\documentclass[runningheads]{llncs}
\usepackage{graphicx}
\usepackage{amsmath,amssymb} % define this before the line numbering.
\usepackage{color}

\usepackage{pdfpages}

\usepackage[width=122mm,left=12mm,paperwidth=146mm,height=193mm,top=12mm,paperheight=217mm]{geometry}

\usepackage{graphicx}

\usepackage{subfigure}
\usepackage{xcolor}

\usepackage{mathtools}
\DeclarePairedDelimiter\floor{\lfloor}{\rfloor}

\newcommand\partialconv{{\textit{partial convolution}}}

\def \saliency {\textup{\saliency}}

\def \path {\mathit{path}}

\begin{document}
% \linenumbers
\pagestyle{headings}
\mainmatter

\title{Image Inpainting for Irregular Holes Using Partial Convolutions} % Replace with your title

\titlerunning{Image Inpainting for Irregular Holes Using Partial Convolutions}

\authorrunning{Guilin Liu et al.}

\author{Guilin Liu \quad Fitsum A. Reda \quad Kevin J. Shih \quad Ting-Chun Wang \\ Andrew Tao \quad Bryan Catanzaro}
\institute{NVIDIA Corporation}

\maketitle

% \teaser{
\begin{figure}
\centering
\includegraphics[width=0.2\columnwidth]{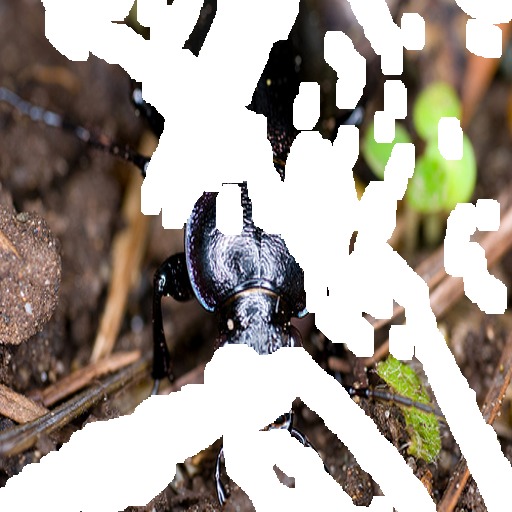}
\includegraphics[width=0.2\columnwidth]{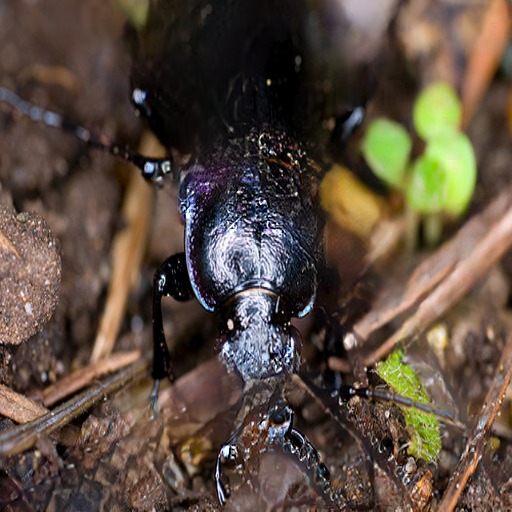}
\includegraphics[width=0.2\columnwidth]{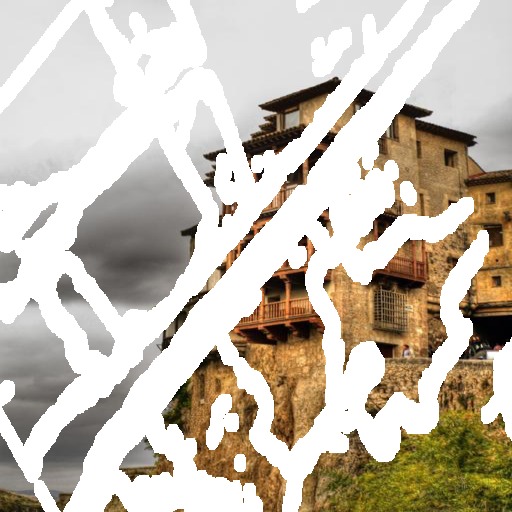}
\includegraphics[width=0.2\columnwidth]{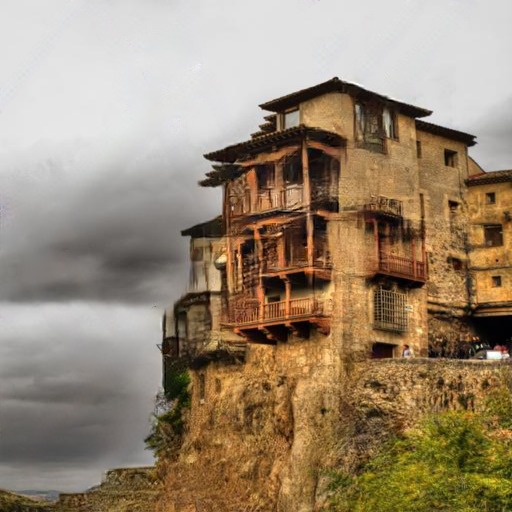} \\
\includegraphics[width=0.2\columnwidth]{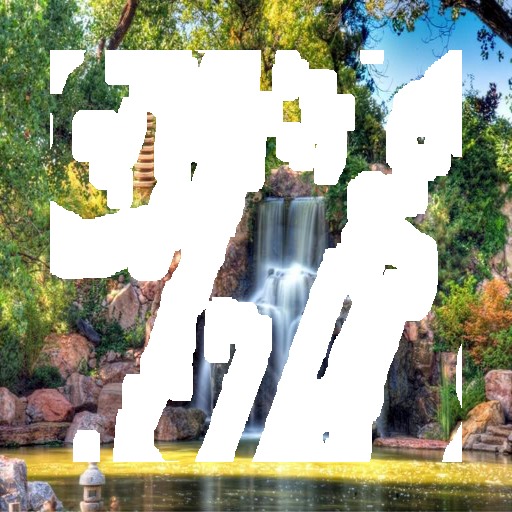} 
\includegraphics[width=0.2\columnwidth]{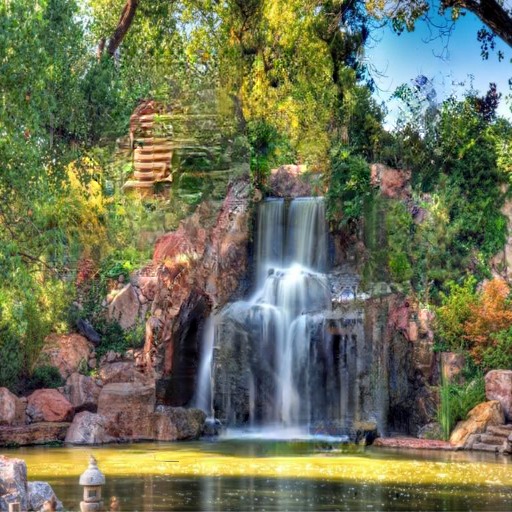} 
\includegraphics[width=0.2\columnwidth]{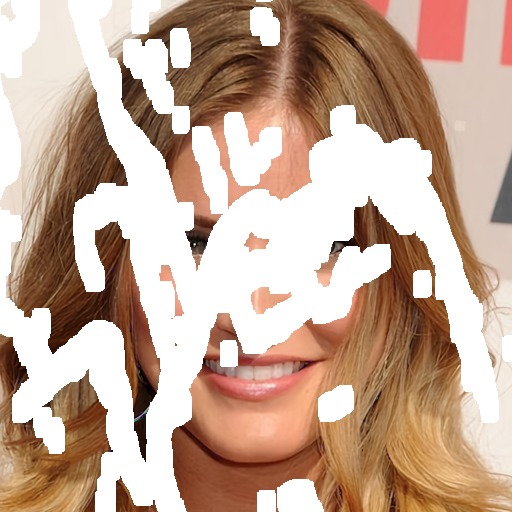}
\includegraphics[width=0.2\columnwidth]{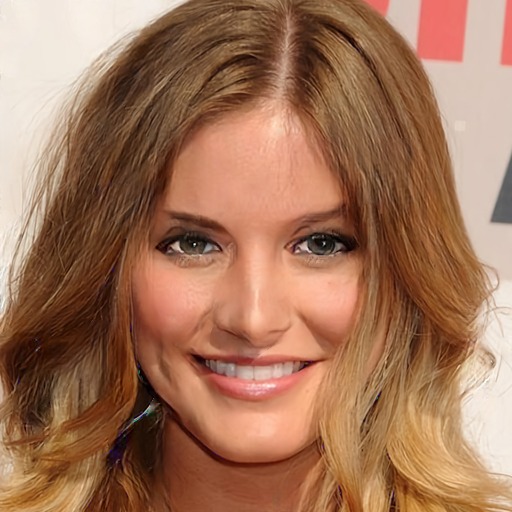} \\
\caption{Masked images and corresponding inpainted results using our partial-convolution based network.}
\label{fig:example}
\end{figure}
%  }

\begin{abstract}

Existing deep learning based image inpainting methods use a standard convolutional network over the corrupted image, using convolutional filter responses conditioned on both valid pixels as well as the substitute values in the masked holes (typically the mean value). This often leads to artifacts such as color discrepancy and blurriness. Post-processing is usually used to reduce such artifacts, but are expensive and may fail. We propose the use of partial convolutions, where the convolution is masked and renormalized to be conditioned on only valid pixels. We further include a mechanism to automatically generate an updated mask for the next layer as part of the forward pass. Our model outperforms other methods for irregular masks. We show qualitative and quantitative comparisons with other methods to validate our approach.

\keywords{Partial Convolution, Image Inpainting}
\end{abstract}

\section{Introduction}
\begin{figure}[ht!]
    \label{fig:intro}
    \centering
    \subfigure[Image with hole]{\label{fig:intro_input}\includegraphics[width=0.23\columnwidth]{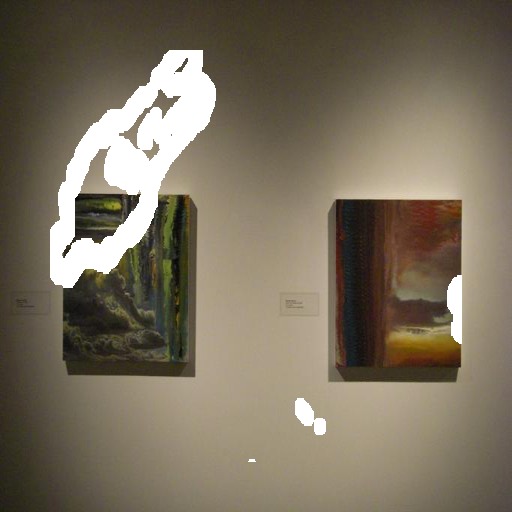}}
    \subfigure[PatchMatch]{\label{fig:intro_patchmatch}\includegraphics[width=0.23\columnwidth]{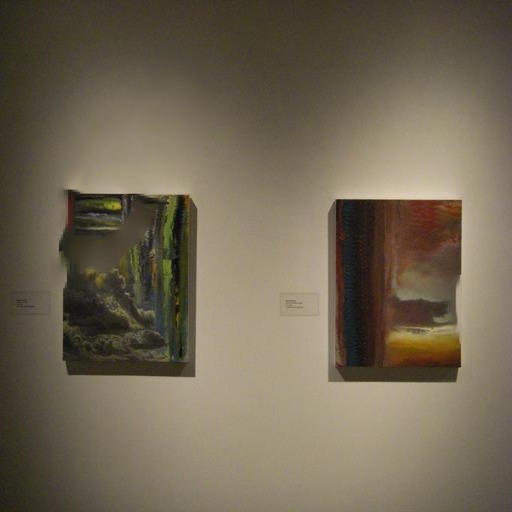}}
    \subfigure[Iizuka et al.\cite{iizuka2017globally}]{\label{fig:intro_globallocal}\includegraphics[width=0.23\columnwidth]{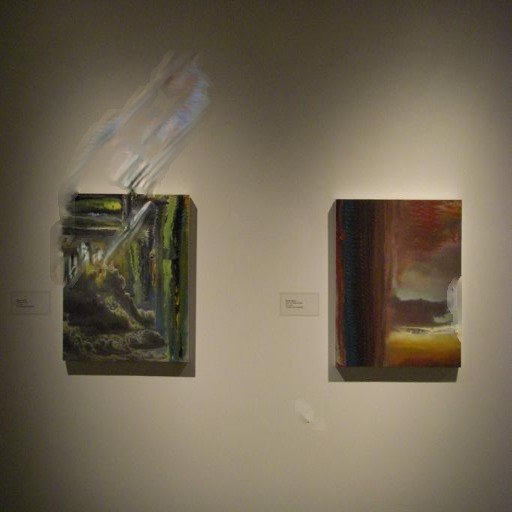}}
    \subfigure[Yu et al.\cite{yu2018generative}]{\label{fig:intro_genInpaint}\includegraphics[width=0.23\columnwidth]{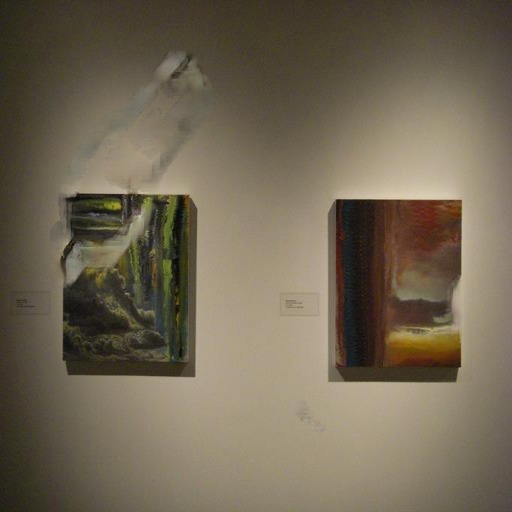}} \\
    \subfigure[Hole$=$127.5]{\label{fig:intro_127init}\includegraphics[width=0.23\columnwidth]{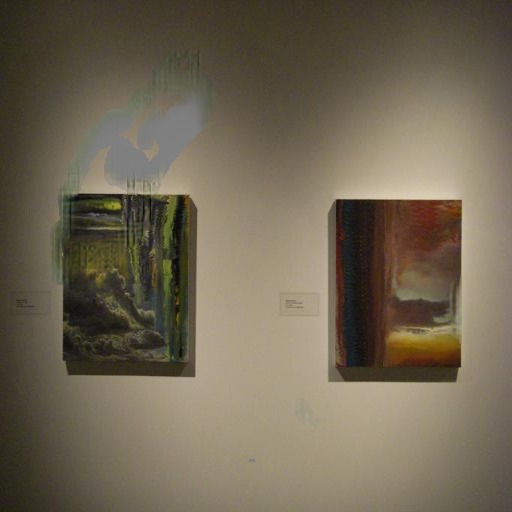}}
    \subfigure[Hole=IN\_Mean]{\label{fig:intro_imgnetInitHyper}\includegraphics[width=0.23\columnwidth]{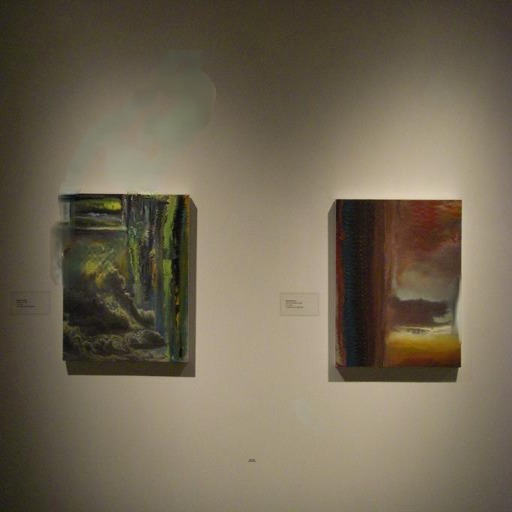}}
    \subfigure[Partial Conv]{\label{fig:intro_partialconv}\includegraphics[width=0.23\columnwidth]{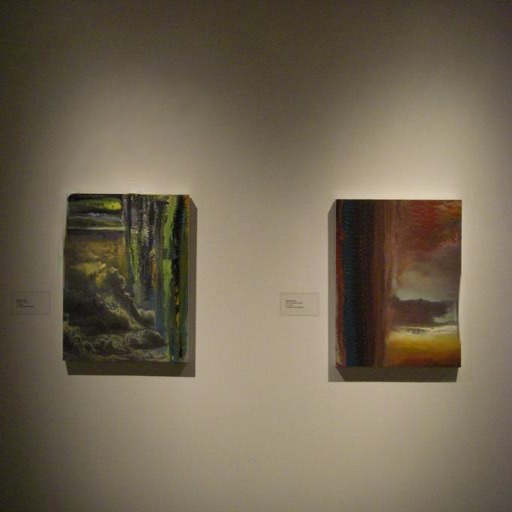}}
    \subfigure[Ground Truth]{\label{fig:intro_gt}\includegraphics[width=0.23\columnwidth]{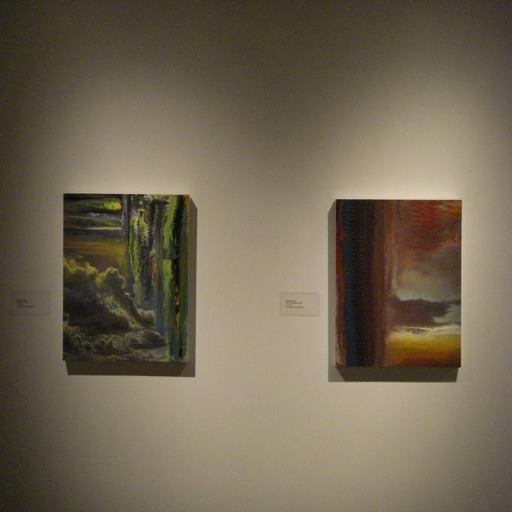}
    }
    \caption{From left to right, top to bottom: \ref{fig:intro_input}: image with hole. \ref{fig:intro_patchmatch}: inpainted result of PatchMatch\cite{barnes2009patchmatch}. \ref{fig:intro_globallocal}: inpainted result of Iizuka et al.\cite{iizuka2017globally}. \ref{fig:intro_genInpaint}: Yu et al.\cite{yu2018generative}. \ref{fig:intro_127init} and \ref{fig:intro_imgnetInitHyper} are using the same network architecture as Section~\ref{sec:sub_network} but using typical convolutional network, \ref{fig:intro_127init} uses the pixel value 127.5 to initialize the holes. \ref{fig:intro_imgnetInitHyper} uses the mean ImageNet pixel value. \ref{fig:intro_partialconv}: our \partialconv\ based results which are agnostic to hole values.}
\end{figure}

Image inpainting, the task of filling in holes in an image, can be used in many applications. For example, it can be used in image editing to remove unwanted image content, while filling in the resulting space with plausible imagery. Previous deep learning approaches have focused on rectangular regions located around the center of the image, and often rely on expensive post-processing. The goal of this work is to propose a model for image inpainting that operates robustly on irregular hole patterns (see Fig. \ref{fig:intro}), and produces semantically meaningful predictions that incorporate smoothly with the rest of the image without the need for any additional post-processing or blending operation.

Recent image inpainting approaches that do not use deep learning use image statistics of the remaining image to fill in the hole. PatchMatch~\cite{barnes2009patchmatch}, one of the state-of-the-art methods, iteratively searches for the best fitting patches to fill in the holes. While this approach generally produces smooth results, it is limited by the available image statistics and has no concept of visual semantics. For example, in Figure~\ref{fig:intro_patchmatch}, PatchMatch was able to smoothly fill in the missing components of the painting using image patches from the surrounding shadow and wall, but a semantically-aware approach would make use of patches from the painting instead.

Deep neural networks learn semantic priors and meaningful hidden representations in an end-to-end fashion, which have been used for recent image inpainting efforts. These networks employ convolutional filters on images, replacing the removed content  with a fixed value. As a result, these approaches suffer from dependence on the initial hole values, which often manifests itself as lack of texture in the hole regions, obvious color contrasts, or artificial edge responses surrounding the hole. Examples using a U-Net architecture with typical convolutional layers with various hole value initialization can be seen in Figure~\ref{fig:intro_127init} and \ref{fig:intro_imgnetInitHyper}. (For both, the training and testing share the same initalization scheme). 

Conditioning the output on the hole values ultimately results in various types of visual artifacts that necessitate expensive post-processing. For example, Iizuka et al. \cite{iizuka2017globally} uses fast marching \cite{telea2004image} and Poisson image blending \cite{perez2003poisson}, while Yu et al. \cite{yu2018generative} employ a following-up refinement network to refine their raw network predictions. However, these refinement cannot resolve all the artifacts shown as \ref{fig:intro_globallocal} and \ref{fig:intro_genInpaint}. Our work aims to achieve well-incorporated hole predictions independent of the hole initialization values and without any additional post-processing.

Another limitation of many recent approaches is the focus on rectangular shaped holes, often assumed to be center in the image. We find these limitations may lead to overfitting to the rectangular holes, and ultimately limit the utility of these models in application. Pathak et al. \cite{pathak2016context} and Yang et al. ~\cite{yang2017high} assume $64\times64$ square holes at the center of a 128$\times$128 image. Iizuka et al. \cite{iizuka2017globally} and Yu et al. \cite{yu2018generative} remove the centered hole assumption and can handle irregular shaped holes, but do not perform an extensive quantitative analysis on a large number of images with irregular masks (51 test images in ~\cite{hays2007scene}). In order to focus on the more practical irregular hole use case, we collect a large benchmark of images with irregular masks of varying sizes. In our analysis, we look at the effects of not just the size of the hole, but also whether the holes are in contact with the image border.

To properly handle irregular masks, we propose the use of a \emph{Partial Convolutional Layer}, comprising a masked and re-normalized convolution operation followed by a mask-update step. The concept of a masked and re-normalized convolution is also referred to as  segmentation-aware convolutions in ~\cite{harley2017segmentation} for the image segmentation task, however they did not make modifications to the input mask. Our use of partial convolutions is such that  given a binary mask our convolutional results depend only on the non-hole regions at every layer. Our main extension is the automatic mask update step, which removes any masking where the partial convolution was able to operate on an unmasked value. Given sufficient layers of successive updates, even the largest masked holes will eventually shrink away, leaving only valid responses in the feature map. The partial convolutional layer ultimately makes our model agnostic to placeholder hole values.

In summary, we make the following contributions:
\begin{itemize}
\item we propose the the use of \emph{partial convolutions} with an \emph{automatic mask update step} for achieving state-of-the-art on image inpainting. % and with potential applications to other tasks with incomplete data, such as 3D point clouds.
\item while previous works fail to achieve good inpainting results with skip links in a U-Net \cite{ulyanov2017deep} with typical convolutions, we demonstrate that substituting convolutional layers with partial convolutions and mask updates can achieve state-of-the-art inpainting results.
\item to the best of our knowledge, we are the first to demonstrate the efficacy of training image-inpainting models on irregularly shaped holes.
\item we propose a large irregular mask dataset, which will be released to public to facilitate future efforts in training and evaluating inpainting models.
\end{itemize}

\section{Related Work}
Non-learning approaches to image inpainting rely on propagating appearance information from neighboring pixels to the target region using some mechanisms like distance field\cite{bertalmio2000image,ballester2001filling,telea2004image}. However, these methods can only handle narrow holes, where the color and texture variance is small. Big holes may result in over-smoothing or artifacts resembling Voronoi regions such as in~\cite{telea2004image}. Patch-based methods such as~\cite{efros2001image,kwatra2005texture} operate by searching for relevant patches from the image's non-hole regions or other source images in an iterative fashion. However, these steps often come at a large computation cost such as in~\cite{simakov2008summarizing}. PatchMatch~\cite{barnes2009patchmatch} speeds it up by proposing a faster similar patch searching algorithm. However, these approaches are still not fast enough for real-time applications and cannot make semantically aware patch selections.

Deep learning based methods typically initialize the holes with some constant placeholder values e.g. the mean pixel value of ImageNet \cite{ILSVRC15}, which is then passed through a convolutional network. Due to the resulting artifacts, post-processing is often used to ameliorate the effects of conditioning on the placeholder values. Content Encoders \cite{pathak2016context} first embed the 128$\times$128 image with 64$\times$64 center hole into low dimensional feature space and then decode the feature to a 64x64 image. Yang et al. \cite{yang2017high} takes the result from Content Encoders as input and then propagates the texture information from non-hole regions to fill the hole regions as postprocessing. Song et al. \cite{song2017image} uses a refinement network in which a blurry initial hole-filling result is used as the input, then iteratively replaced with patches from the closest non-hole regions in the feature space. Li et al. \cite{li2017generative} and Iizuka et al. \cite{iizuka2017globally} extended Content Encoders by defining both global and local discriminators; then Iizuka et al. \cite{iizuka2017globally} apply Poisson blending as a post-process. Following \cite{iizuka2017globally}, Yu et al. \cite{yu2018generative} replaced the post-processing with a refinement network powered by the contextual attention layers.  

Amongst the deep learning approaches, several other efforts also ignore the mask placeholder values. In Yeh et al. \cite{yeh2016semantic}, searches for the closest encoding to the corrupted image in a latent space, which is then used to condition the output of a hole-filling generator. Ulyanov et al. \cite{ulyanov2017deep} further found that the network needs no external dataset training and can rely on the structure of the generative network itself to complete the corrupted image. However, this approach can require a different set of hyper parameters for every image, and applies several iterations to achieve good results. Moreover, their design \cite{ulyanov2017deep} is not able to use skip links, which are known to produce detailed output. With standard convolutional layers, the raw features of noise or wrong hole initialization values in the encoder stage will propagate to the decoder stage. Our work also does not depend on placeholder values in the hole regions, but we also aim to achieve good results in a single feedforward pass and enable the use of skip links to create detailed predictions.

Our work makes extensive use of a masked or reweighted convolution operation, which allows us to condition output only on valid inputs. Harley et al.~\cite{harley2017segmentation} recently made use of this approach with a soft attention mask for semantic segmentation. It has also been used for full-image generation in PixelCNN~\cite{van2016conditional}, to condition the next pixel only on previously synthesized pixels.
Uhrig et al.~\cite{uhrig2017sparsity} proposed sparsity invariant CNNs with reweighted convolution and max pooling based mask updating mechanism for depth completion. For image inpainting, Ren et al.~\cite{ren2015shepard} proposed shepard convolution layer where the same kernel is applied for both feature and mask convolutions. The mask convolution result acts as both the reweighting denominator and updated mask, which does not guarantee the hole to evolve during updating due to the kernel's possible negative entries. It cannot handle big holes properly either. Discussions of other CNN variants like~\cite{dai2017deformable} are beyond the scope of this work.

\section{Approach}
\label{sec:approach}

Our proposed model uses stacked partial convolution operations and mask updating steps to perform image inpainting. We first define our convolution and mask update mechanism, then discuss model architecture and loss functions.

\subsection{Partial Convolutional Layer}
We refer to our partial convolution operation and mask update function jointly as the \emph{Partial Convolutional Layer}. Let $\mathbf{W}$ be the convolution filter weights for the convolution filter and $b$ its the corresponding bias. $\mathbf{X}$ are the feature values (pixels values) for the current convolution (sliding) window and $\mathbf{M}$ is the corresponding binary mask. The partial convolution at every location, similarly defined in~\cite{harley2017segmentation}, is expressed as:

\begin{equation}
\label{eq:partconv}
    x' = \begin{cases}
     \mathbf{W}^T(\mathbf{X}\odot\mathbf{M})\frac{\text{sum}(\mathbf{1})}{\text{sum}(\mathbf{M})} + b,  & \text{if }  \text{sum}(\mathbf{M})>0  \\
        0,  & \text{otherwise}
    \end{cases}
\end{equation}
where $\odot$ denotes element-wise multiplication, and $\mathbf{1}$ has same shape as $M$ but with all elements being 1. As can be seen, output values depend only on the unmasked inputs. The scaling factor $\text{sum}(\mathbf{1})/\text{sum}(\mathbf{M})$ applies appropriate scaling to adjust for the varying amount of valid (unmasked) inputs.

After each partial convolution operation, we then update our mask as follows: if the convolution was able to condition its output on at least one valid input value, then we mark that location to be valid. This is expressed as:
\begin{equation}
m'=\begin{cases}
    1, & \text{if } \text{sum}(\mathbf{M})>0  \\
    0, & \text{otherwise}
\end{cases}
\end{equation}
and can easily be implemented in any deep learning framework as part of the forward pass. With sufficient successive applications of the partial convolution layer, any mask will eventually be all ones, if the input contained any valid pixels.

\subsection{Network Architecture and Implementation}
\label{sec:sub_network}

\textbf{Implementation}. Partial convolution layer is implemented by extending existing standard PyTorch\cite{paszke2017automatic}, although it can be improved both in time and space using custom layers. The straightforward implementation is to define binary masks of size C$\times$H$\times$W, the same size with their associated images/features, and then to implement mask updating is implemented using a fixed convolution layer, with the same kernel size as the partial convolution operation, but with weights identically set to 1 and no bias. The entire network inference on a 512$\times$512 image takes 0.029s on a single NVIDIA V100 GPU, regardless of the hole size.

\textbf{Network Design}. We design a UNet-like architecture~\cite{ronneberger2015u} similar to the one used in \cite{isola2017image}, replacing all convolutional layers with partial convolutional layers and using nearest neighbor up-sampling in the decoding stage.
The skip links will concatenate two feature maps and two masks respectively, acting as the feature and mask inputs for the next partial convolution layer. The last partial convolution layer's input will contain the concatenation of the original input image with hole and original mask, making it possible for the model to copy non-hole pixels. Network details are found in the supplementary file.

\textbf{Partial Convolution as Padding}. We use the partial convolution with appropriate masking at image boundaries in lieu of typical padding . This ensures that the inpainted content at the image border will not be affected by invalid values outside of the image -- which can be interpreted as another hole.

\subsection{Loss Functions}
\label{sec:loss_func}

Our loss functions target both per-pixel reconstruction accuracy as well as composition, i.e. how smoothly the predicted hole values transition into their surrounding context.

Given input image with hole $\mathbf{I}_{in}$, initial binary mask $\mathbf{M}$ (0 for holes), the network prediction $\mathbf{I}_{out}$, and the ground truth image $\mathbf{I}_{gt}$, we first define our per-pixel losses  $\mathcal{L}_{hole} = \frac{1}{N_{\mathbf{I}_{gt}}}\|(1-M)\odot(\mathbf{I}_{out} - \mathbf{I}_{gt})\|_1$, $\mathcal{L}_{valid} = \frac{1}{N_{\mathbf{I}_{gt}}}\|M \odot (\mathbf{I}_{out} -\mathbf{I}_{gt})\|_1$, where $N_{\mathbf{I}_{gt}}$ denotes the number of elements in $\mathbf{I}_{gt}$ ($N_{\mathbf{I}_{gt}} = C * H * W$ and C, H and W are the channel size, height and width of image $\mathbf{I}_{gt}$). These are the $L^1$ losses on the network output for the hole and the non-hole pixels respectively.

Next, we define the perceptual loss, introduced by Gatys et al.~\cite{gatys2015neural}:
\begin{equation}
    \mathcal{L}_{perceptual} = \sum_{p=0}^{P-1} {\frac{\|{\Psi}_{p}^{\mathbf{I}_{out}}-{\Psi}_{p}^{\mathbf{I}_{gt}}\|_{1}}{N_{{\Psi}_{p}^{\mathbf{I}_{gt}}}}} + \sum_{p=0}^{P-1} {\frac{\|{\Psi}_{p}^{\mathbf{I}_{comp}}-{\Psi}_{p}^{\mathbf{I}_{gt}}\|_{1}}{N_{{\Psi}_{p}^{\mathbf{I}_{gt}}}}}
\end{equation}
Here, $\mathbf{I}_{comp}$ is the raw output image $\mathbf{I}_{out}$, but with the non-hole pixels directly set to ground truth; $N_{{\Psi}_{p}^{\mathbf{I}_{gt}}}$ is the number of elements in ${\Psi}_{p}^{\mathbf{I}_{gt}}$. The perceptual loss computes the $L^1$ distances between both $\mathbf{I}_{out}$ and $\mathbf{I}_{comp}$ and the ground truth, but after projecting these images into higher level feature spaces using an ImageNet-pretrained VGG-16~\cite{simonyan2014very}. ${\Psi}_{p}^{\mathbf{I}_{*}}$ is the activation map of the $p$th selected layer given original input $\mathbf{I}_{*}$. We use layers $pool1$, $pool2$ and $pool3$ for our loss.  

We further include the style-loss term, which is similar to the perceptual loss~\cite{gatys2015neural}, but we first perform an autocorrelation (Gram matrix) on each feature map before applying the $L^1$.
\begin{equation}
 \mathcal{L}_{style_{out}} = \sum_{p=0}^{P-1}{\frac{1}{C_{p} C_{p}} {\Big|\Big|K_p\big({{\big(\Psi}_{p}^{\mathbf{I}_{out}}\big)}^{\intercal}{{\big(\Psi}_{p}^{\textbf{I}_{out}}\big)} - {\big({\Psi}_{p}^{\mathbf{I}_{gt}}\big)}^{\intercal}{\big({\Psi}_{p}^{\mathbf{I}_{gt}}\big)\big)}\Big|\Big|}_{1} }
 \end{equation}
 \begin{equation}
    \mathcal{L}_{style_{comp}} = \sum_{p=0}^{P-1}{\frac{1}{C_{p} C_{p}} {\Big|\Big|K_p\big({{\big(\Psi}_{p}^{\mathbf{I}_{comp}}\big)}^{\intercal}{{\big(\Psi}_{p}^{\mathbf{I}_{comp}}\big)} - {\big({\Psi}_{p}^{\mathbf{I}_{gt}}\big)}^{\intercal}{\big({\Psi}_{p}^{\mathbf{I}_{gt}}\big)\big)}\Big|\Big|}_{1} }
\end{equation}
Here, we note that the matrix operations assume that the high level features $\Psi(x)_p$ is of shape $(H_pW_p)\times C_p$, resulting in a $C_p\times C_p$ Gram matrix, and $K_p$ is the normalization factor $1/C_pH_pW_p$ for the $p$th selected layer. Again, we include loss terms for both raw output and composited output.

Our final loss term is the total variation (TV) loss $\mathcal{L}_{tv}$: which is the smoothing penalty \cite{johnson2016perceptual} on $R$, where $R$ is the region of 1-pixel dilation of the hole region.
\begin{equation}
    \mathcal{L}_{tv} = \sum_{(i,j) \in R, (i,j+1) \in R} {\frac{\|\mathbf{I}_{comp}^{i,j+1} - \mathbf{I}_{comp}^{i,j}\|_{1}}{N_{\mathbf{I}_{comp}}}} + \sum_{(i,j) \in R, (i+1,j) \in R}{\frac{\|\mathbf{I}_{comp}^{i+1,j} - \mathbf{I}_{comp}^{i,j}\|_{1}}{N_{\mathbf{I}_{comp}}}}
\end{equation}
where, $N_{\mathbf{I}_{comp}}$ is the number of elements in $\mathbf{I}_{comp}$.

The total loss $\mathcal{L}_{total}$ is the combination of all the above loss functions.
\begin{equation}
    \mathcal{L}_{total} = \mathcal{L}_{valid} + 6\mathcal{L}_{hole} + 0.05\mathcal{L}_{perceptual} + 120(\mathcal{L}_{style_{out}} + \mathcal{L}_{style_{comp}}) + 0.1\mathcal{L}_{tv}
\end{equation} 

The loss term weights were determined by performing a hyperparameter search on 100 validation images.

\begin{figure}[t]
\centering
\subfigure[Input]{\label{fig:vggstyle_input}{\includegraphics[width=0.2\columnwidth]{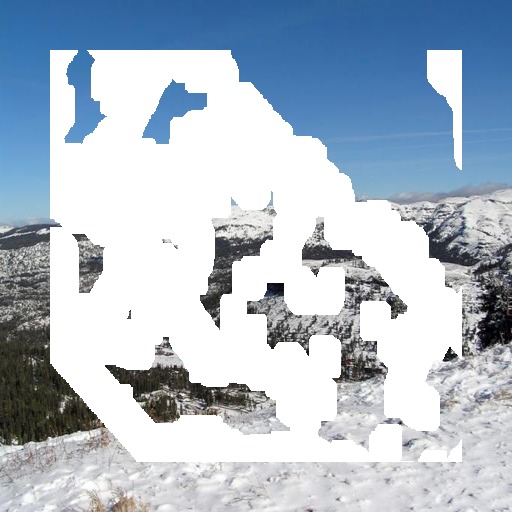}}}
\subfigure[no $\mathcal{L}_{style}$]{\label{fig:vggstyle_nostyle}{\includegraphics[width=0.2\columnwidth]{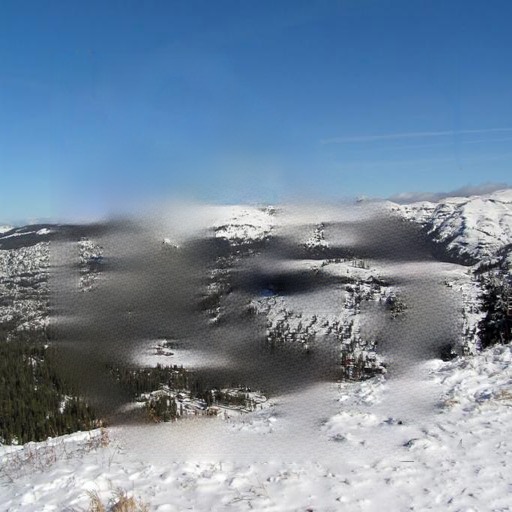}}}
\subfigure[full $\mathcal{L}_{total}$]{\label{fig:vggstyle_style}{\includegraphics[width=0.2\columnwidth]{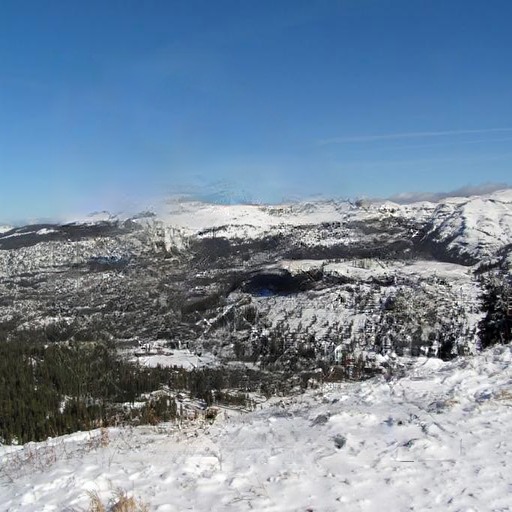}}}
\subfigure[GT]{\label{fig:vggstyle_input}{\includegraphics[width=0.2\columnwidth]{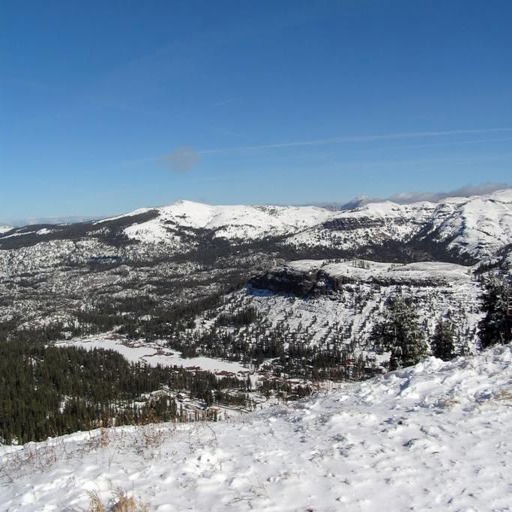}}} \\
\subfigure[Input]{\label{fig:stylewgt_input}\includegraphics[width=0.2\columnwidth]{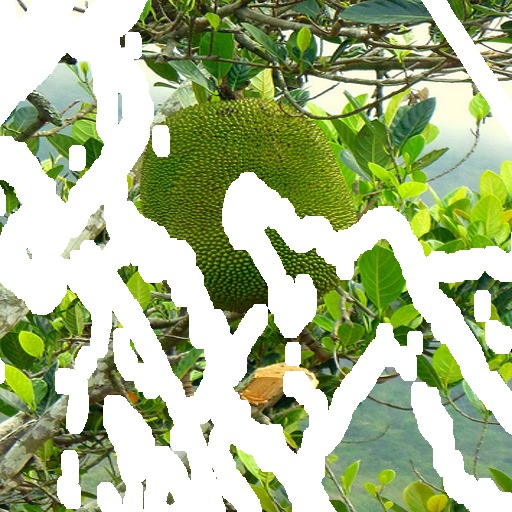}}
\subfigure[Small $\mathcal{L}_{style}$]{\label{fig:stylewgt_small}\includegraphics[width=0.2\columnwidth]{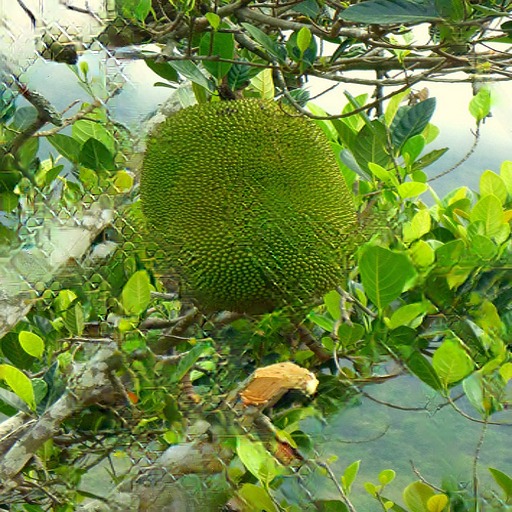}}
\subfigure[full $\mathcal{L}_{total}$]{\label{fig:stylewgt_full}\includegraphics[width=0.2\columnwidth]{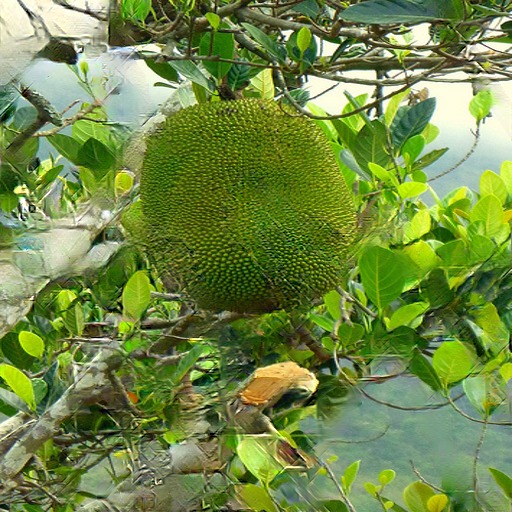}}
\subfigure[GT]{\label{fig:stylewgt_gt}\includegraphics[width=0.2\columnwidth]{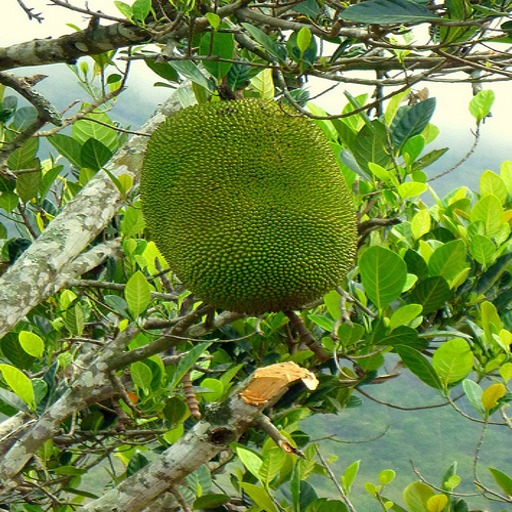}} \\
\subfigure[Input]{\label{fig:perceptua_input}\includegraphics[width=0.2\columnwidth]{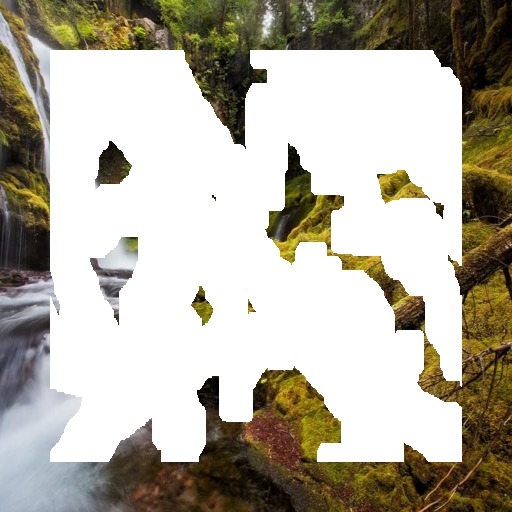}}
\subfigure[no $\mathcal{L}_{perceptual}$]{\label{fig:perceptua_no}\includegraphics[width=0.2\columnwidth]{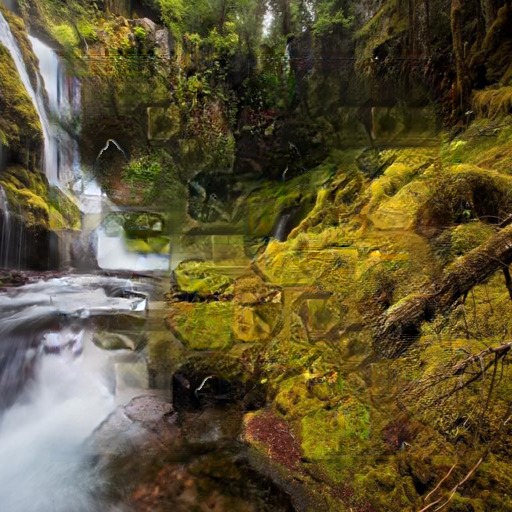}}
\subfigure[full $\mathcal{L}_{total}$]{\label{fig:perceptua_full}\includegraphics[width=0.2\columnwidth]{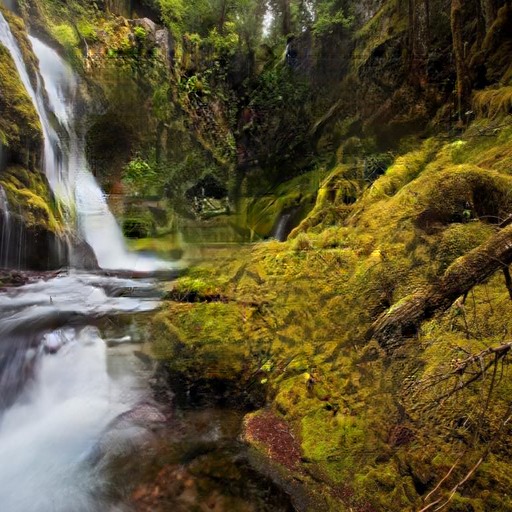}}
\subfigure[GT]{\label{fig:perceptua_gt}\includegraphics[width=0.2\columnwidth]{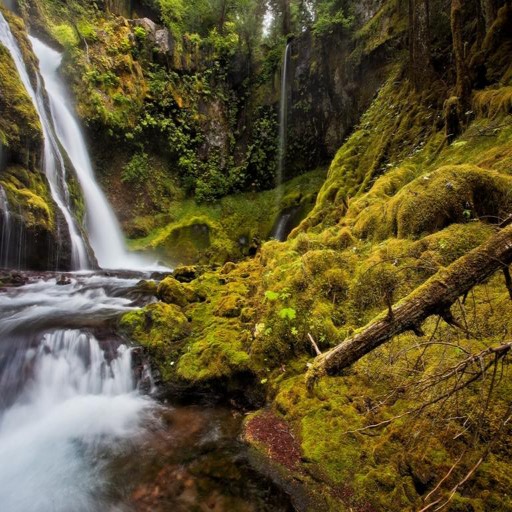}}
\caption{In top row, from left to right: input image with hole, result without style loss, result using full $\mathcal{L}_{total}$, and ground truth. In middle row, from left to right: input image with hole, result using small style loss weight, result using full $\mathcal{L}_{total}$, and ground truth. In bottom row, from left to right: input image with hole, result without perceptual loss, result using full $\mathcal{L}_{total}$, and ground truth.}
\label{fig:vggstyle}
\end{figure}

\textbf{Ablation Study of Different Loss Terms}. Perceptual loss~\cite{johnson2016perceptual} is known to generate \emph{checkerboard artifacts}. Johnson et al. \cite{johnson2016perceptual} suggests to ameliorate the problem by using the total variation (TV) loss. We found this not to be the case for our model. Figure~\ref{fig:vggstyle_nostyle} shows the result of the model trained by removing $\mathcal{L}_{style_{out}}$ and $\mathcal{L}_{style_{comp}}$ from $\mathcal{L}_{total}$. For our model, the additional style loss term is necessary. However, not all the loss weighting schemes for the style loss will generate plausible results. Figure~\ref{fig:stylewgt_small} shows the result of the model trained with a small style loss weight. Compared to the result of the model trained with full $\mathcal{L}_{total}$ in Figure~\ref{fig:stylewgt_full}, it has many \emph{fish scale artifacts}. However, perceptual loss is also important; grid-shaped artifacts are less prominent in the results with full $\mathcal{L}_{total}$ (Figure~\ref{fig:perceptua_full}) than the results without perceptual loss (Figure~\ref{fig:perceptua_no}). We hope this discussion will be useful to readers interested in employing VGG-based high level losses.

\section{Experiments}
\label{sec:experiment}

\begin{figure}%[t]
    \centering
    \scalebox{0.8}{
    \centering
    \includegraphics[width=0.16\columnwidth]{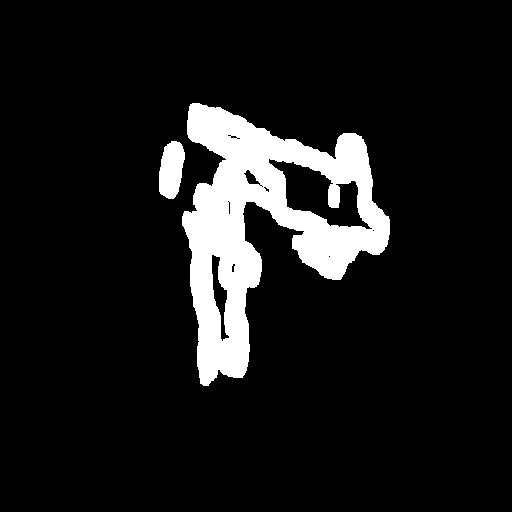}
    \includegraphics[width=0.16\columnwidth]{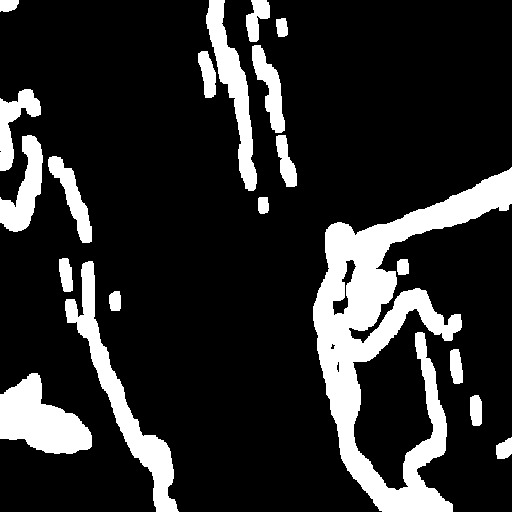}
    \includegraphics[width=0.16\columnwidth]{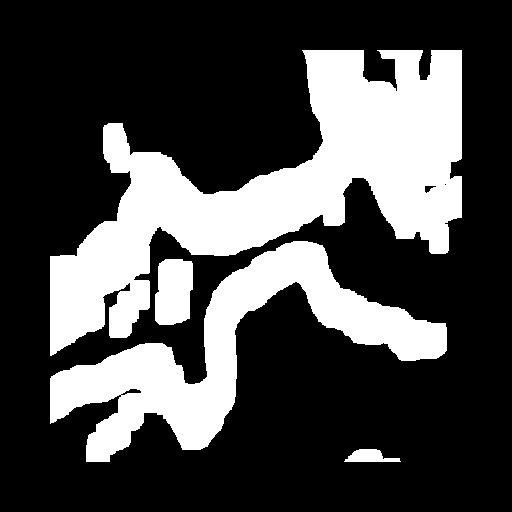}
    \includegraphics[width=0.16\columnwidth]{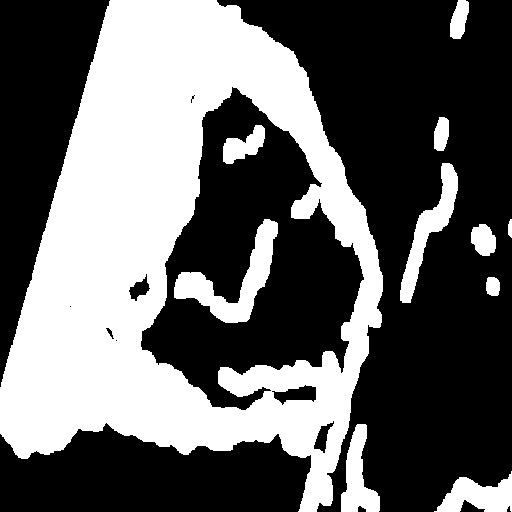}
    \includegraphics[width=0.16\columnwidth]{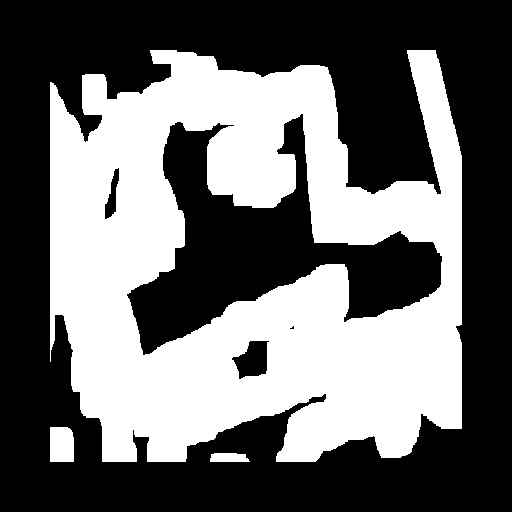}
    \includegraphics[width=0.16\columnwidth]{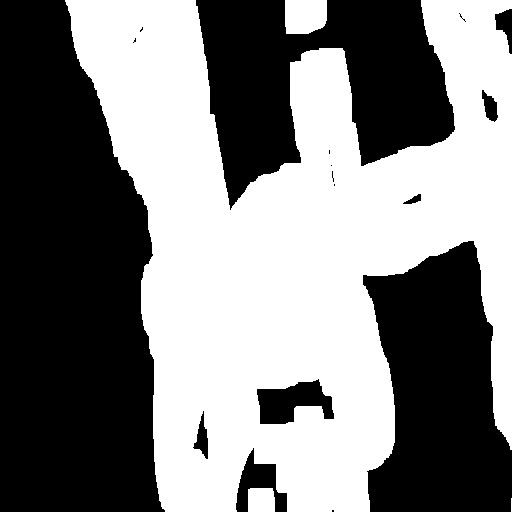}
    }
    \caption{Some test masks for each hole-to-image area ratio category. 1, 3 and 5 are shown using their examples with border constraint; 2, 4 and 6 are shown using their examples without border constraint.}
    \label{fig:rep_mask}
\end{figure}

\subsection{Irregular Mask Dataset}
Previous works generate holes in their datasets by randomly removing rectangular regions within their image. We consider this insufficient in creating the diverse hole shapes and sizes that we need. As such, we begin by collecting masks of random streaks and holes of arbitrary shapes. We found the results of occlusion/dis-occlusion mask estimation method between two consecutive frames for videos described in \cite{sundaram2010dense} to be a good source of such patterns. We generate 55,116 masks for the training and 24,866 masks for testing. During training, we augment the mask dataset by randomly sampling a mask from 55,116 masks and later perform random dilation, rotation and cropping. All the masks and images for training and testing are with the size of 512$\times$512.

We create a test set by starting with the 24,866 raw masks and adding random dilation, rotation and cropping. Many previous methods such as \cite{iizuka2017globally} have degraded performance at holes near the image borders. As such, we divide the test set into two: masks with and without holes close to border. The split that has holes distant from the border ensures a distance of at least 50 pixels from the border. 

We also further categorize our masks by hole size. Specifically, we generate 6 categories of masks with different hole-to-image area ratios: (0.01, 0.1], (0.1, 0.2], (0.2, 0.3], (0.3, 0.4], (0.4, 0.5], (0.5, 0.6]. Each category contains 1000 masks with and without border constraints. In total, we have created $6\times 2 \times1000=12,000$ masks. Some examples of each category's masks can be found in Figure~\ref{fig:rep_mask}.

\subsection{Training Process}

\textbf{Training Data} We use 3 separate image datasets for training and testing: ImageNet dataset \cite{ILSVRC15}, Places2 dataset \cite{zhou2017places} and CelebA-HQ \cite{liu2015faceattributes,karras2017progressive}. We use the original train, test, and val splits for ImageNet and Places2. For CelebA-HQ, we randomly partition into 27K images for training and 3K images for testing. 

\textbf{Training Procedure}.
We initialize the weights using the initialization method described in \cite{he2015delving} and use Adam \cite{kingma2014adam} for optimization. We train on a single NVIDIA V100 GPU (16GB) with a batch size of 6. 

\textbf{Initial Training and Fine-Tuning}. Holes present a problem for Batch Normalization because the mean and variance will be computed for hole pixels, and so it would make sense to disregard them at masked locations. However, holes are gradually filled with each application and usually completely gone by the decoder stage.

In order to use Batch Normalization in the presence of holes, we first turn on Batch Normalization for the initial training using a learning rate of 0.0002. Then, we fine-tune using a learning rate of 0.00005 and freeze the Batch Normalization parameters in the encoder part of the network. We keep Batch Normalization enabled in the decoder. This not only avoids the incorrect mean and variance issues, but also helps us to achieve faster convergence. ImageNet and Places2 models train for 10 days, whereas CelebA-HQ trains in 3 days. All fine-tuning is performed in one day.

\subsection{Comparisons}
\label{sec:sub_comparison}

\begin{figure}%[ht]
    \centering
    \includegraphics[width=0.158\columnwidth]{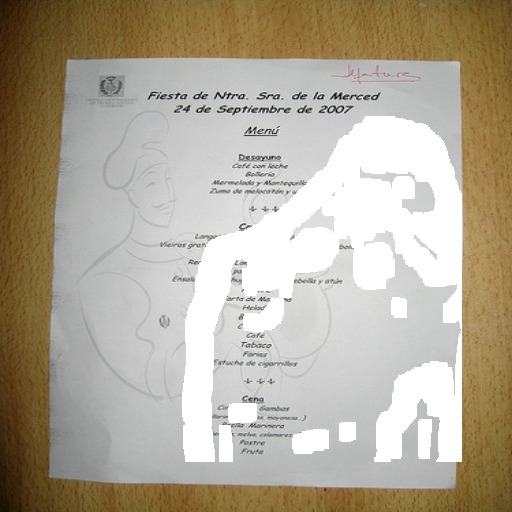}
    \includegraphics[width=0.158\columnwidth]{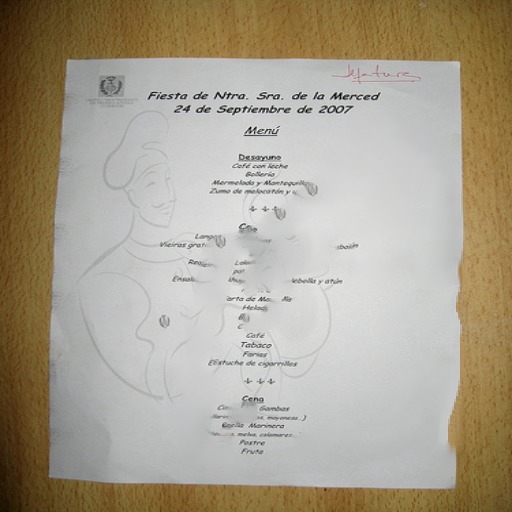}
    \includegraphics[width=0.158\columnwidth]{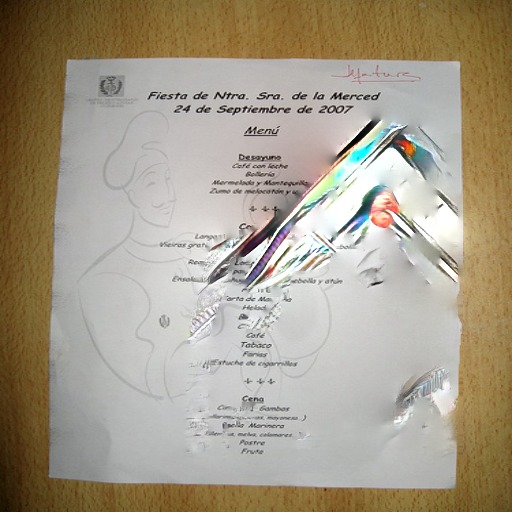}
    \includegraphics[width=0.158\columnwidth]{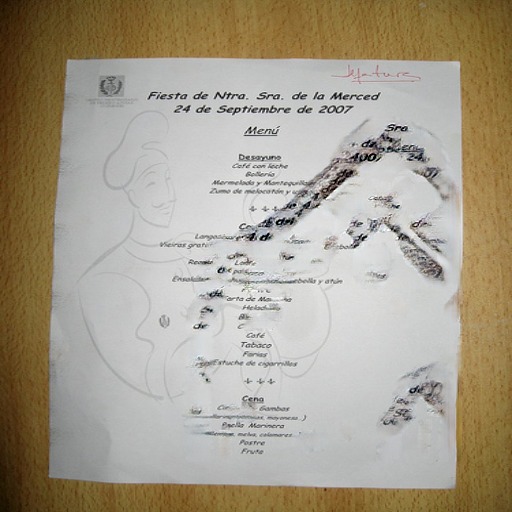}
    \includegraphics[width=0.158\columnwidth]{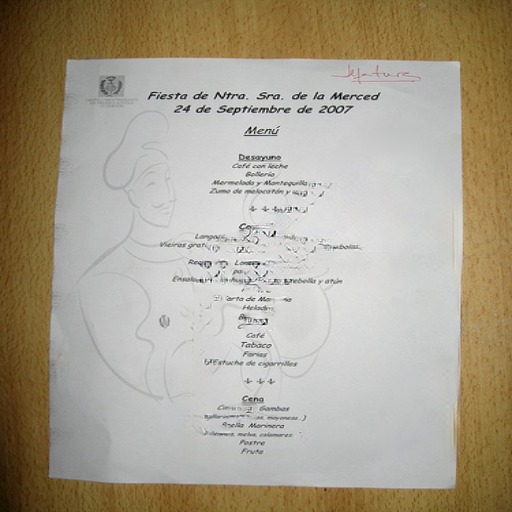}
    \includegraphics[width=0.158\columnwidth]{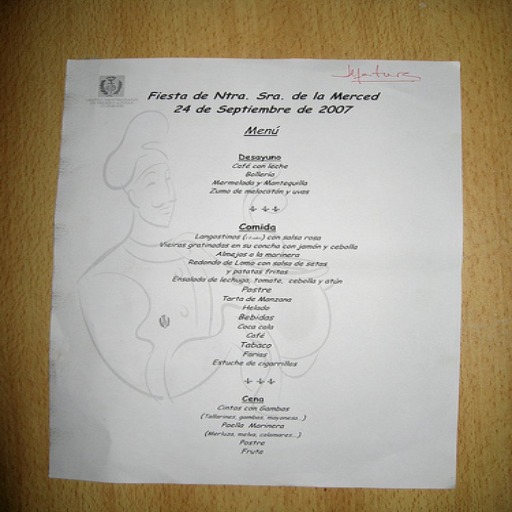} \\        
    \includegraphics[width=0.158\columnwidth]{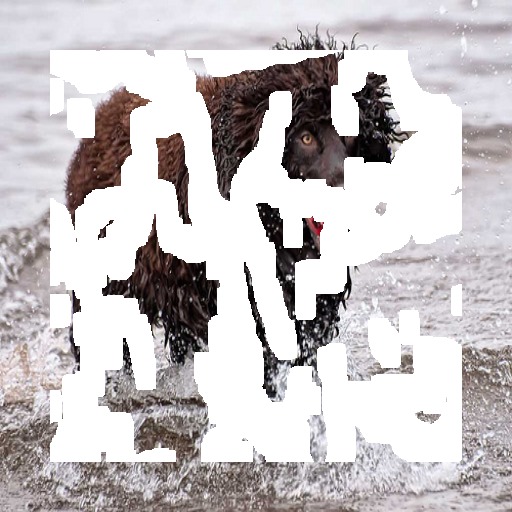}
    \includegraphics[width=0.158\columnwidth]{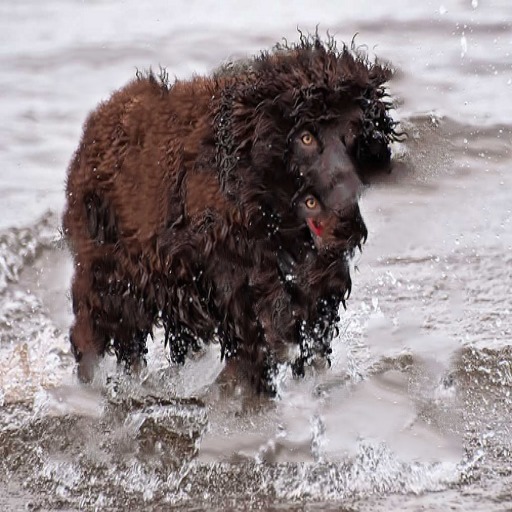}
    \includegraphics[width=0.158\columnwidth]{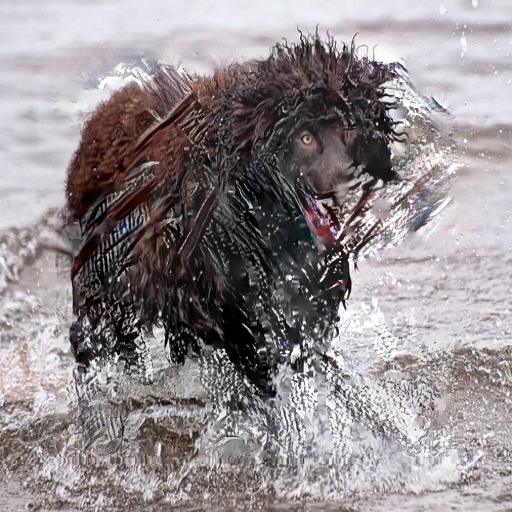}
    \includegraphics[width=0.158\columnwidth]{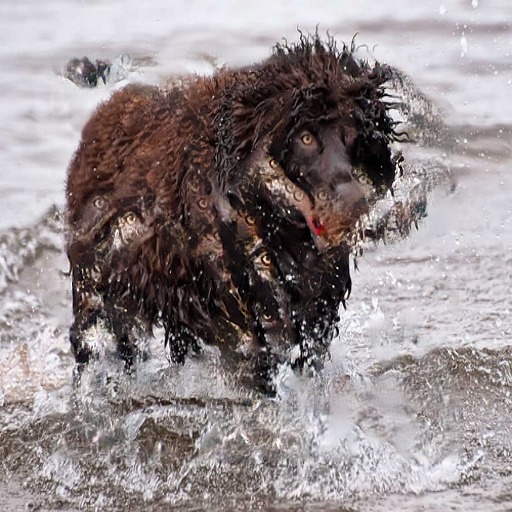}
    \includegraphics[width=0.158\columnwidth]{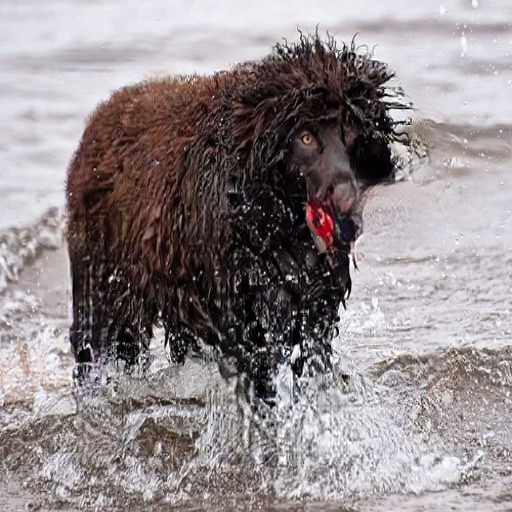}
    \includegraphics[width=0.158\columnwidth]{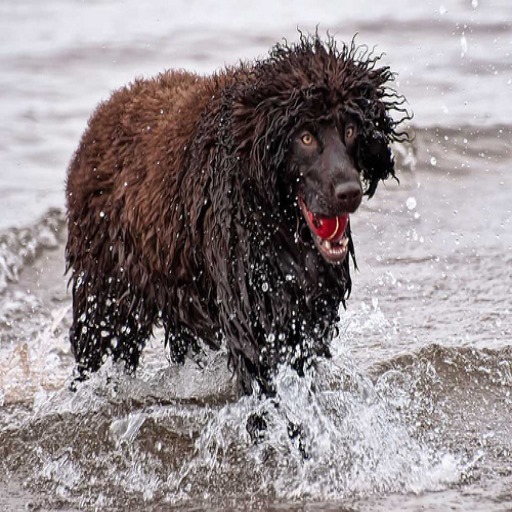} \\    
    \includegraphics[width=0.158\columnwidth]{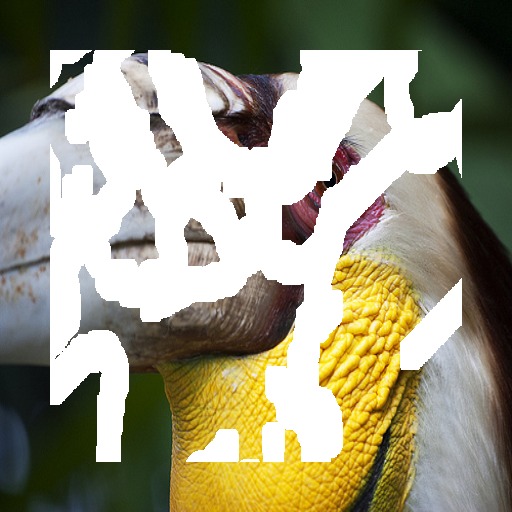}
    \includegraphics[width=0.158\columnwidth]{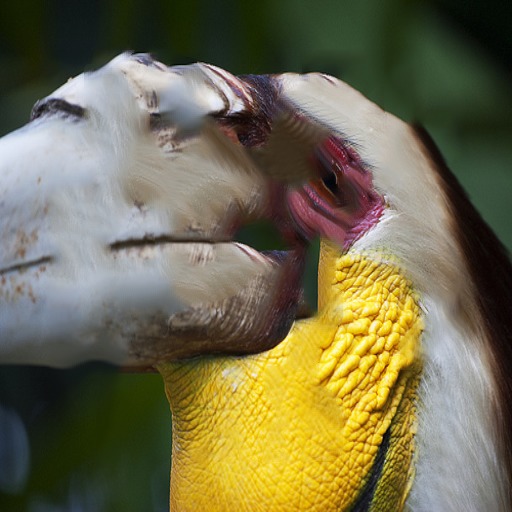}
    \includegraphics[width=0.158\columnwidth]{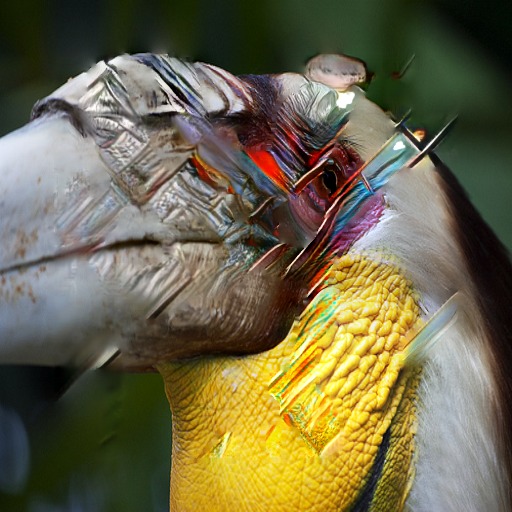}
    \includegraphics[width=0.158\columnwidth]{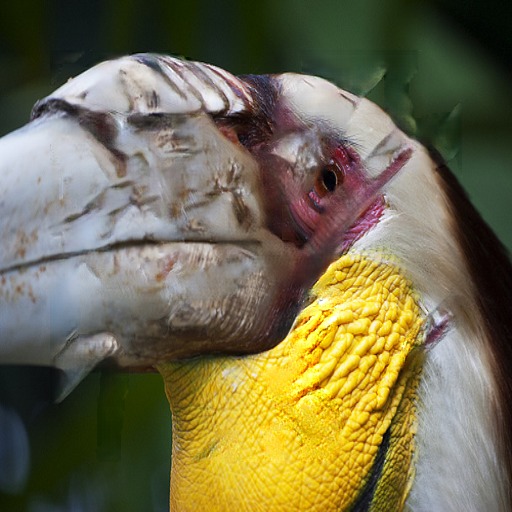}
    \includegraphics[width=0.158\columnwidth]{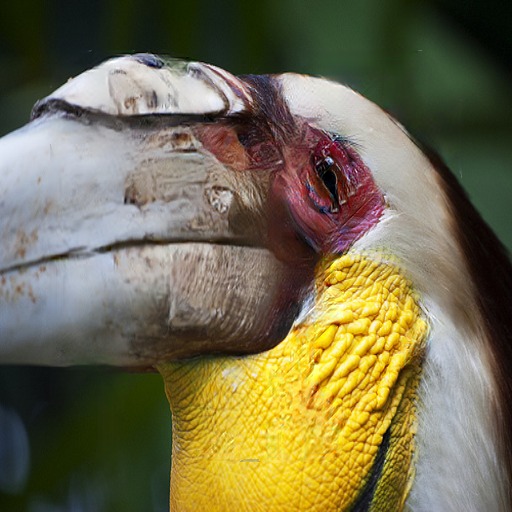}
    \includegraphics[width=0.158\columnwidth]{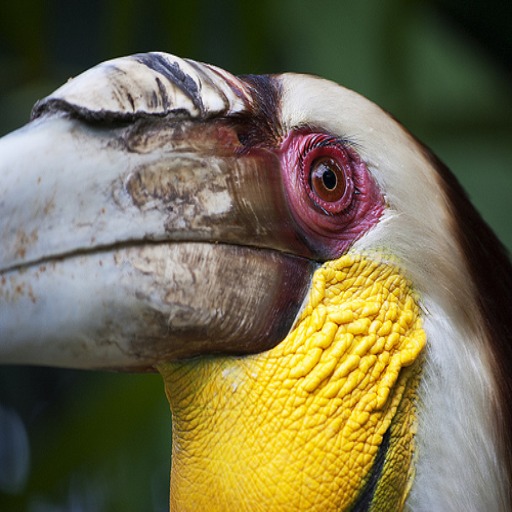}    
    \subfigure[Input]{\includegraphics[width=0.158\columnwidth]{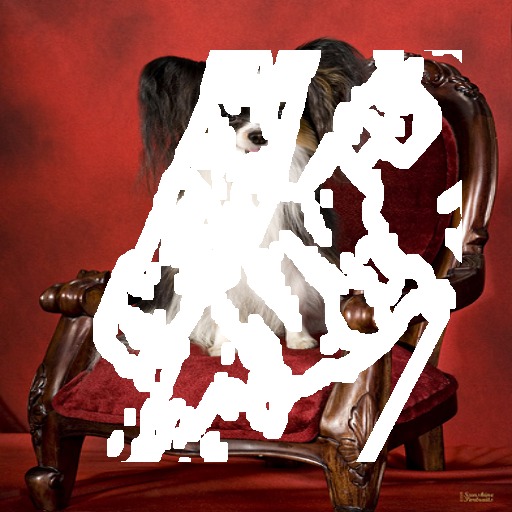}}
    \subfigure[PM]{\includegraphics[width=0.158\columnwidth]{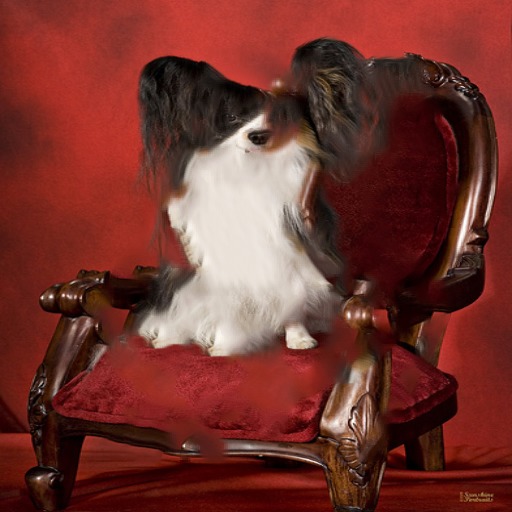}}
    \subfigure[GL]{\includegraphics[width=0.158\columnwidth]{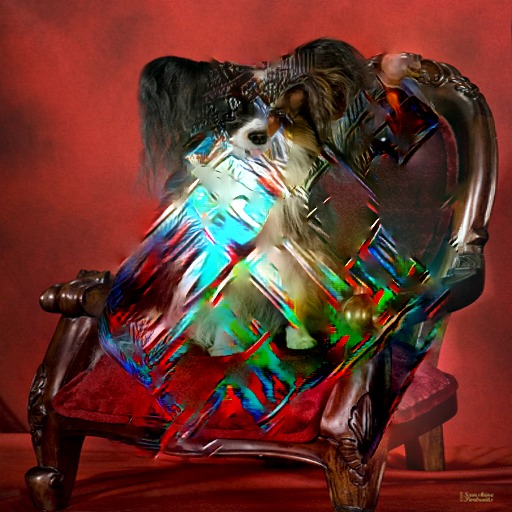}}
    \subfigure[GntIpt]{\includegraphics[width=0.158\columnwidth]{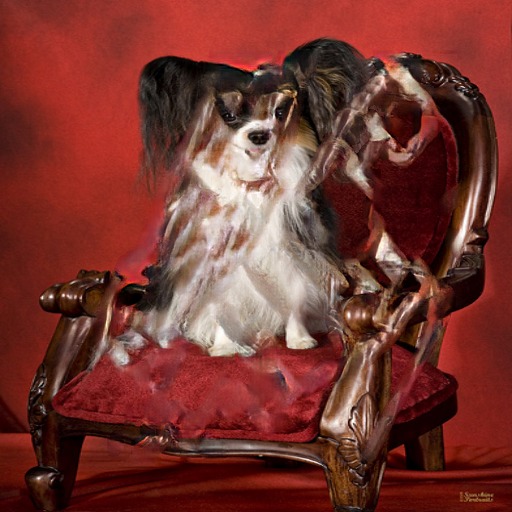}}
    \subfigure[PConv]{\includegraphics[width=0.158\columnwidth]{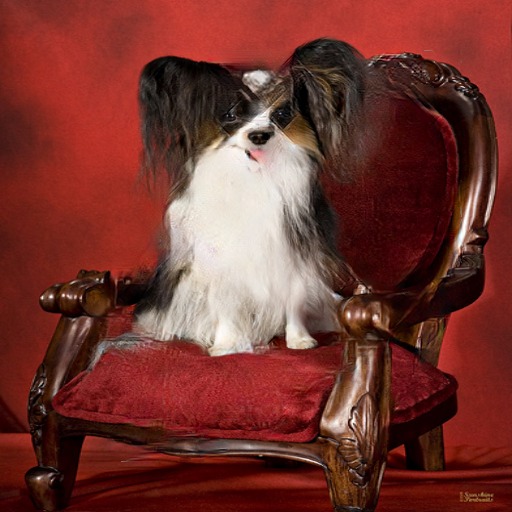}}
    \subfigure[GT]{\includegraphics[width=0.158\columnwidth]{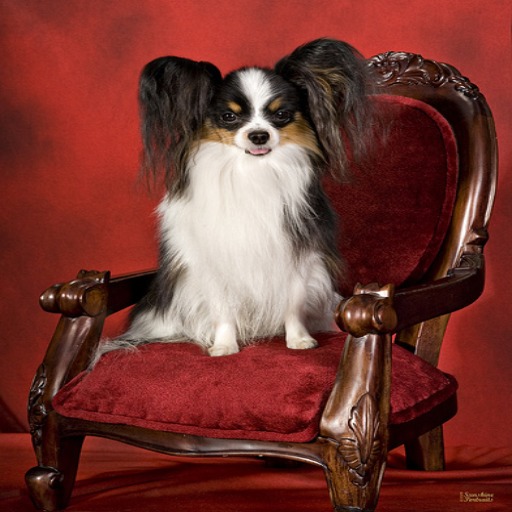}} \\  
    \caption{Comparisons of test results on ImageNet}
    \label{fig:compare_imgnet}
\end{figure}

We compare with 4 methods: 
\begin{itemize}
\item \textbf{PM}: PatchMatch \cite{barnes2009patchmatch}, the state-of-the-art non-learning based approach
\item \textbf{GL}: Method proposed by Iizuka et al. \cite{iizuka2017globally}
\item \textbf{GntIpt}: Method proposed by Yu et al. \cite{yu2018generative}
\item \textbf{Conv}: Same network structure as our method but using typical convolutional layers. Loss weights were re-determined via hyperparameter search.
\end{itemize}

Our method is denoted as \textbf{PConv}. A fair comparison with GL and GntIpt would require retraining their models on our data. However, the training of both approaches use local discriminators assuming availability of the local bounding boxes of the holes, which would not make sense for the shape of our masks. As such, we directly use their released pre-trained models\footnote{\texttt{https://github.com/satoshiiizuka/siggraph2017\_inpainting}, \texttt{https://github.com/JiahuiYu/generative\_inpainting}}. For PatchMatch, we used a third-party implementation\footnote{\texttt{https://github.com/younesse-cv/patchmatch}}. As we do not know their train-test splits, our own splits will likely differ from theirs. We evaluate on 12,000 images randomly assigning our masks to images without replacement.

\begin{figure}[t]%[t]
    \centering
    \includegraphics[width=0.158\columnwidth]{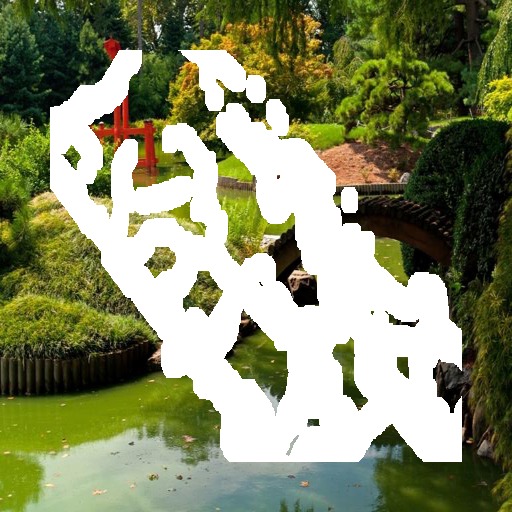}
    \includegraphics[width=0.158\columnwidth]{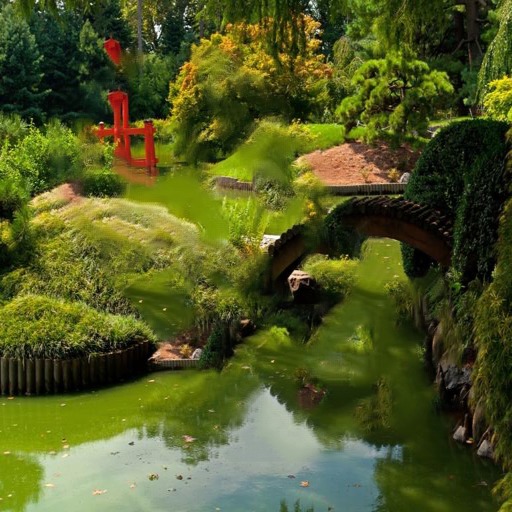}
    \includegraphics[width=0.158\columnwidth]{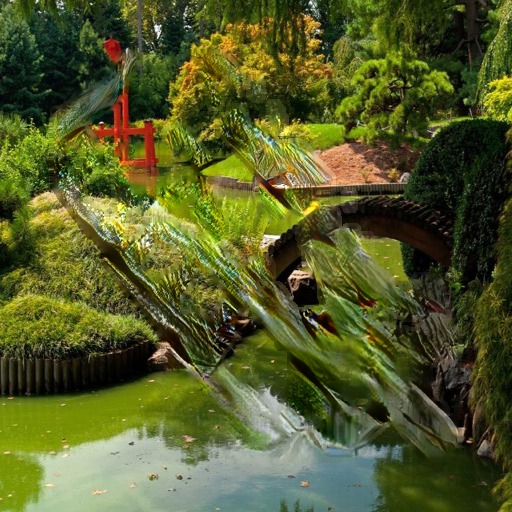}
    \includegraphics[width=0.158\columnwidth]{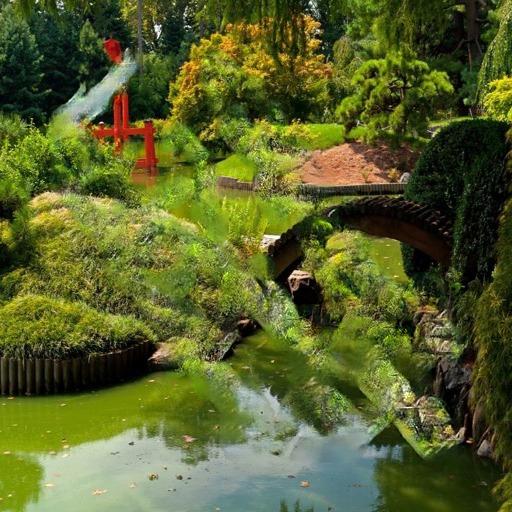}
    \includegraphics[width=0.158\columnwidth]{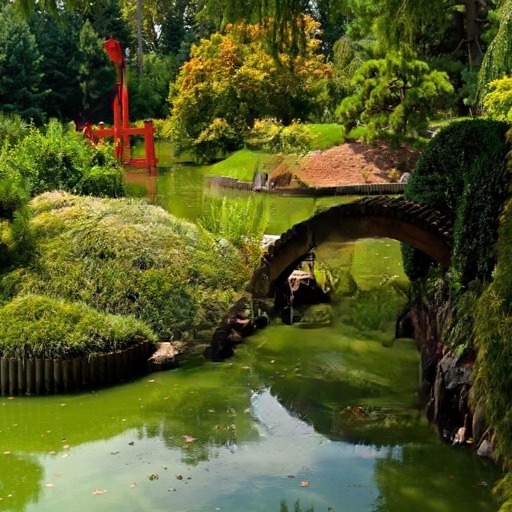}
    \includegraphics[width=0.158\columnwidth]{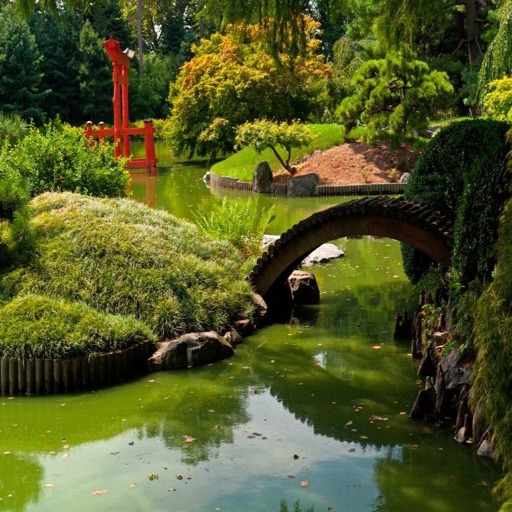} \\
    \includegraphics[width=0.158\columnwidth]{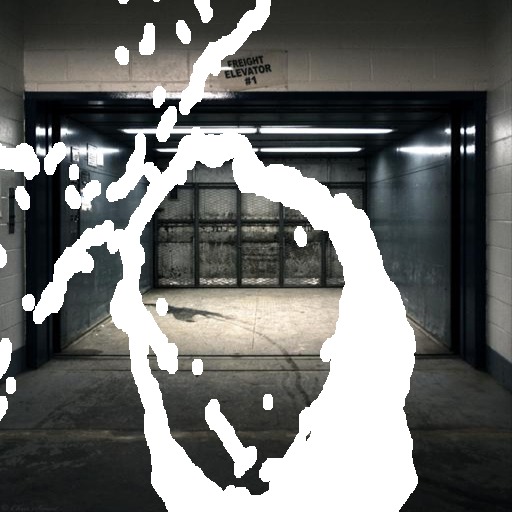}
    \includegraphics[width=0.158\columnwidth]{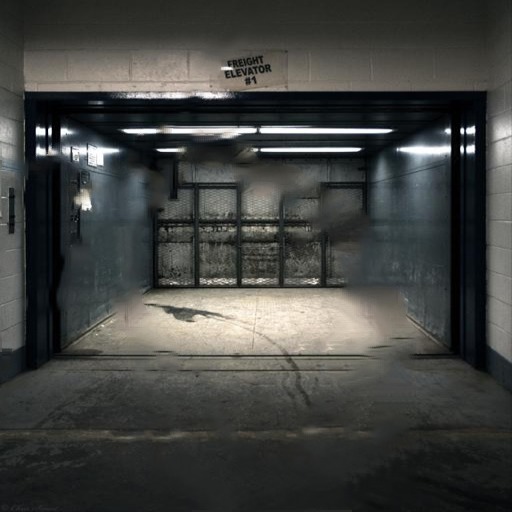}
    \includegraphics[width=0.158\columnwidth]{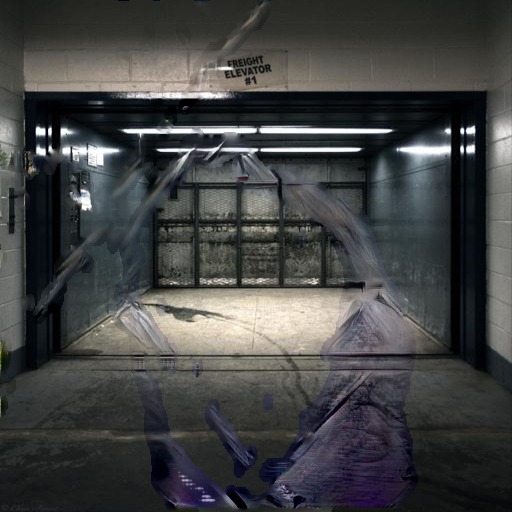}
    \includegraphics[width=0.158\columnwidth]{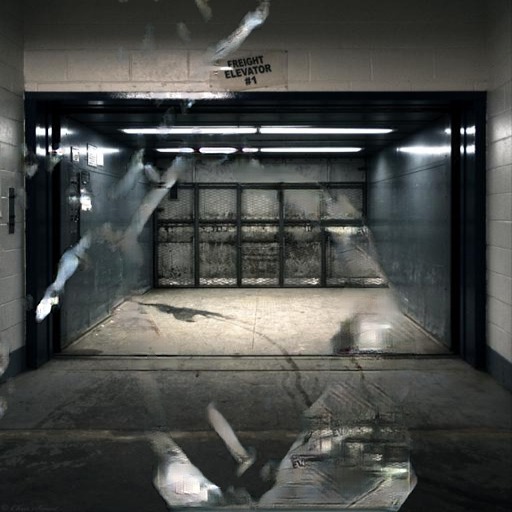}
    \includegraphics[width=0.158\columnwidth]{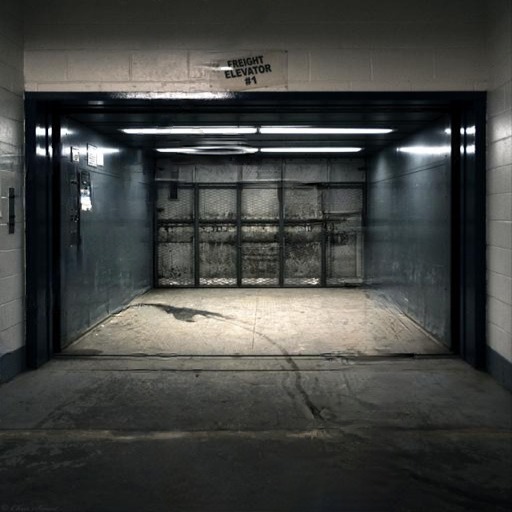}
    \includegraphics[width=0.158\columnwidth]{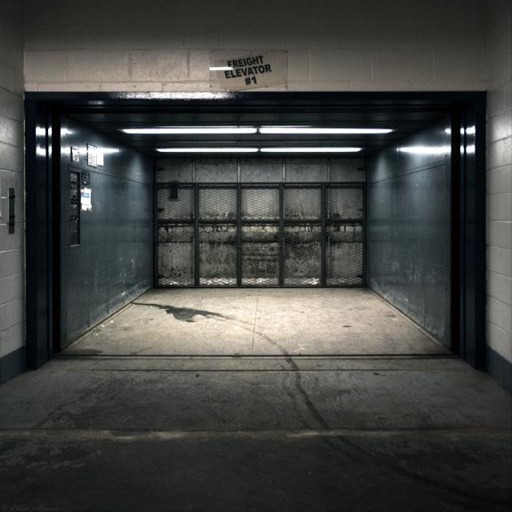} \\          
    \includegraphics[width=0.158\columnwidth]{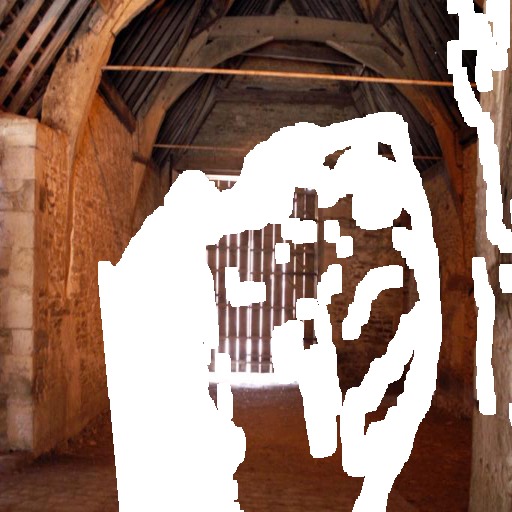}
    \includegraphics[width=0.158\columnwidth]{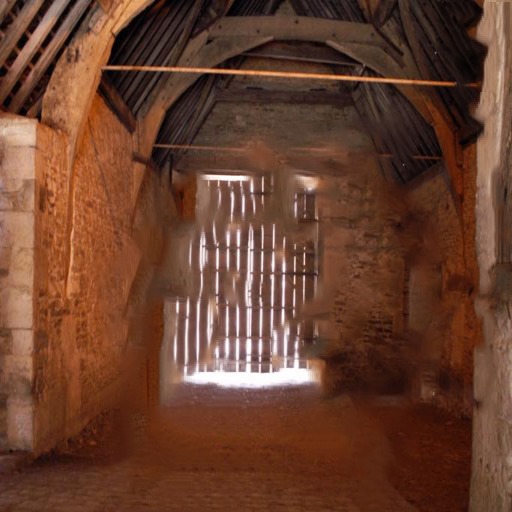}
    \includegraphics[width=0.158\columnwidth]{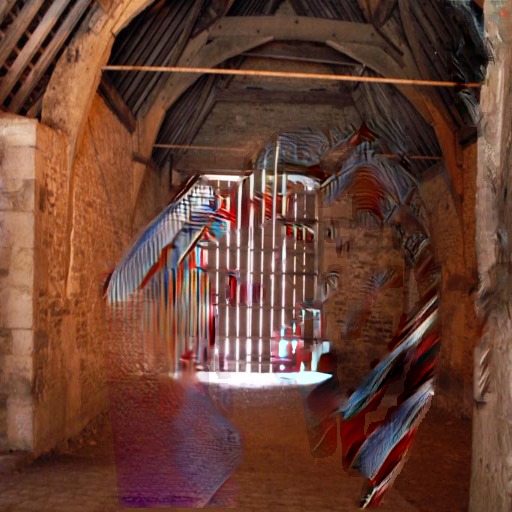}
    \includegraphics[width=0.158\columnwidth]{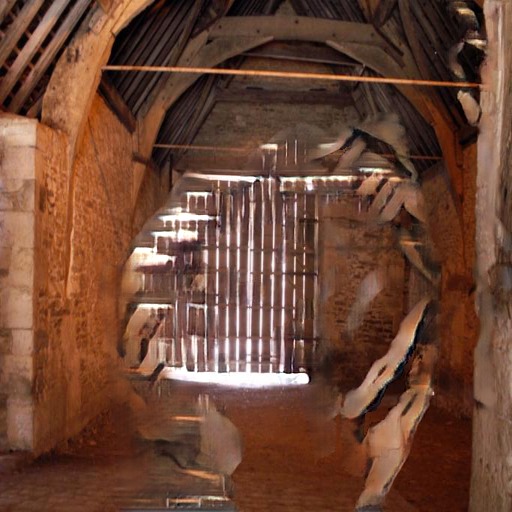}
    \includegraphics[width=0.158\columnwidth]{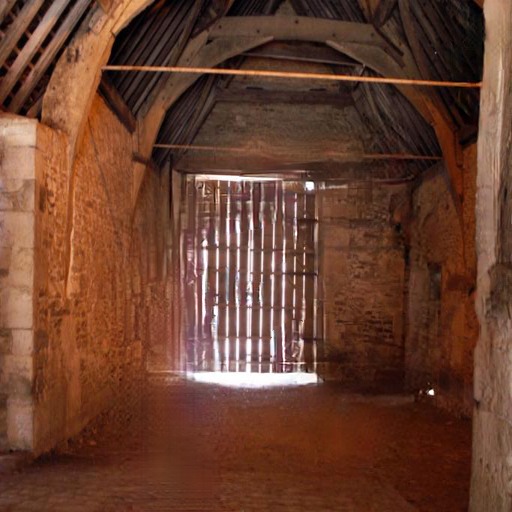}
    \includegraphics[width=0.158\columnwidth]{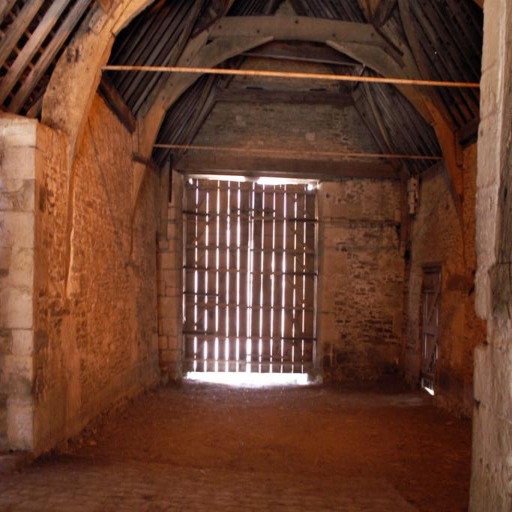}
    \subfigure[Input]{\includegraphics[width=0.158\columnwidth]{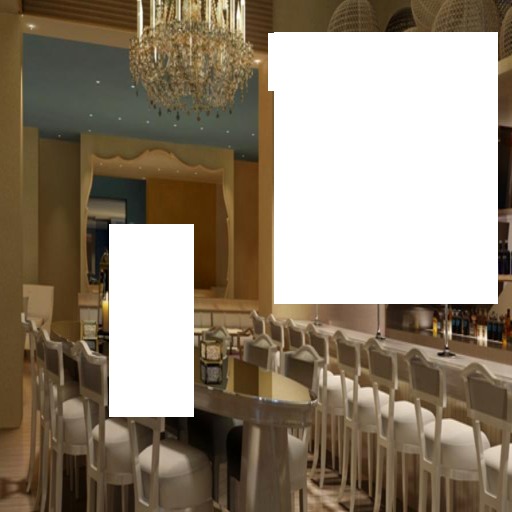}}
    \subfigure[PM]{\includegraphics[width=0.158\columnwidth]{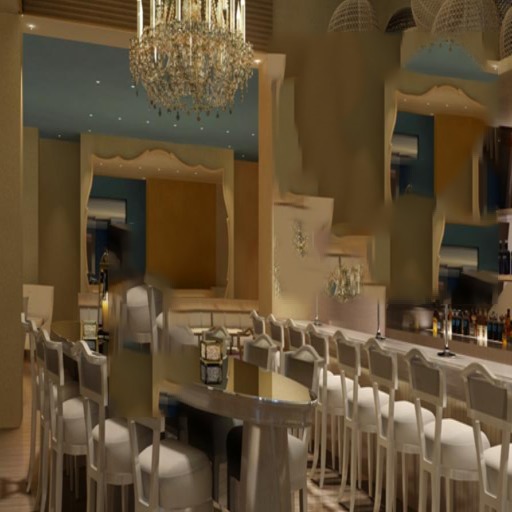}}
    \subfigure[GL]{\includegraphics[width=0.158\columnwidth]{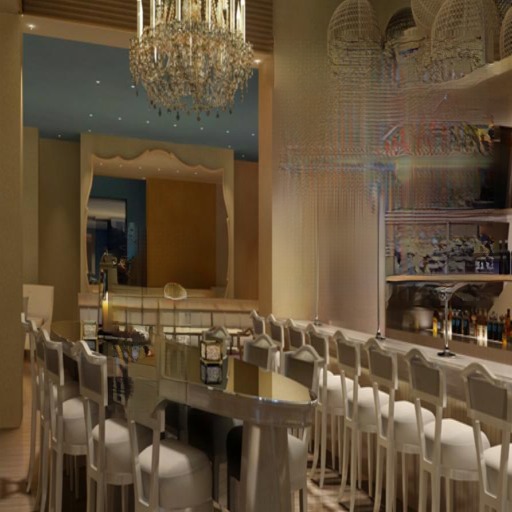}}
    \subfigure[GntIpt]{\includegraphics[width=0.158\columnwidth]{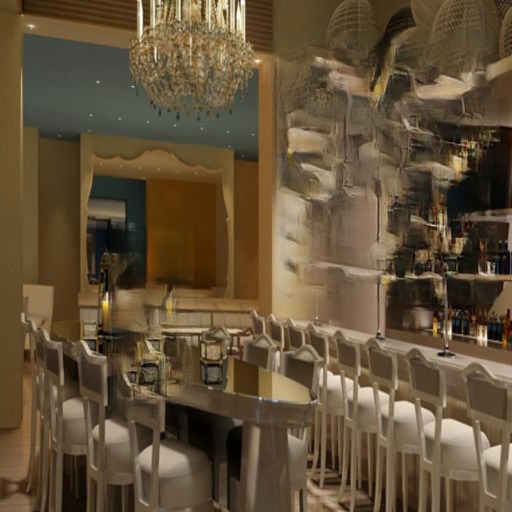}}
    \subfigure[PConv]{\includegraphics[width=0.158\columnwidth]{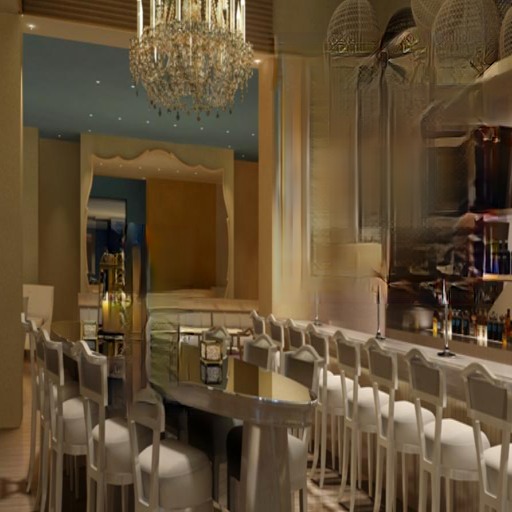}}
    \subfigure[GT]{\includegraphics[width=0.158\columnwidth]{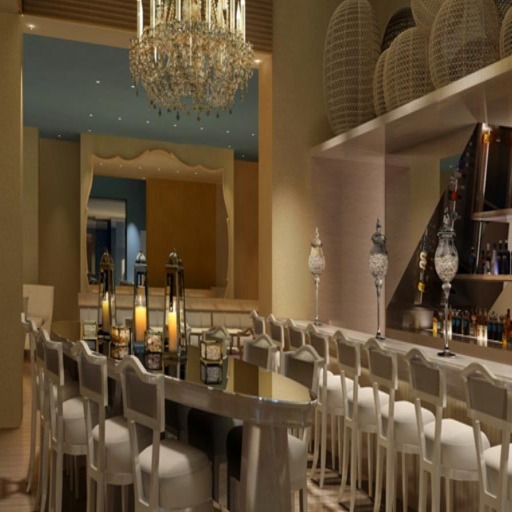}} \\    
    \caption{Comparison of test results on Places2 images}
    \label{fig:compare_place2}
\end{figure}

\begin{figure}[b] %[h]
\vspace{-.3cm}
\centering
\begin{tabular}{ccc|ccc}
\multicolumn{6}{c}{}\\
Input & Conv & PConv & Input & Conv & PConv \\
    \includegraphics[width=0.158\columnwidth]{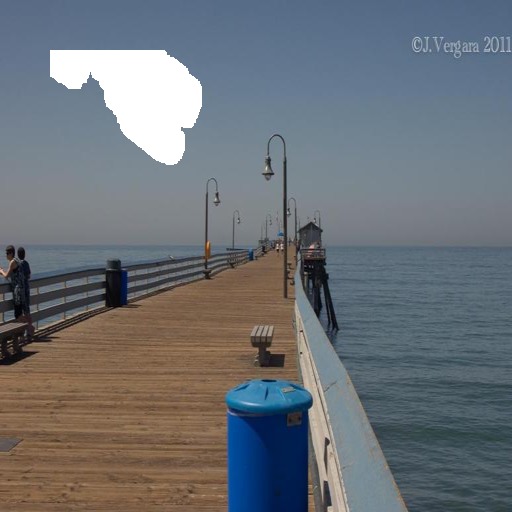} &
    \includegraphics[width=0.158\columnwidth]{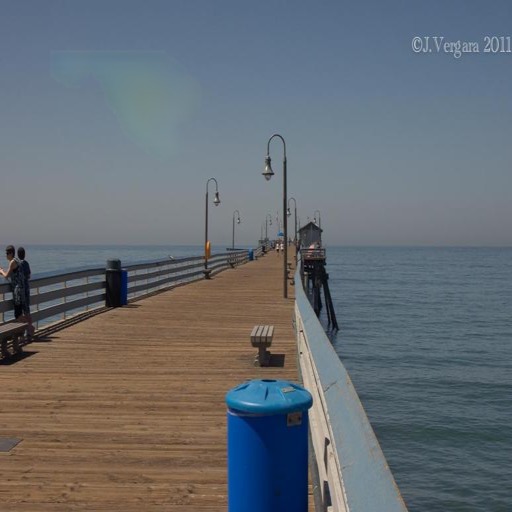} &
    \includegraphics[width=0.158\columnwidth]{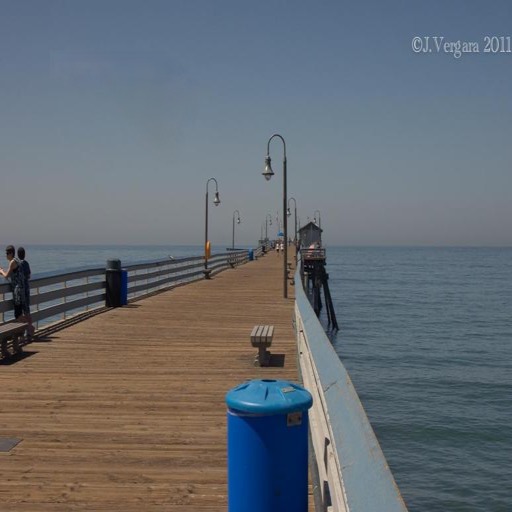} &
    \includegraphics[width=0.158\columnwidth]{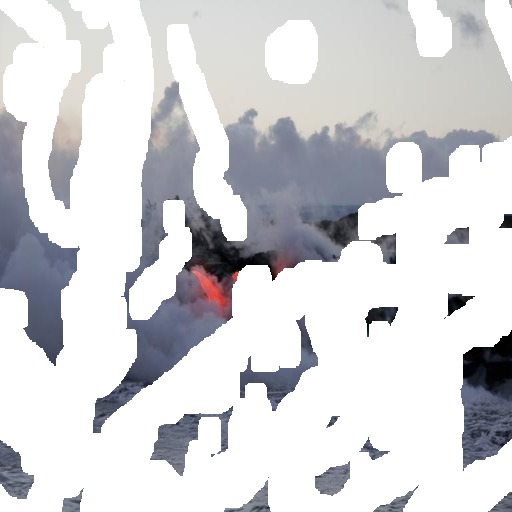} &
    \includegraphics[width=0.158\columnwidth]{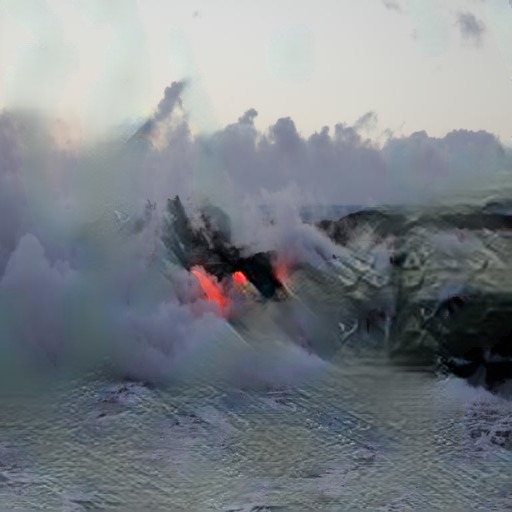} &
    \includegraphics[width=0.158\columnwidth]{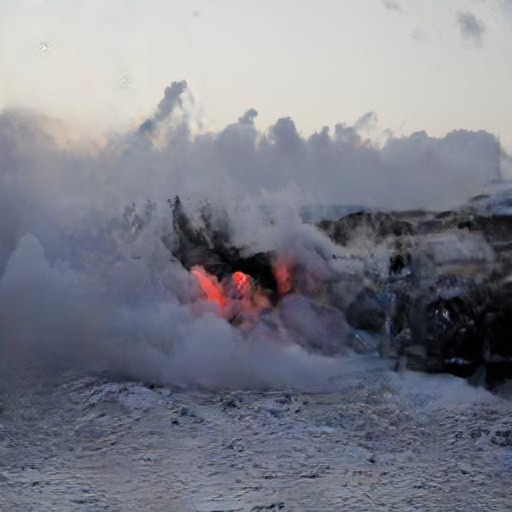} \\
\end{tabular}
\caption{Comparison between typical convolution layer based results (Conv) and partial convolution layer based results (PConv).}
\label{fig:compare_typicalconv}
\end{figure}

\begin{figure}[h] %[t]%[b!] %[ht!]
\centering
\includegraphics[width=0.23\columnwidth]{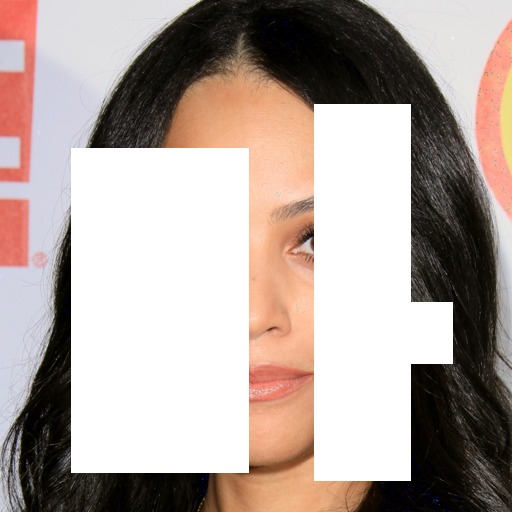}
\includegraphics[width=0.23\columnwidth]{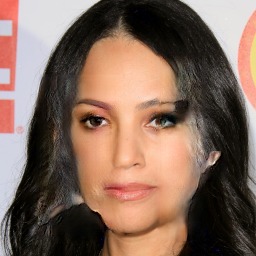}
\includegraphics[width=0.23\columnwidth]{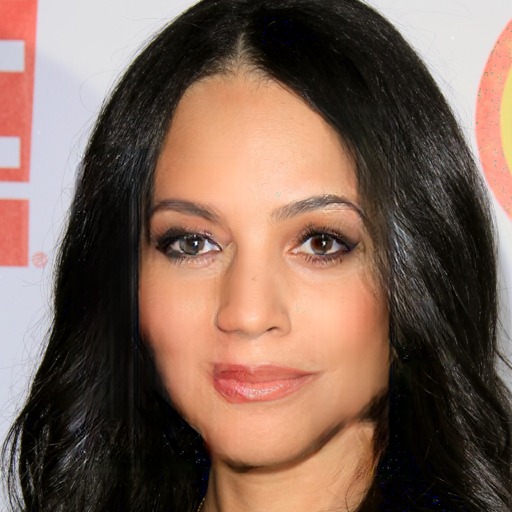}
\includegraphics[width=0.23\columnwidth]{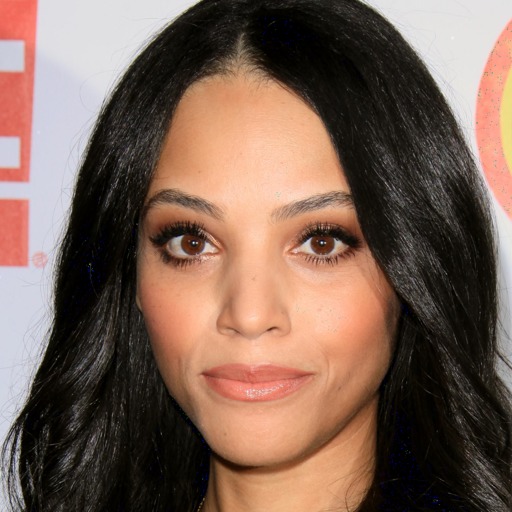} \\
\includegraphics[width=0.23\columnwidth]{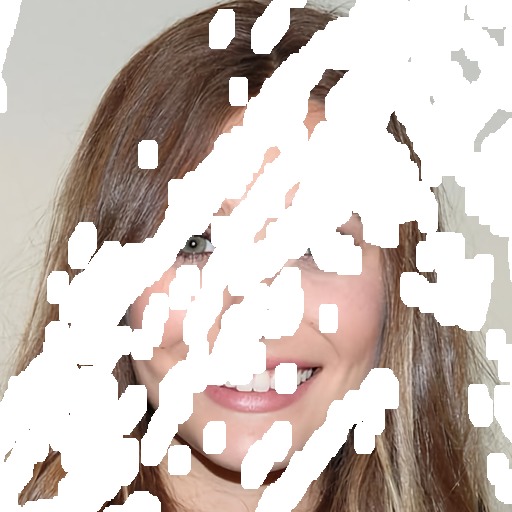}
\includegraphics[width=0.23\columnwidth]{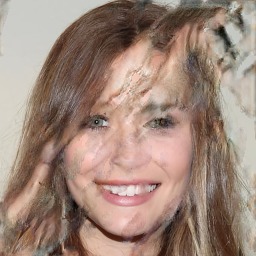}
\includegraphics[width=0.23\columnwidth]{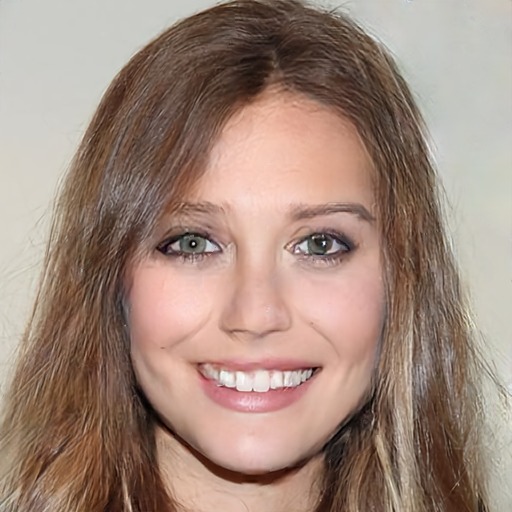}
\includegraphics[width=0.23\columnwidth]{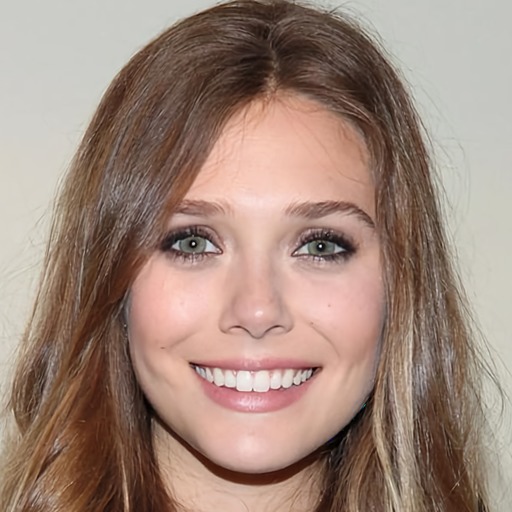}
\subfigure[Input]{\includegraphics[width=0.23\columnwidth]{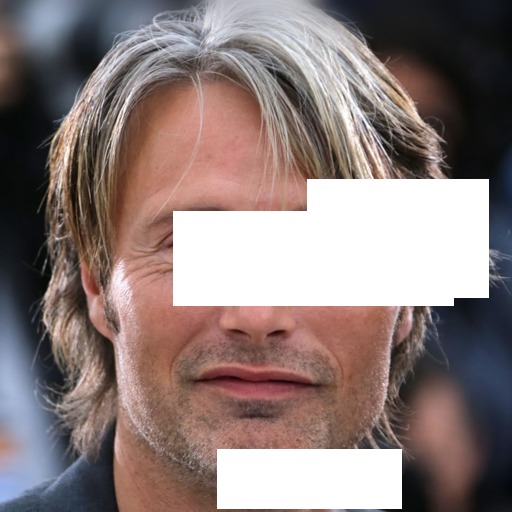}}
\subfigure[GntIpt]{\includegraphics[width=0.23\columnwidth]{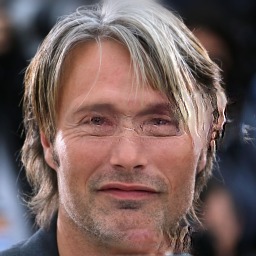}}
\subfigure[PConv(Ours)]{\includegraphics[width=0.23\columnwidth]{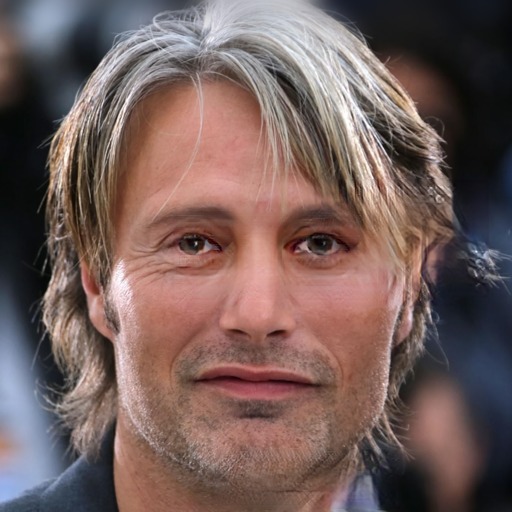}}
\subfigure[Ground Truth]{\includegraphics[width=0.23\columnwidth]{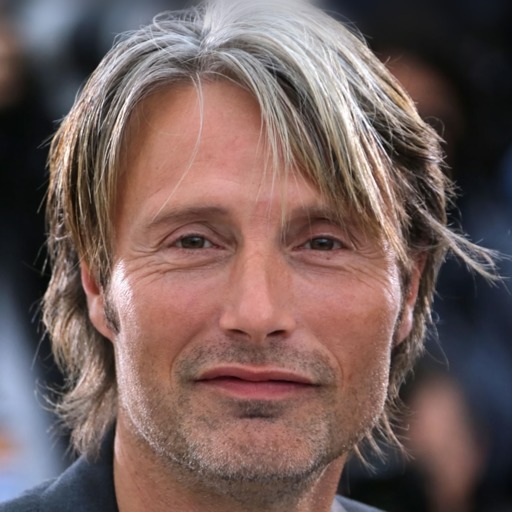}} \\
\caption{Test results on CelebA-HQ.}
\label{fig:compare_face}
\end{figure}

\textbf{Qualitative Comparisons}. Figure~\ref{fig:compare_imgnet} and Figure~\ref{fig:compare_place2} shows the comparisons on ImageNet and Places2 respectively. GT represents the ground truth. We compare with GntIpt\cite{yu2018generative} on CelebA-HQ in Figure~\ref{fig:compare_face}. GntIpt tested CelebA-HQ on 256$\times$256 so we downsample the images to be 256$\times$256 before feeding into their model. It can be seen that PM may copy semantically incorrect patches to fill holes, while GL and GntIpt sometimes fail to achieve plausible results through post-processing or refinement network.  Figure~\ref{fig:compare_typicalconv} shows the results of Conv, which are with the distinct artifacts from conditioning on hole placeholder values.

\textbf{Quantitative comparisons}. 
As mentioned in \cite{yu2018generative}, there is no good numerical metric to evaluate image inpainting results due to the existence of many possible solutions. Nevertheless we follow the previous image inpainting works \cite{yang2017high,yu2018generative} by reporting  $\ell_1$ error, PSNR, SSIM \cite{wang2004image}, and the inception score \cite{salimans2016improved}. $\ell_1$ error, PSNR and SSIM are reported on Places2, whereas the Inception score (IScore) is reported on ImageNet. Note that the released model for \cite{iizuka2017globally} was trained only on Places2, which we use for all evaluations. Table~\ref{tab:evaliregular} shows the comparison results. It can be seen that our method outperforms all the other methods on these measurements on irregular masks.

\begin{table}[t] %[b!]
\centering
\scalebox{0.8}{
{
\begin{tabular}{l|c|c||c|c||c|c||c|c||c|c||c|c}
\multicolumn{1}{c|}{} & \multicolumn{2}{c||}{\textbf{[0.01,0.1]}} & \multicolumn{2}{c||}{\textbf{(0.1,0.2]}} & \multicolumn{2}{c||}{\textbf{(0.2,0.3]}} & \multicolumn{2}{c||}{\textbf{(0.3,0.4]}} & \multicolumn{2}{c||}{\textbf{(0.4,0.5]}} & \multicolumn{2}{c}{\textbf{(0.5,0.6]}} \\
\hline
        & N & B & N & B & N & B & N & B & N & B & N & B \\
\hline
%GT & 1 & 0 & 1 \\
$\ell_1$(PM)($\%$) & \textbf{0.45} & \textbf{0.42} & 1.25 & 1.16 & 2.28 & 2.07 & 3.52 & 3.17 & 4.77 & 4.27 & 6.98 & 6.34 \\
% L1(ContEnc) & & &  & & &  & & &  & & \\
$\ell_1$(GL)($\%$) & 1.39 & 1.53 & 3.01 & 3.22 & 4.51 & 5.00 & 6.05 & 6.77 & 7.34 & 8.20 & 8.60 & 9.78 \\
$\ell_1$(GnIpt)($\%$) & 0.78 & 0.88 & 1.98 & 2.09 & 3.34 & 3.72 & 4.98 & 5.50 & 6.51 & 7.13 & 8.33 & 9.19 \\
%$\ell_1$(Conv-S)($\%$) & 0.59 & 0.57 & 1.43 & 1.35 & 2.53 & 2.36 & 3.86 & 3.56 & 5.30 & 4.85 & 7.82 & 7.15 \\
$\ell_1$(Conv)($\%$) & 0.52 & 0.50 & 1.26 & 1.17 & 2.20 & 2.01 & 3.37 & 3.03 & 4.58 & 4.10 & 6.66 & 6.01 \\
$\ell_1$(PConv)($\%$) & 0.49 & 0.47 & \textbf{1.18} & \textbf{1.09} & \textbf{2.07} & \textbf{1.88} & \textbf{3.19} & \textbf{2.84} & \textbf{4.37} & \textbf{3.85} & \textbf{6.45} & \textbf{5.72}\\
\hline
\hline
PSNR(PM) & 32.97 & 33.68 & 26.87 & 27.51 & 23.70 & 24.35 & 21.27 & 22.05 & 19.70 & 20.58 & 17.60 & 18.22 \\
% PSNR(ContEnc) & & &  & & &  & & &  & & \\
PSNR(GL) & 30.17 & 29.74 & 23.87 & 23.83 & 20.92 & 20.73 & 18.80 & 18.61 & 17.60 & 17.38 & 16.90 & 16.37\\
PSNR(GnIpt) & 29.07 & 28.38 & 23.20 & 22.86 & 20.58 & 19.86 & 18.53 & 17.85 & 17.31 & 16.68 & 16.24 & 15.52\\
%PSNR(Conv-S) & 32.22 & 32.60 & 26.45 & 26.95 & 23.33 & 23.91 & 20.94 & 21.66 & 19.28 & 20.05 & 17.14 & 17.81 \\
PSNR(Conv) & 33.21 & 33.79 & 27.30 & 27.89 & 24.23 & 24.90 & 21.79 & 22.60 & 20.20 & 21.13 & \textbf{18.24} & 18.94\\
PSNR(PConv) & \textbf{33.75} & \textbf{34.34} & \textbf{27.71} & \textbf{28.32} & \textbf{24.54} & \textbf{25.25} & \textbf{22.01} & \textbf{22.89} & \textbf{20.34} & \textbf{21.38} & 18.21 & \textbf{19.04}\\
\hline
\hline
SSIM(PM) & \textbf{0.946} & \textbf{0.947} & 0.861 & 0.865 & 0.763 & 0.768 & 0.666 & 0.675 & 0.568 & 0.579 & 0.459 & 0.472\\
% SSIM(ContEnc) & & &  & & &  & & &  & & \\
SSIM(GL) & 0.929 & 0.923 & 0.831 & 0.829 & 0.732 & 0.721 & 0.638 & 0.627 & 0.543 & 0.533 & 0.446 & 0.440\\
SSIM(GnIpt) & 0.940 & 0.938 & 0.855 & 0.855 & 0.760 & 0.758 & 0.666 & 0.666 & 0.569 & 0.570 & 0.465 & 0.470\\
%SSIM(Conv-S) & 0.939 & 0.939 & 0.854 & 0.857 & 0.758 & 0.761 & 0.663 & 0.669 & 0.564 & 0.573 & 0.453 & 0.467 \\
SSIM(Conv) & 0.943 & 0.943 & 0.862 & 0.865 & 0.769 & 0.772 & 0.674 & 0.682 & 0.576 & 0.587 & 0.463 & 0.478\\
SSIM(PConv) &\textbf{0.946} & 0.945 & \textbf{0.867} & \textbf{0.870} & \textbf{0.775} & \textbf{0.779} & \textbf{0.681} & \textbf{0.689} & \textbf{0.583} & \textbf{0.595} & \textbf{0.468} & \textbf{0.484}\\
\hline
\hline
IScore(PM) & 0.090 & 0.058 & 0.307 & 0.204 & 0.766 & 0.465 & 1.551 & 0.921 & 2.724 & 1.422 & 4.075 & 2.226 \\
IScore(GL) & 0.183 & 0.112 & 0.619 & 0.464 & 1.607 & 1.046 & 2.774 & 1.941 & 3.920 & 2.825 & 4.877 & 3.362 \\
IScore(GnIpt) & 0.127 & 0.088 & 0.396 & 0.307 & 0.978 & 0.621 & 1.757 & 1.126 & 2.759 & 1.801 & 3.967 & 2.525 \\
%IScore(Conv-S) & 0.109 & 0.071 & 0.345 & 0.236 & 0.982 & 0.562 & 1.927 & 1.088 & 3.352 & 1.841 & 4.682 & 2.788 \\
IScore(Conv) & 0.068 & 0.041 & 0.228 & 0.149 & 0.603 & 0.366 & 1.264 & 0.731 & 2.368 & 1.189 & 4.162 & 2.224 \\
IScore(PConv) & \textbf{0.051} & \textbf{0.032} & \textbf{0.163} & \textbf{0.109} & \textbf{0.446} & \textbf{0.270} & \textbf{0.954} & \textbf{0.565} & \textbf{1.881} & \textbf{0.838} & \textbf{3.603} & \textbf{1.588} \\
\hline
\hline
\end{tabular}
}
}
\caption{Comparisons with various methods. Columns represent different hole-to-image area ratios. N=no border, B=border}
%\vspace{-1cm}
\label{tab:evaliregular}
\vspace{-0.3cm}
\end{table}

\begin{figure}[h!] %[h] %[ht]
    \centering
    \includegraphics[width=0.85\columnwidth]{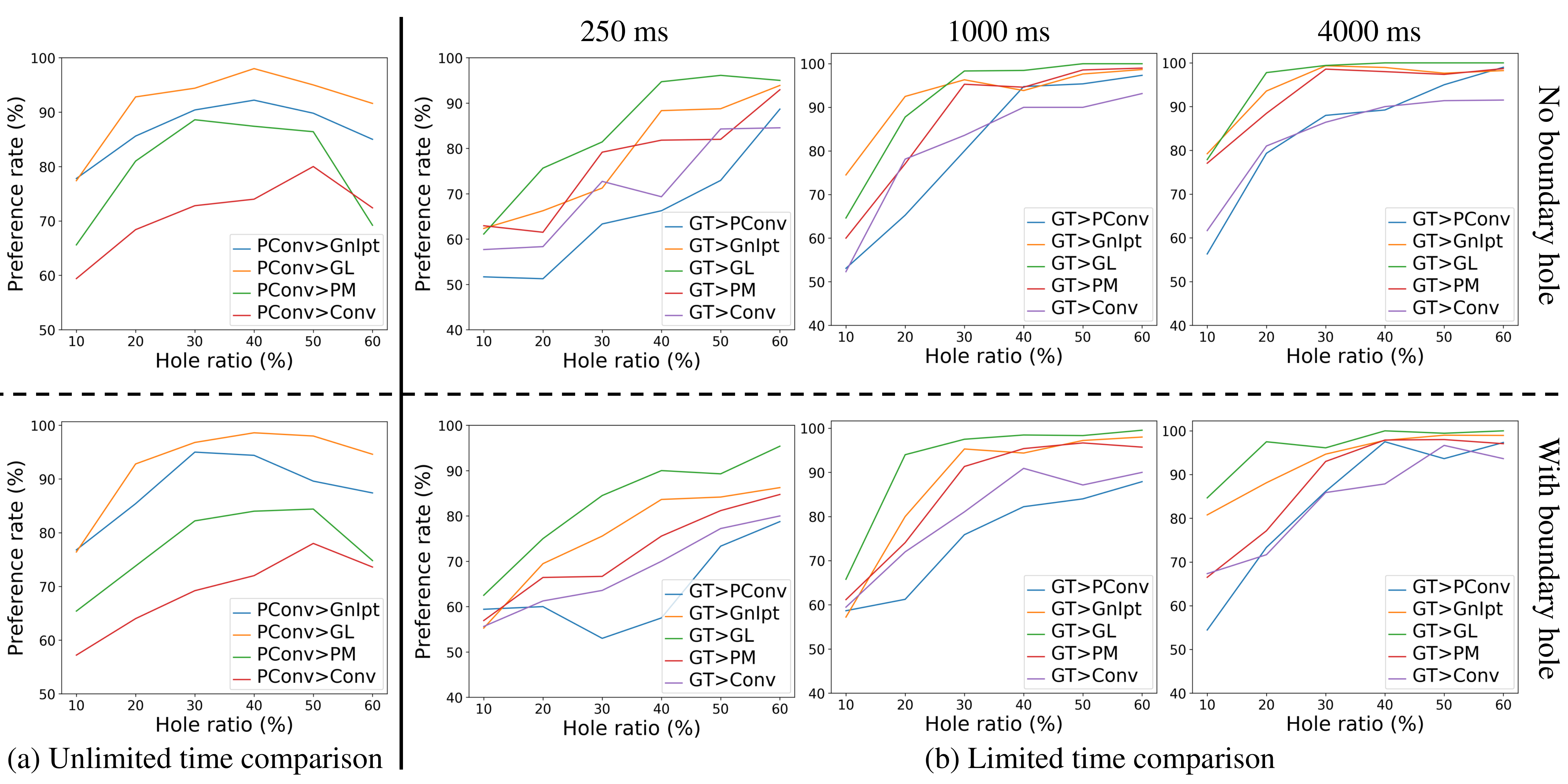}
    \caption{User study results. We perform two kinds of experiments: unlimited time and limited time. (a) In the unlimited time setting, we compare our result with the result generated by another method. The rate where our result is preferred is graphed. 50\% means two methods are equal. In the first row, the holes are not allowed to touch the image boundary, while in the second row it is allowed. (b) In the limited time setting, we compare all methods to the ground truth. The subject is given some limited time (250ms, 1000ms or 4000ms) to select which image is more realistic. The rate where ground truth is preferred over the other method is reported. The lower the curve, the better. }
    \label{fig:mturk}
   \vspace{-0.3cm}
\end{figure}

\textbf{User Study}
In addition to quantitative comparisons, we also evaluate our algorithm via a human subjective study. 
We perform pairwise A/B tests without showing hole positions or original input image with holes, deployed on the Amazon Mechanical Turk (MTurk) platform.
We perform two different kinds of experiments: unlimited time and limited time.
We also report the cases with and without holes close to the image boundaries separately.
For each situation, We randomly select $300$ images for each method, where each image is compared $10$ times.

For the unlimited time setting, the workers are given two images at once: each generated by a different method. 
The workers are then given unlimited time to select which image looks more realistic.
We also shuffle the image order to ensure unbiased comparisons. 
The results across all different hole-to-image area ratios are summarized in Fig.~\ref{fig:mturk}(a).
The first row shows the results where the holes are at least 50 pixels away from the image border, while the second row shows the case where the holes may be close to or touch image border.
As can be seen, our method performs significantly better than all the other methods (50\% means two methods perform equally well) in both cases.

For the limited time setting, we compare all methods (including ours) to the ground truth.
In each comparison, the result of one method is chosen and shown to the workers along with the ground truth for a limited amount of time.
The workers are then asked to select which image looks more natural.
This evaluates how quickly the difference between the images can be perceived.
The comparison results for different time intervals are shown in Fig.~\ref{fig:mturk}(b). 
Again, the first row shows the case where the holes do not touch the image boundary while the second row allows that.
Our method outperforms the other methods in most cases across different time periods and hole-to-image area ratios.

\section{Discussion \& Extension}
\subsection{Discussion}
\begin{figure}%[h] %[t!]
\centering
\begin{tabular}{c|c|c|c|c|c}
    \includegraphics[width=0.158\columnwidth]{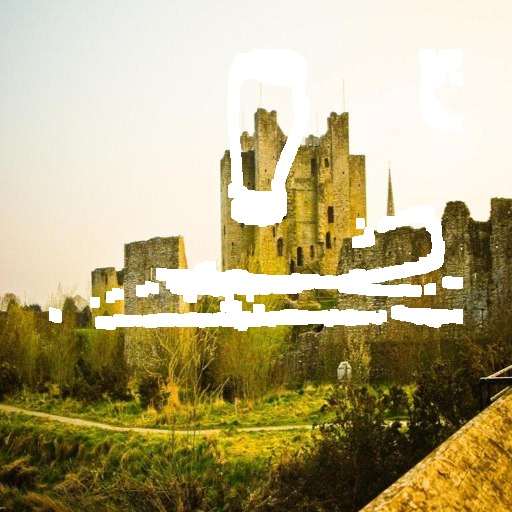} &
    \includegraphics[width=0.158\columnwidth]{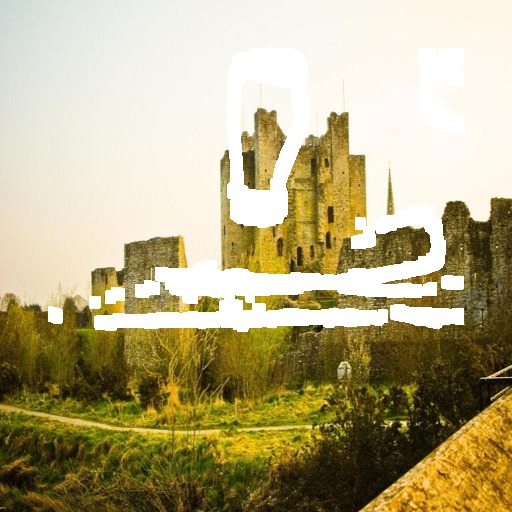} &
    \includegraphics[width=0.158\columnwidth]{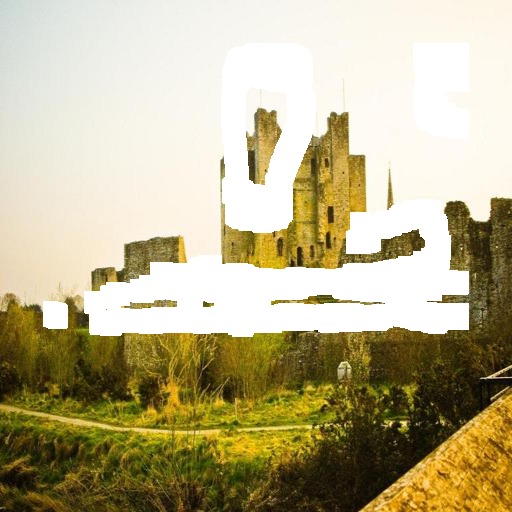} &
    \includegraphics[width=0.158\columnwidth]{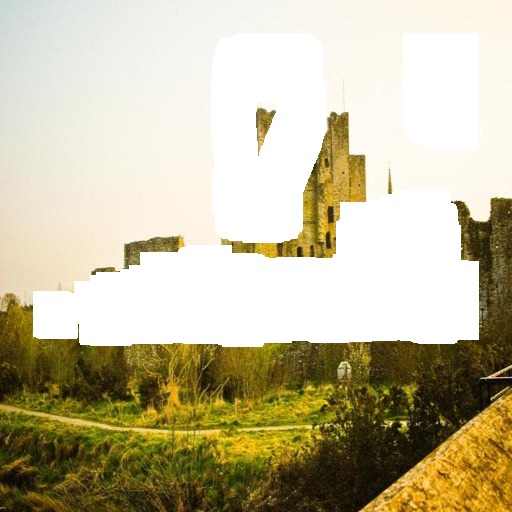} &
    \includegraphics[width=0.158\columnwidth]{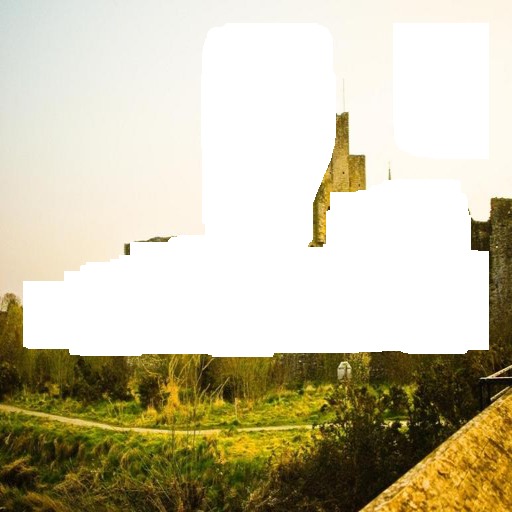} &
    \includegraphics[width=0.158\columnwidth]{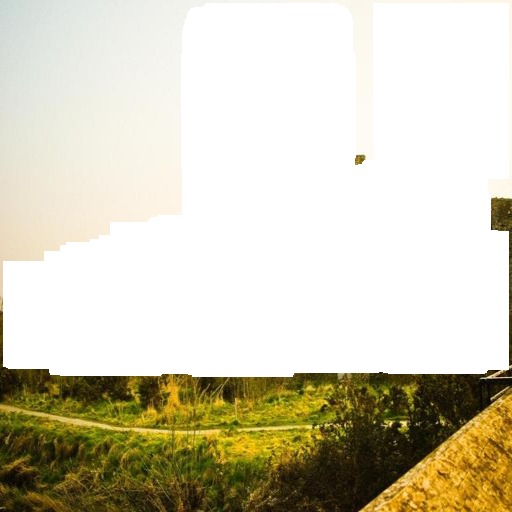} \\    
    \includegraphics[width=0.158\columnwidth]{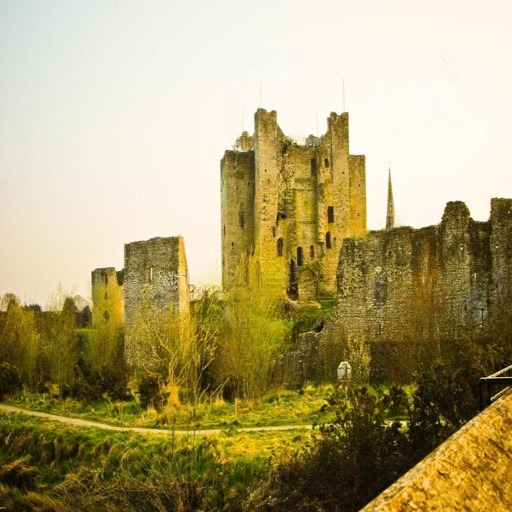} &
    \includegraphics[width=0.158\columnwidth]{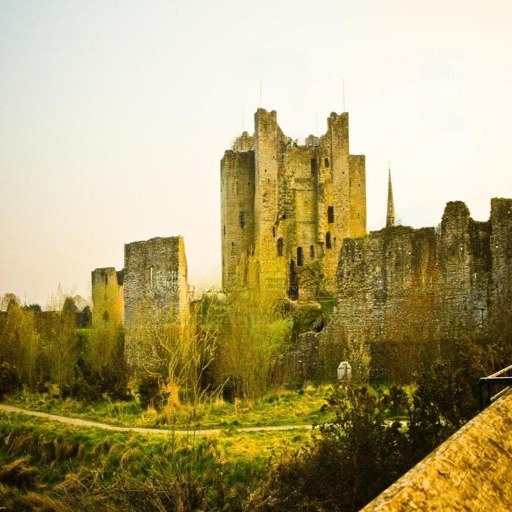} &
    \includegraphics[width=0.158\columnwidth]{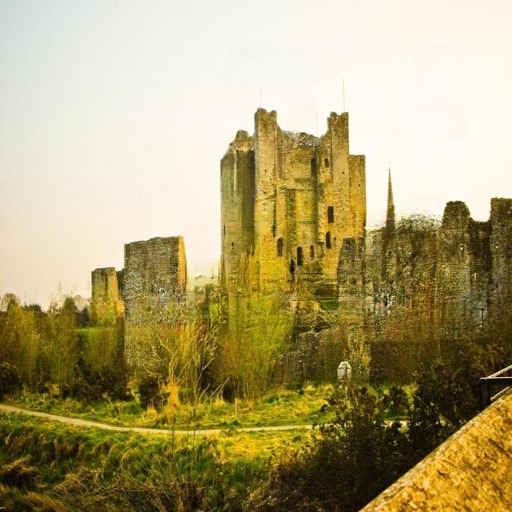} &
    \includegraphics[width=0.158\columnwidth]{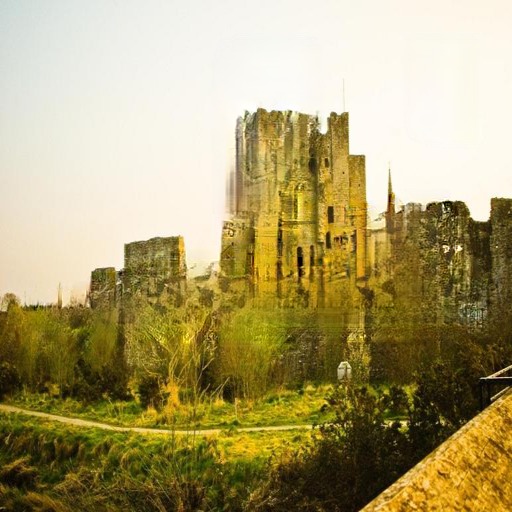} &
    \includegraphics[width=0.158\columnwidth]{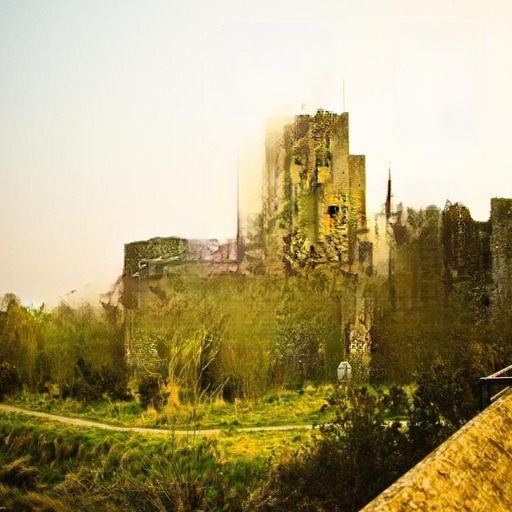} &
    \includegraphics[width=0.158\columnwidth]{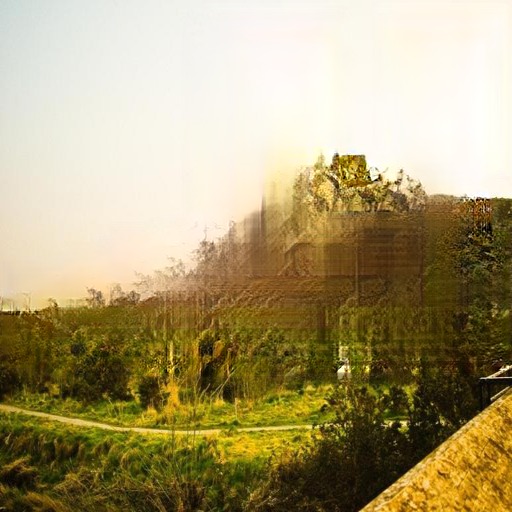} \\
\end{tabular}
\caption{Inpainting results with various dilation of the hole region from left to right: 0, 5, 15, 35, 55, and 95 pixels dilation respectively. Top row: input; bottom row: corresponding inpainted results.}
\label{fig:compare_sensitivity}
\vspace{-1cm}
\end{figure}

\begin{figure} %[h] %[b!]
    \centering
    \includegraphics[width=0.158\columnwidth]{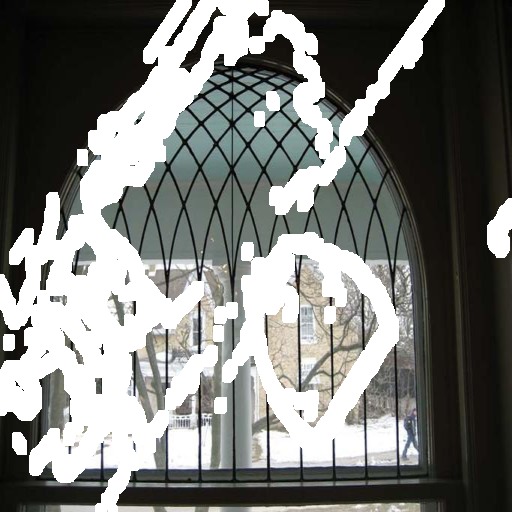}
    \includegraphics[width=0.158\columnwidth]{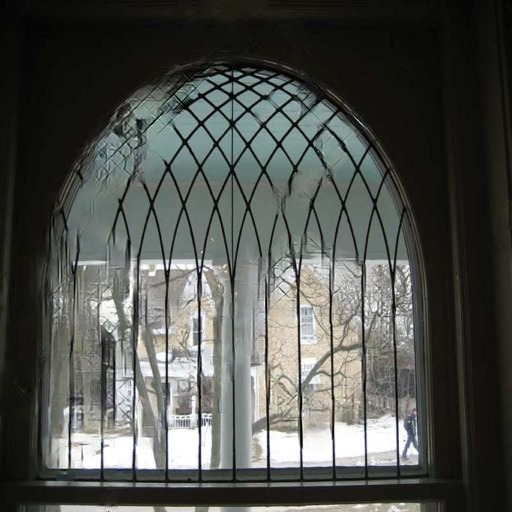}
    \includegraphics[width=0.158\columnwidth]{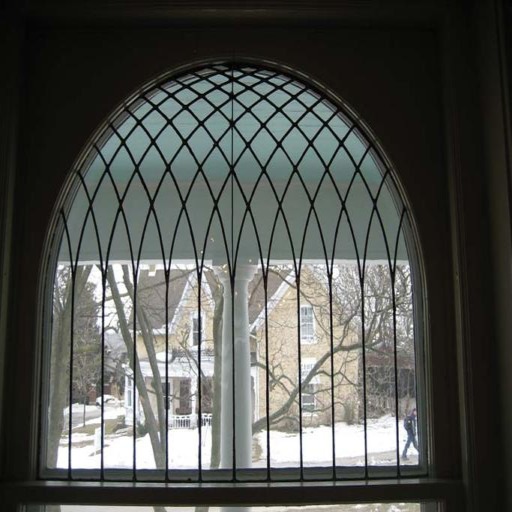}
    \includegraphics[width=0.158\columnwidth]{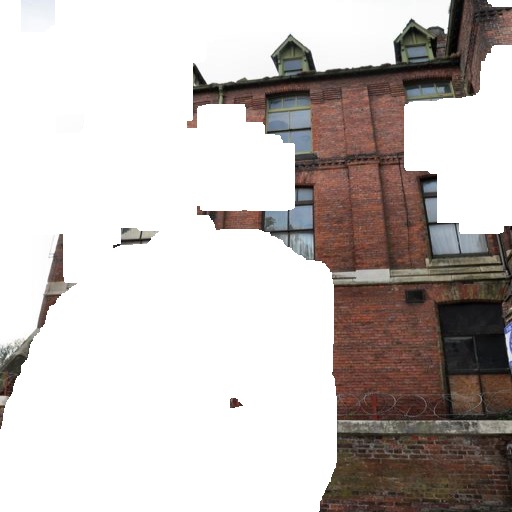}
    \includegraphics[width=0.158\columnwidth]{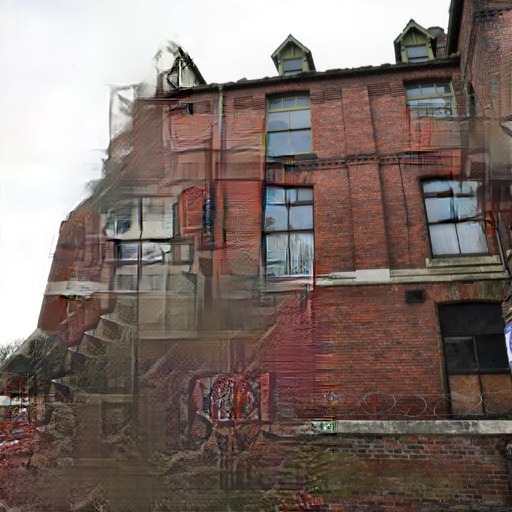}
    \includegraphics[width=0.158\columnwidth]{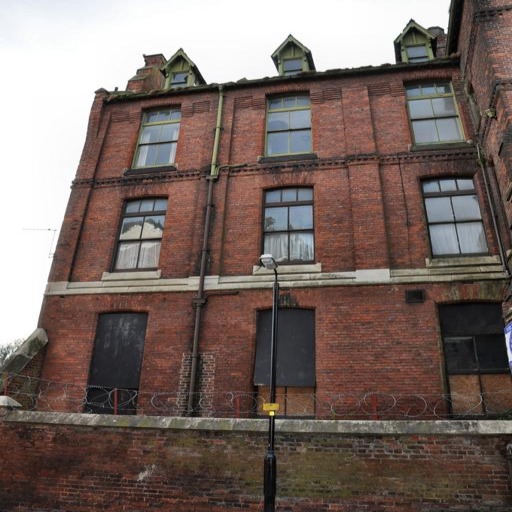} \\    
    \caption{Failure cases. Each group is ordered as input, our result and ground truth.}
    \label{fig:compare_failure}
    \vspace{-0.5cm}
\end{figure}

We propose the use of a partial convolution layer with an automatic mask updating mechanism and achieve state-of-the-art image inpainting results. Our model can robustly handle holes of any shape, size location, or distance from the image borders. Further, our performance does not deteriorate catastrophically as holes increase in size, as seen in Figure~\ref{fig:compare_sensitivity}. However, one limitation of our method is that it fails for some sparsely structured images such as the bars on the door in Figure~\ref{fig:compare_failure}, and, like most methods, struggles on the largest of holes. 

\begin{figure}[t] %[h!]
\centering
   \subfigure[Low Res]{
   \includegraphics[width=0.18\columnwidth]{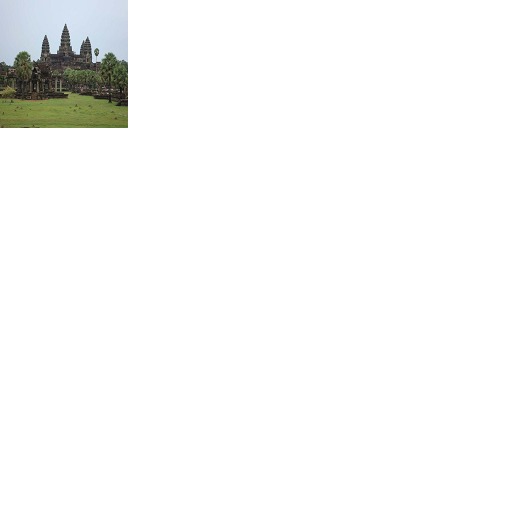}}
   \subfigure[Input]{\includegraphics[width=0.18\columnwidth]{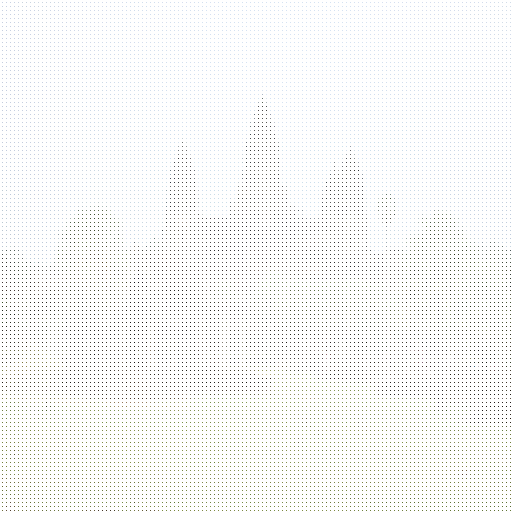}}
   \subfigure[Input Mask]{\includegraphics[width=0.18\columnwidth]{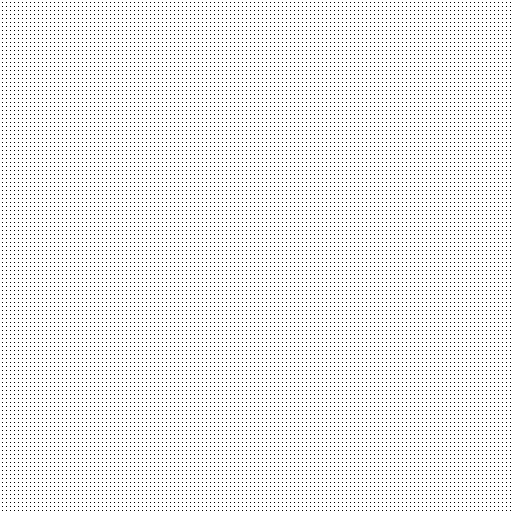}}
   \subfigure[Output]{\includegraphics[width=0.18\columnwidth]{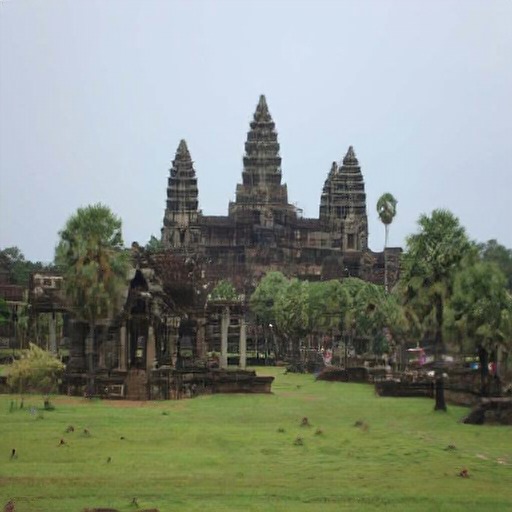}}
    \subfigure[GT]{\includegraphics[width=0.18\columnwidth]{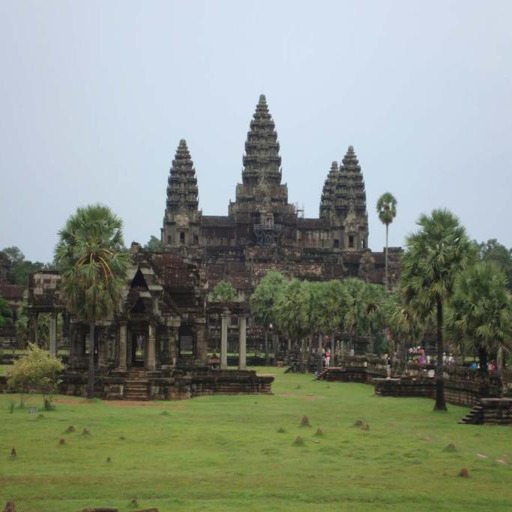}}
    \caption{Example input and output for image super resolution task. Input to the network is constructed from the low resolution image by offsetting pixels and inserting holes using the way described in Section~\ref{sec:superres}.}
\label{fig:superres_input}
\vspace{-0.5cm}
\end{figure}

\begin{figure} %[h] %[h!]
\centering
    \scalebox{0.85}{
    \begin{tabular}{ccccc}
    \multicolumn{5}{c}{}\\
    Bicubic & SRGAN & MDSR+ & PConv & GT \\
    \includegraphics[width=0.19\columnwidth]{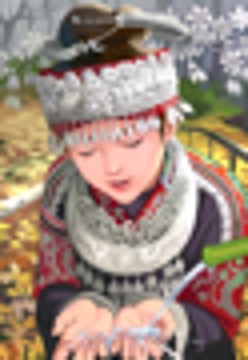} &
    \includegraphics[width=0.19\columnwidth]{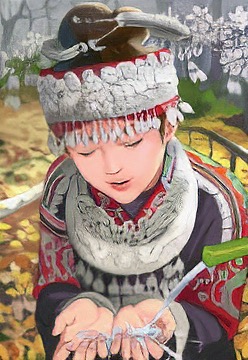} &
    \includegraphics[width=0.19\columnwidth]{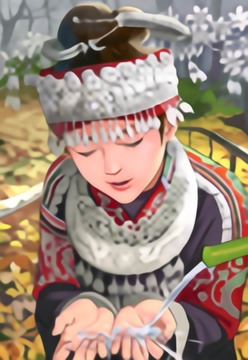} &
    \includegraphics[width=0.19\columnwidth]{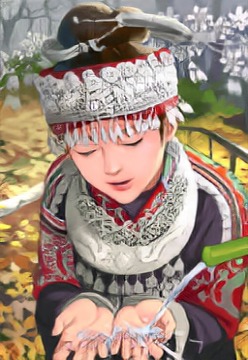} &
    \includegraphics[width=0.19\columnwidth]{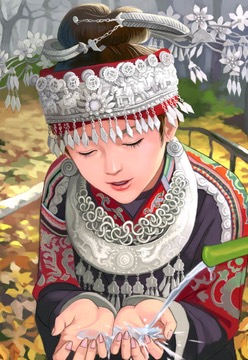} \\    
    \includegraphics[width=0.19\columnwidth]{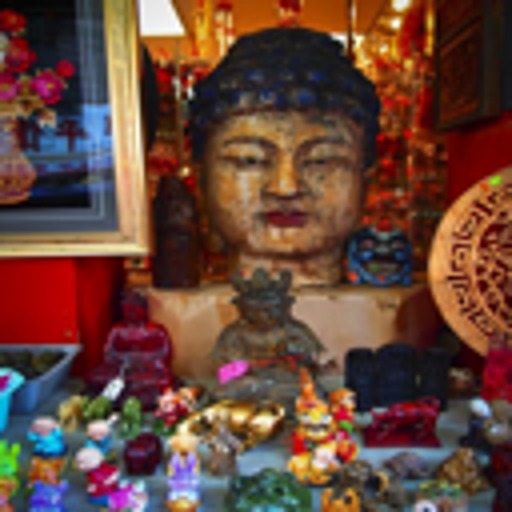} &
    \includegraphics[width=0.19\columnwidth]{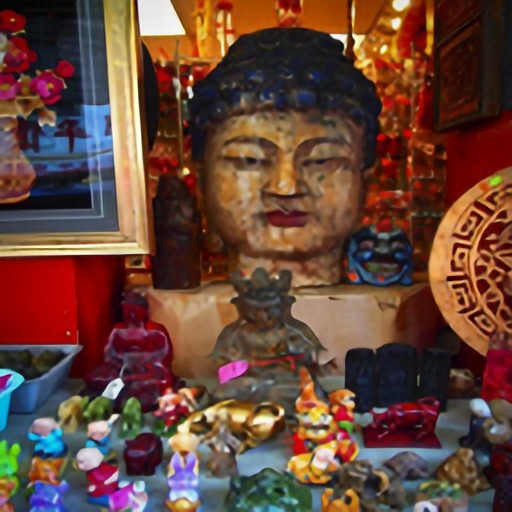} &
    \includegraphics[width=0.19\columnwidth]{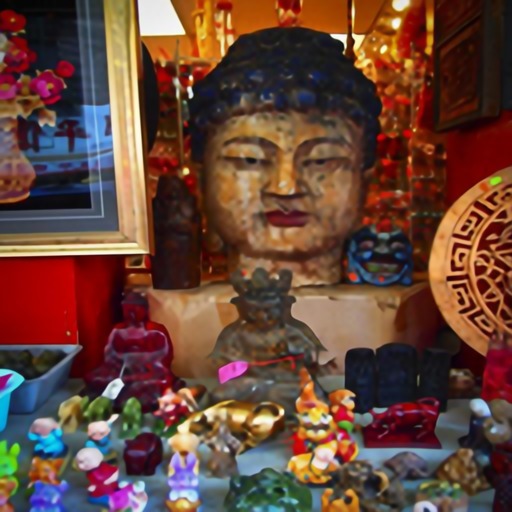} &
    \includegraphics[width=0.19\columnwidth]{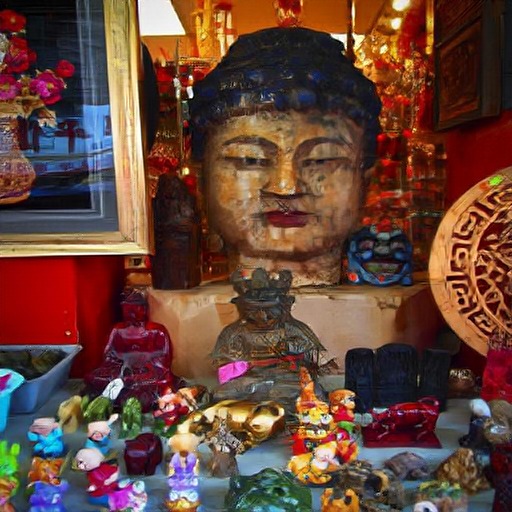} &
    \includegraphics[width=0.19\columnwidth]{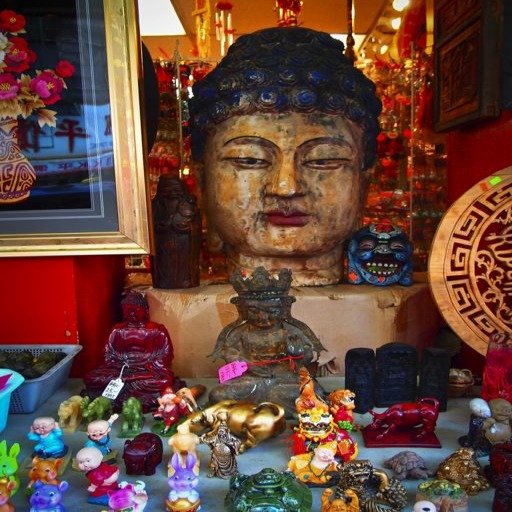} \\      
    \includegraphics[width=0.19\columnwidth]{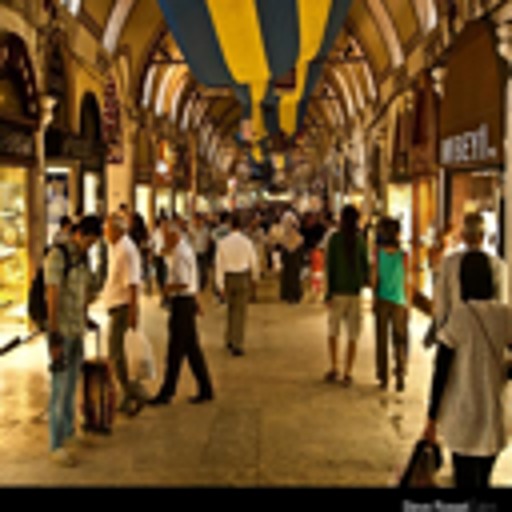} &
    \includegraphics[width=0.19\columnwidth]{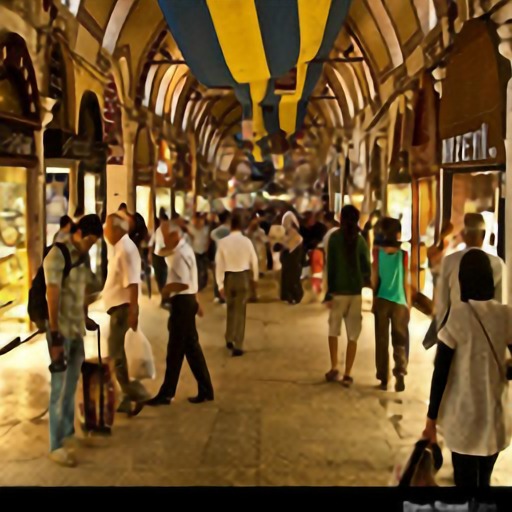} &
    \includegraphics[width=0.19\columnwidth]{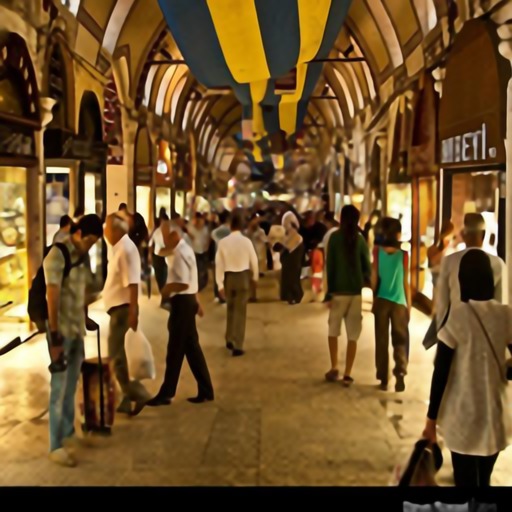} &
    \includegraphics[width=0.19\columnwidth]{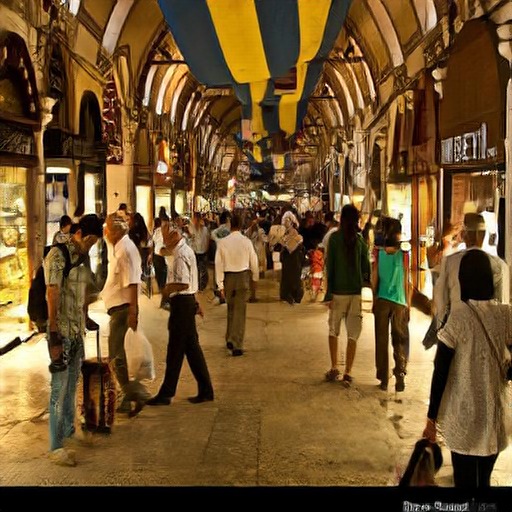} &
    \includegraphics[width=0.19\columnwidth]{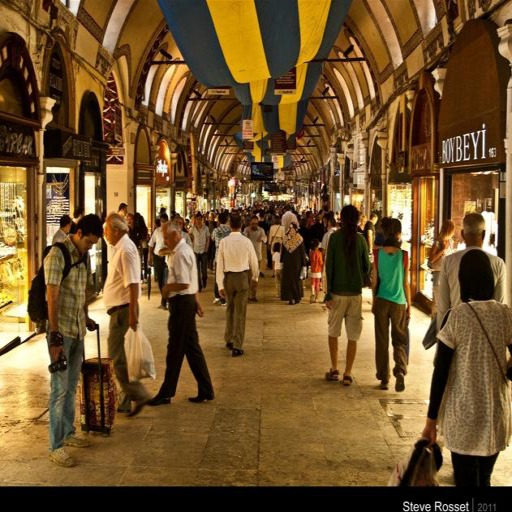} \\    
    \end{tabular}
    }
    \caption{Comparison with SRGAN and MDSR+ for image super resolution task.}
\label{fig:compare_superres_0}
\vspace{-0.3cm}
\end{figure}

\subsection{Extension to Image Super Resolution}
\label{sec:superres}

We also extend our framework to image super resolution tasks by offsetting pixels and inserting holes. Specifically, given a low resolution image $I$ with height $H$ and width $W$ and up-scaling factor $K$, we construct the input $I'$ with height $K$*$H$ and width $K$*$W$ for the network using the following: for each pixel $(x, y)$ in $I$, we put it at $(K$*$x$+$\floor{K/2}$, $K$*$y$+$\floor{K/2})$ in $I'$ and mark this position to have mask value be 1. One example input setting and corresponding output with $K$=4 can be found in Figure~\ref{fig:superres_input}. We compare with two well-known image super-resolution approaches SRGAN\cite{ledig2016photo} and MDSR+\cite{lim2017enhanced} with $K$=4 in Figure~\ref{fig:compare_superres_0}.

\textbf{Acknowledgement}. We would like to thank Jonah Alben, Rafael Valle Costa, Karan Sapra, Chao Yang, Raul Puri, Brandon Rowlett and other NVIDIA colleagues for valuable discussions, and Chris Hebert for technical support.

\clearpage

\bibliographystyle{splncs04}
\bibliography{egbib}

% \clearpage

\section*{Appendix}

\subsection*{Details of Network Architecture}

\begin{table}[ht!]
    \centering
    \scalebox{0.73}{
    \begin{tabular}{c|c|c|c|c|c}
        Module Name& Filter Size & \# Filters/Channels & Stride/Up Factor & BatchNorm & Nonlinearity\\
        \hline
         PConv1& 7$\times$7& 64& 2& -& ReLU\\
         PConv2& 5$\times$5& 128& 2& Y& ReLU\\
         PConv3& 5$\times$5& 256& 2& Y& ReLU\\
         PConv4& 3$\times$3& 512& 2& Y& ReLU\\
         PConv5& 3$\times$3& 512& 2& Y& ReLU\\
         PConv6& 3$\times$3& 512& 2& Y& ReLU\\
         PConv7& 3$\times$3& 512& 2& Y& ReLU\\
         PConv8& 3$\times$3& 512& 2& Y& ReLU\\
         \hline
         NearestUpSample1& & 512& 2 & - & -\\
         Concat1(w/ PConv7)& &512+512&  & - & -\\
         PConv9&3$\times$3& 512 &1& Y& LeakyReLU(0.2)\\
         \hline
         NearestUpSample2& & 512& 2 & - & -\\
         Concat2(w/ PConv6)& &512+512&  & - & -\\
         PConv10&3$\times$3& 512 &1& Y& LeakyReLU(0.2)\\
         \hline
         NearestUpSample3& & 512& 2 & - & -\\
         Concat3(w/ PConv5)& &512+512&  & - & -\\
         PConv11&3$\times$3& 512 &1& Y& LeakyReLU(0.2)\\
         \hline
         NearestUpSample4& & 512& 2 & - & -\\
         Concat4(w/ PConv4)& &512+512&  & - & -\\
         PConv12&3$\times$3& 512 &1& Y& LeakyReLU(0.2)\\
         \hline
         NearestUpSample5& & 512& 2 & - & -\\
         Concat5(w/ PConv3)& &512+256&  & - & -\\
         PConv13&3$\times$3& 256 &1& Y& LeakyReLU(0.2)\\
         \hline
         NearestUpSample6& & 256& 2 & - & -\\
         Concat6(w/ PConv2)& &256+128&  & - & -\\
         PConv14&3$\times$3& 128 &1& Y& LeakyReLU(0.2)\\
         \hline
         NearestUpSample7& & 128& 2 & - & -\\
         Concat7(w/ PConv1)& &128+64&  & - & -\\
         PConv15&3$\times$3& 64 &1& Y& LeakyReLU(0.2)\\
         \hline
         NearestUpSample8& & 64& 2 & - & -\\
         Concat8(w/ Input)& &64+3&  & - & -\\
         PConv16&3$\times$3& 3 &1& - & -\\         
    \end{tabular}
    }
    \caption{PConv is defined as a partial convolutional layer with the specified filter size, stride and number of filters. PConv1-8 are in encoder stage, whereas PConv9-16 are in decoder stage. The BatchNorm column indicates whether PConv is followed by a Batch Normalization layer. The Nonlinearity column shows whether and what nonlinearity layer is used (following the BatchNorm if BatchNorm is used). Skip links are shown using Concat$*$, which concatenate the previous nearest neighbor upsampled results with the corresponding mentioned PConv\# results from the encoder stage.}
    \label{tab:net_detail}
\end{table}

 \clearpage

\subsection*{More Comparisons on Irregular Masks}
% \vspace{-2.cm}
\begin{figure}[h]
\vspace{-1cm}
    \centering
    \scalebox{0.86}{
    \begin{tabular}{cccccc}
    \multicolumn{6}{c}{}\\
    Input & PM & GL & GntIpt & PConv & GT \\
    \includegraphics[width=0.153\columnwidth]{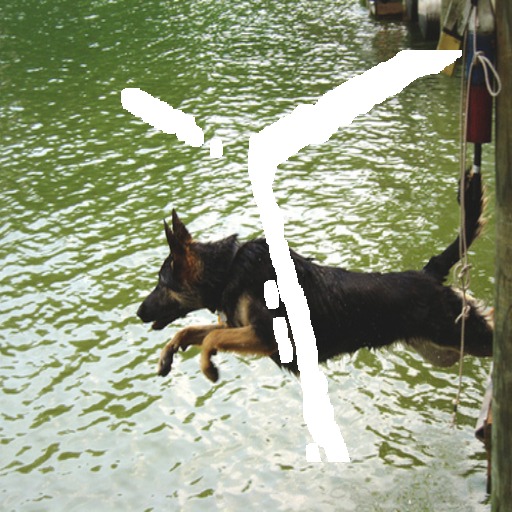} &
    \includegraphics[width=0.153\columnwidth]{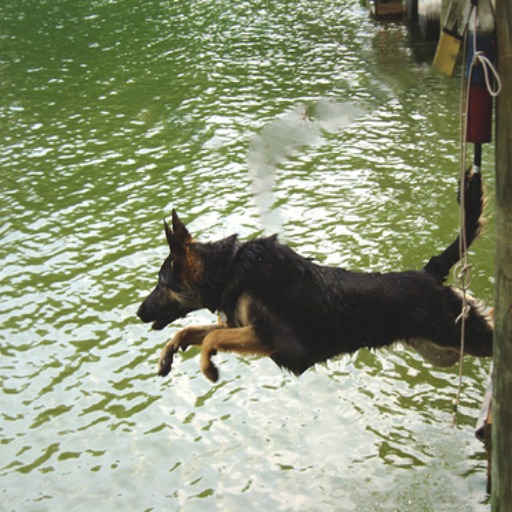} &
    \includegraphics[width=0.153\columnwidth]{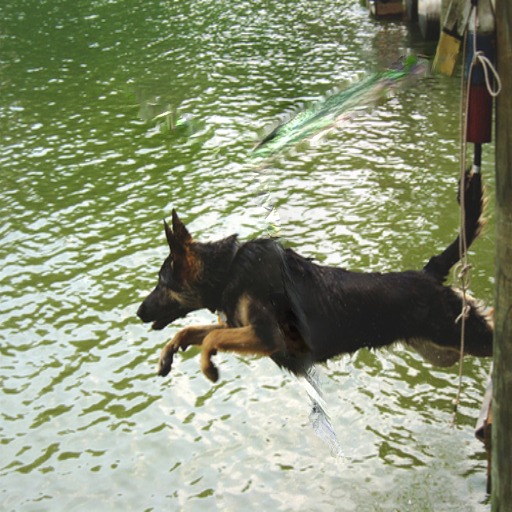} &
    \includegraphics[width=0.153\columnwidth]{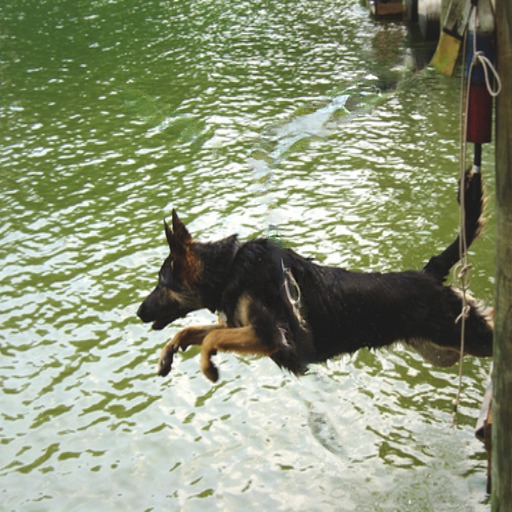} &
    \includegraphics[width=0.153\columnwidth]{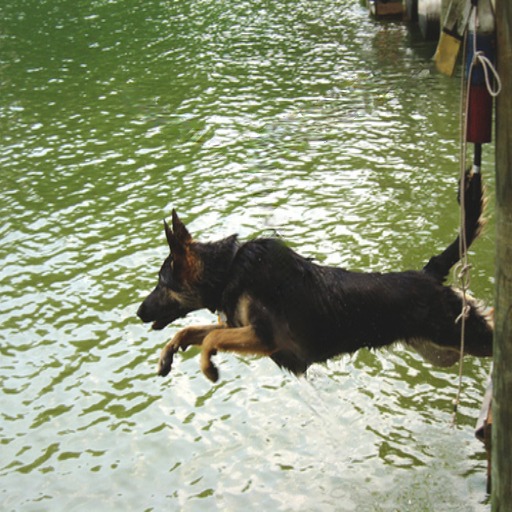} &
    \includegraphics[width=0.153\columnwidth]{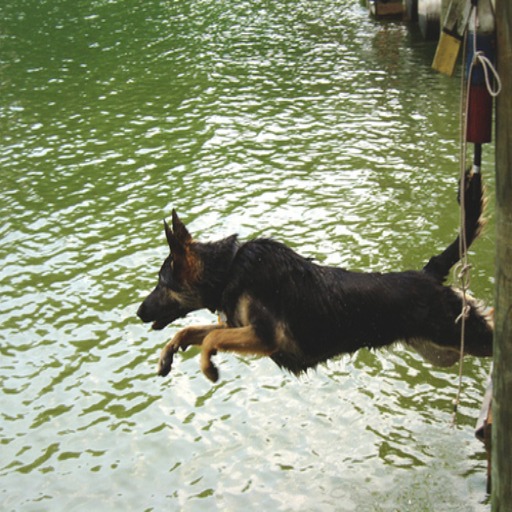} \\
    \includegraphics[width=0.153\columnwidth]{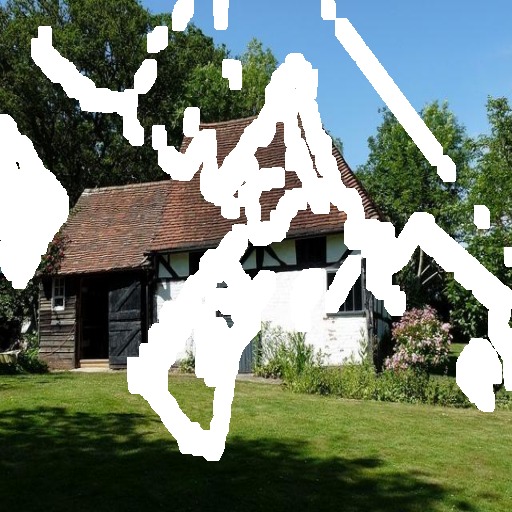} &
    \includegraphics[width=0.153\columnwidth]{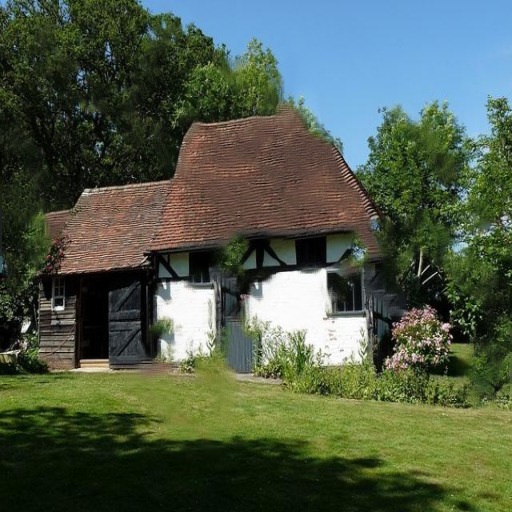} &
    \includegraphics[width=0.153\columnwidth]{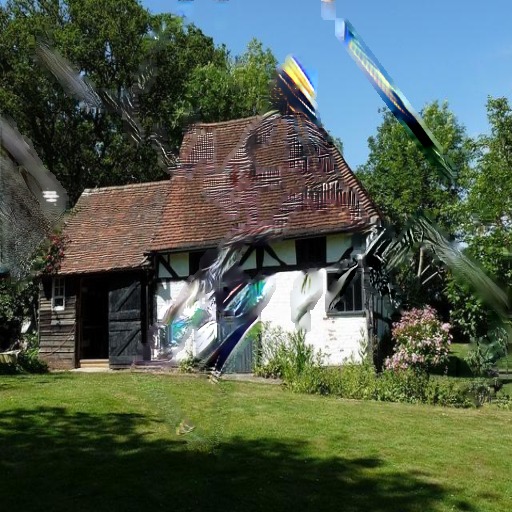} &
    \includegraphics[width=0.153\columnwidth]{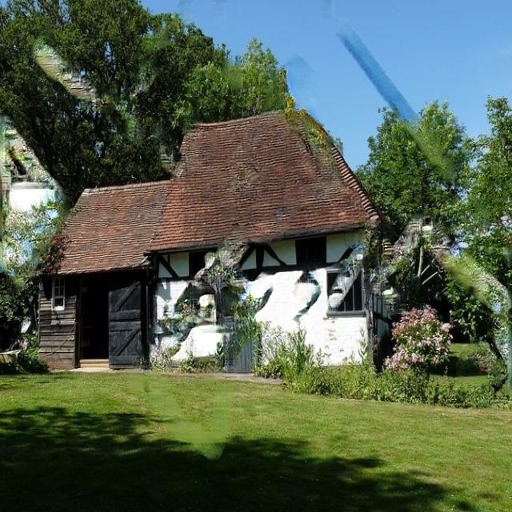} &
    \includegraphics[width=0.153\columnwidth]{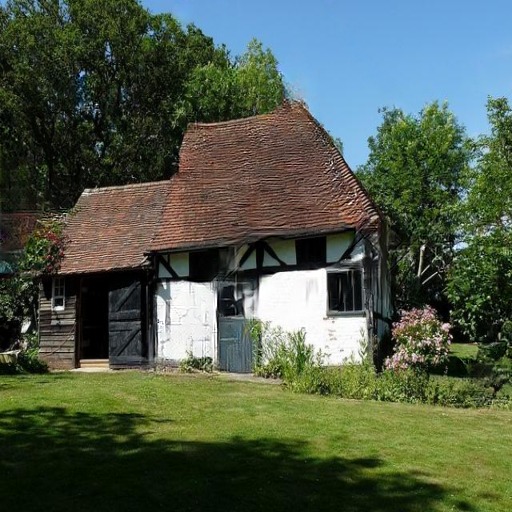} &
    \includegraphics[width=0.153\columnwidth]{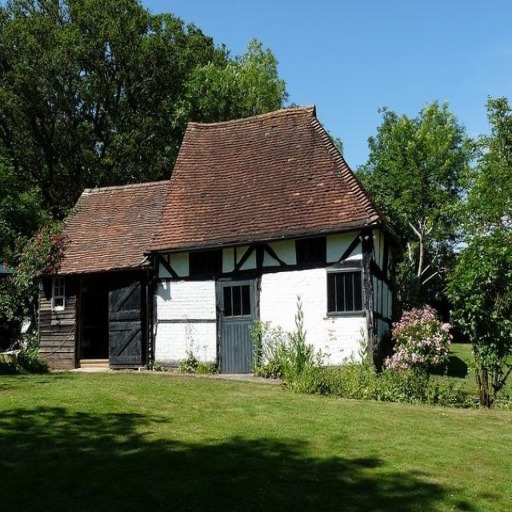} \\    
    \includegraphics[width=0.153\columnwidth]{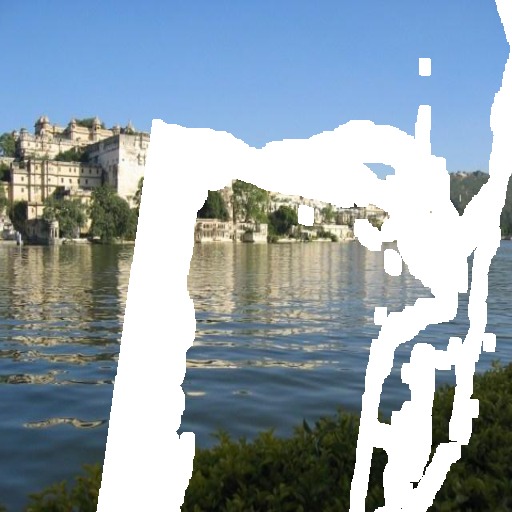} &
    \includegraphics[width=0.153\columnwidth]{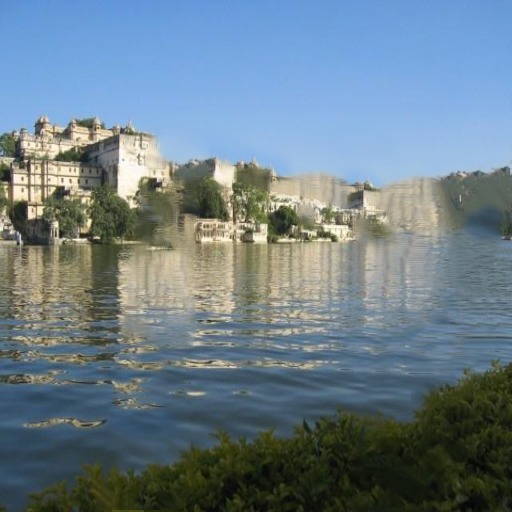} &
    \includegraphics[width=0.153\columnwidth]{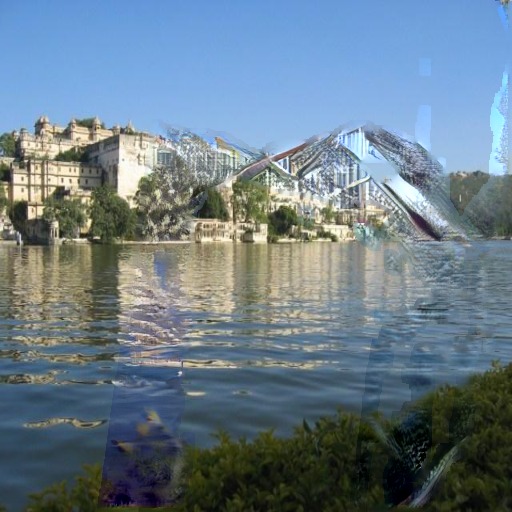} &
    \includegraphics[width=0.153\columnwidth]{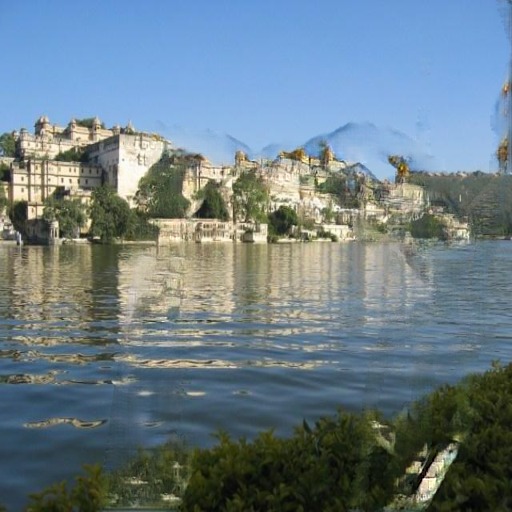} &
    \includegraphics[width=0.153\columnwidth]{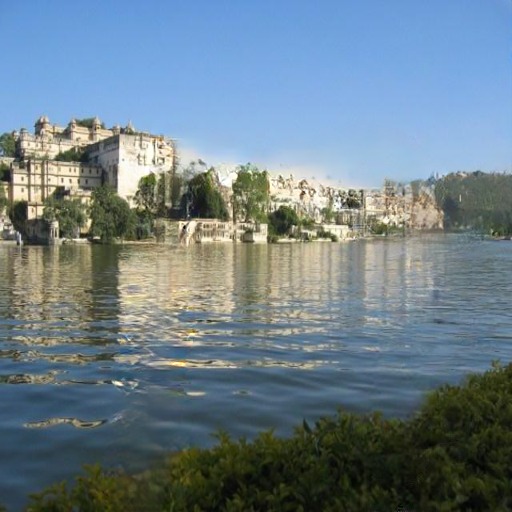} &
    \includegraphics[width=0.153\columnwidth]{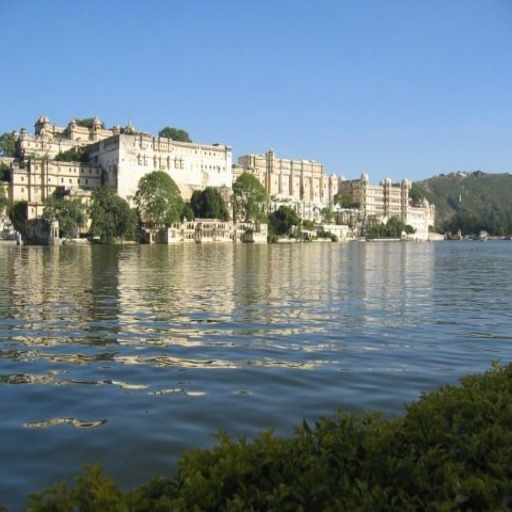} \\           
    \includegraphics[width=0.153\columnwidth]{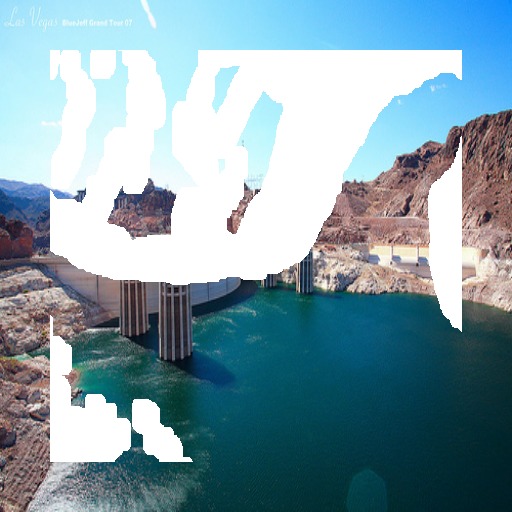} &
    \includegraphics[width=0.153\columnwidth]{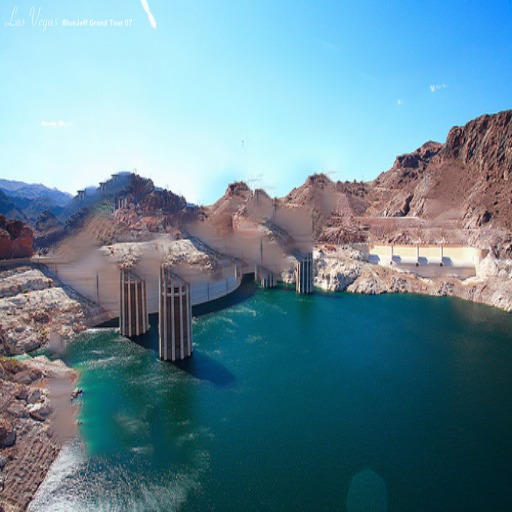} &
    \includegraphics[width=0.153\columnwidth]{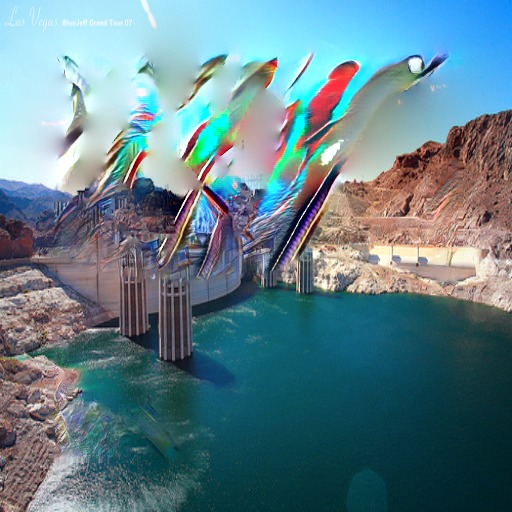} &
    \includegraphics[width=0.153\columnwidth]{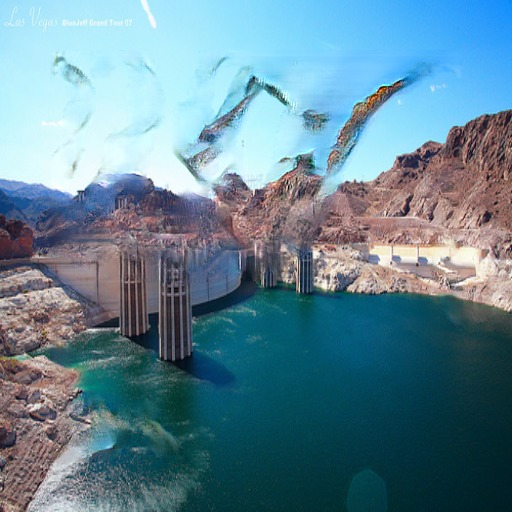} &
    \includegraphics[width=0.153\columnwidth]{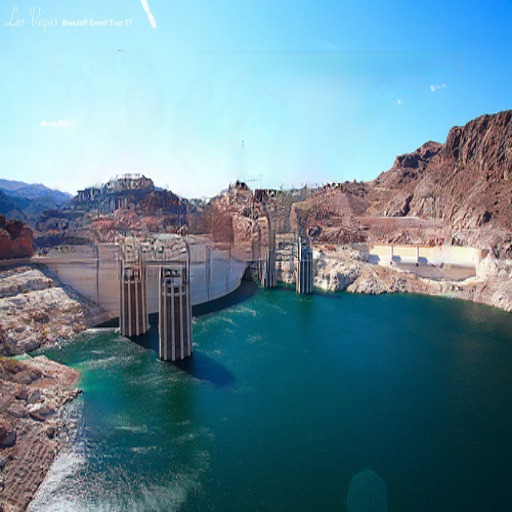} &
    \includegraphics[width=0.153\columnwidth]{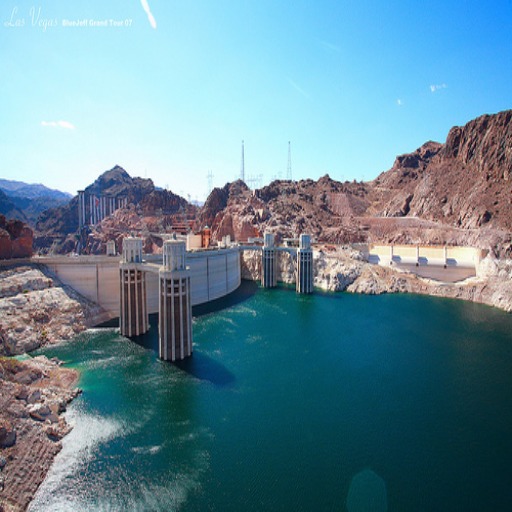} \\         
    \includegraphics[width=0.153\columnwidth]{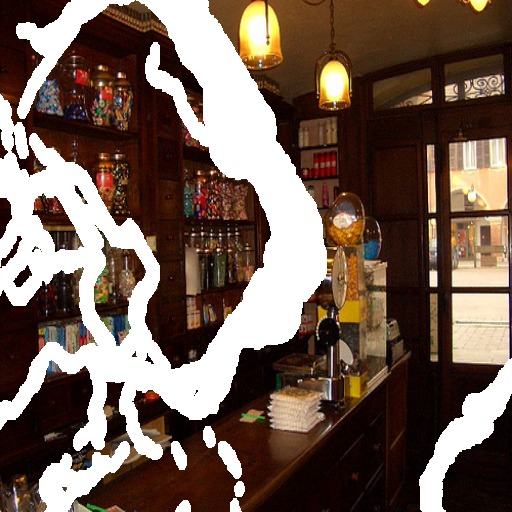} &
    \includegraphics[width=0.153\columnwidth]{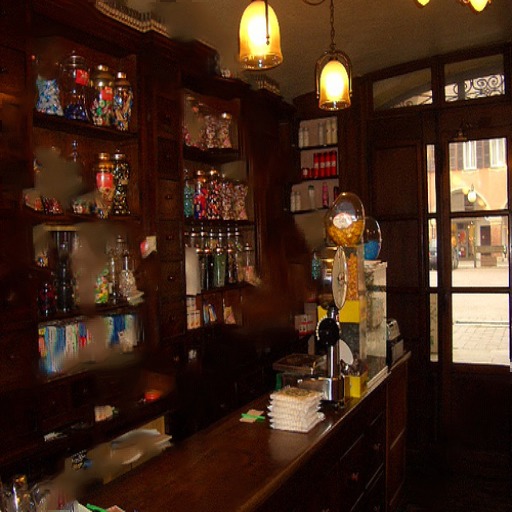} &
    \includegraphics[width=0.153\columnwidth]{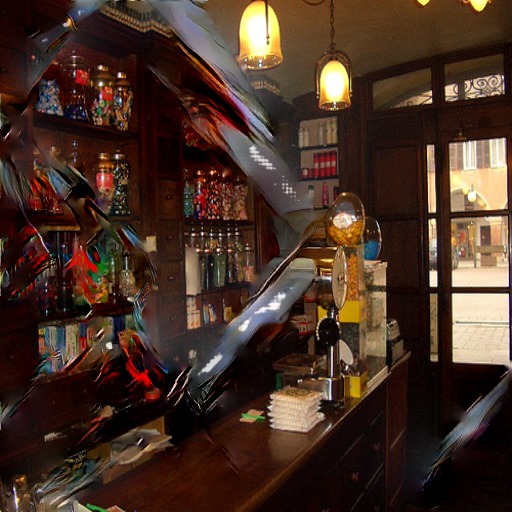} &
    \includegraphics[width=0.153\columnwidth]{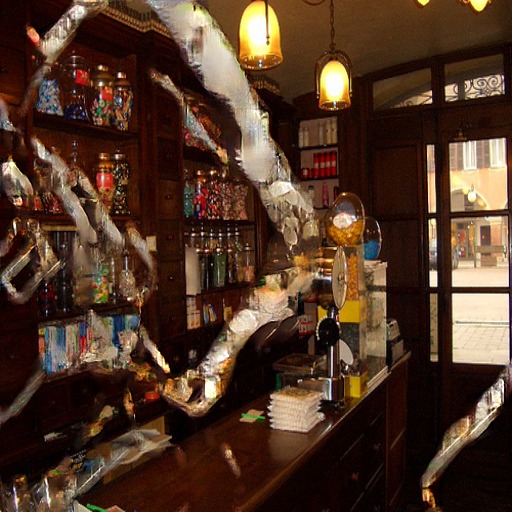} &
    \includegraphics[width=0.153\columnwidth]{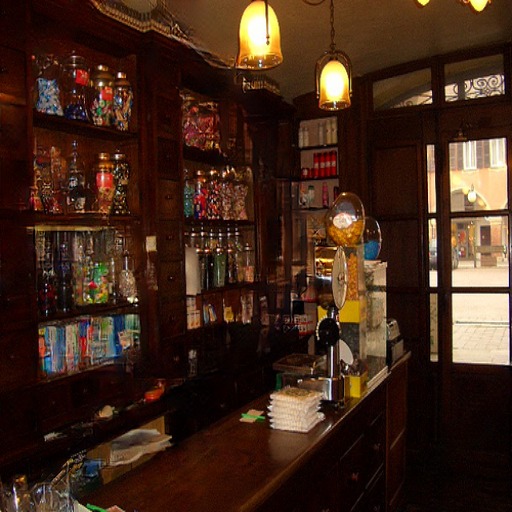} &
    \includegraphics[width=0.153\columnwidth]{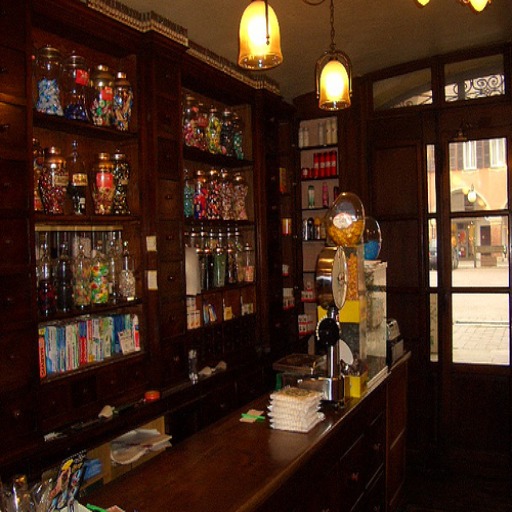} \\     
    \includegraphics[width=0.153\columnwidth]{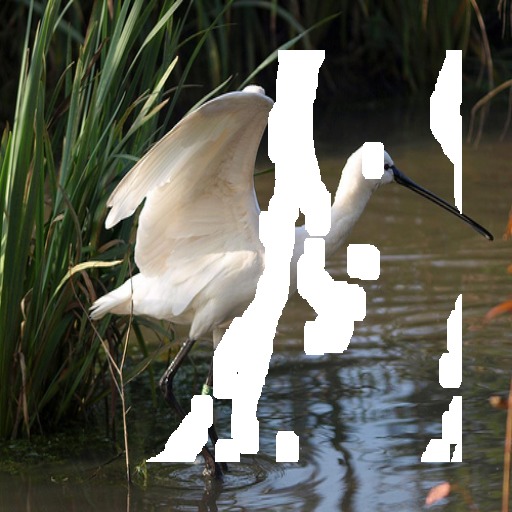} &
    \includegraphics[width=0.153\columnwidth]{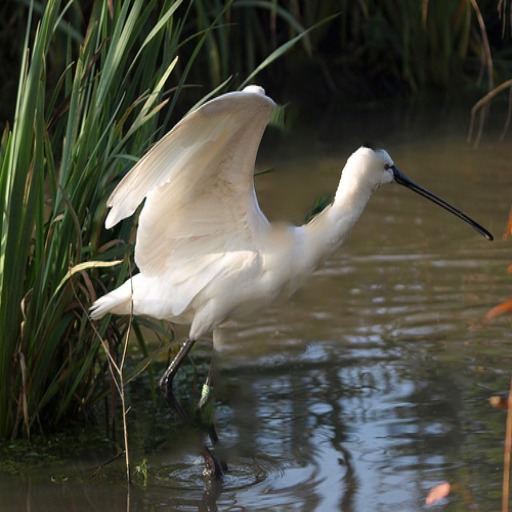} &
    \includegraphics[width=0.153\columnwidth]{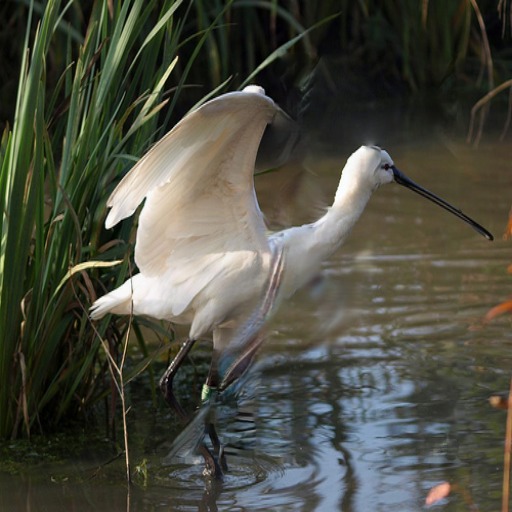} &
    \includegraphics[width=0.153\columnwidth]{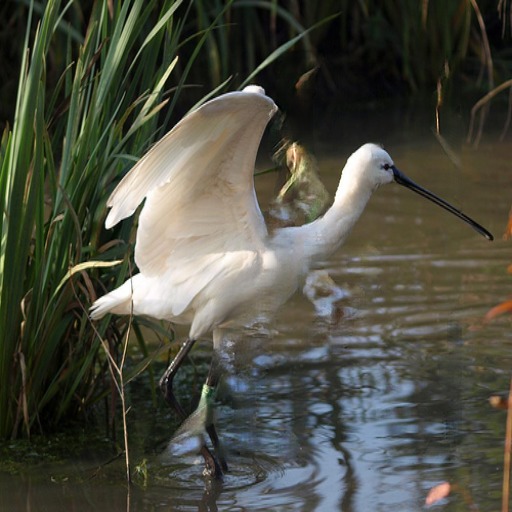} &
    \includegraphics[width=0.153\columnwidth]{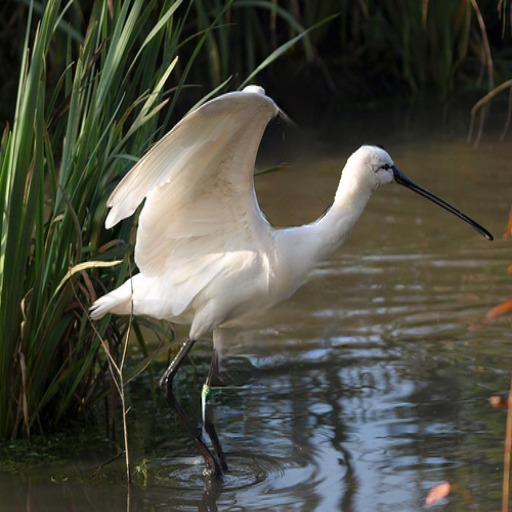} &
    \includegraphics[width=0.153\columnwidth]{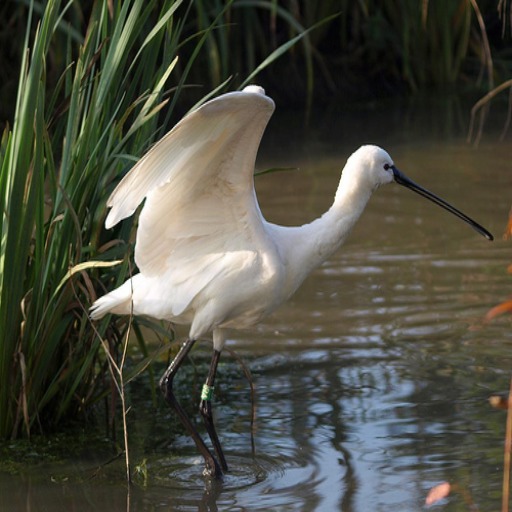} \\     
    \includegraphics[width=0.153\columnwidth]{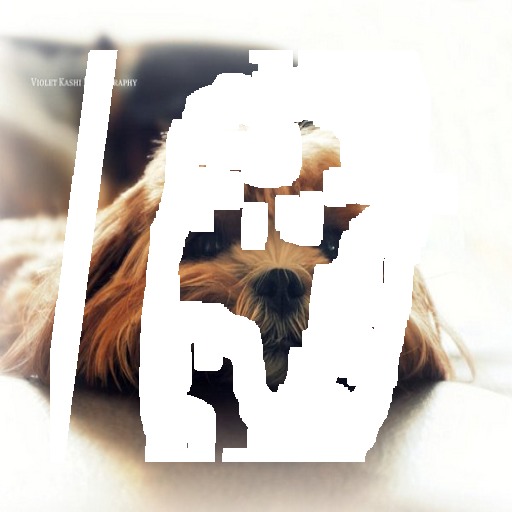} &
    \includegraphics[width=0.153\columnwidth]{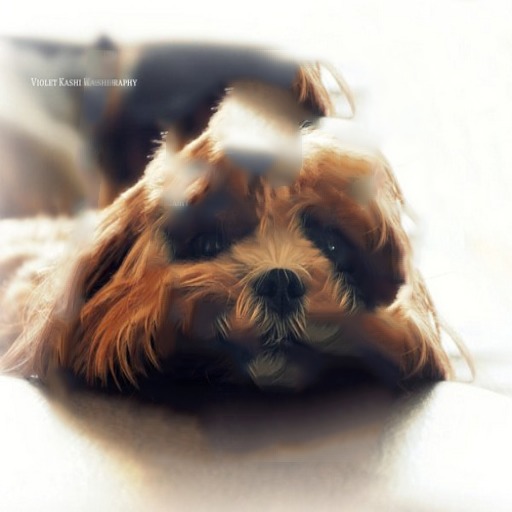} &
    \includegraphics[width=0.153\columnwidth]{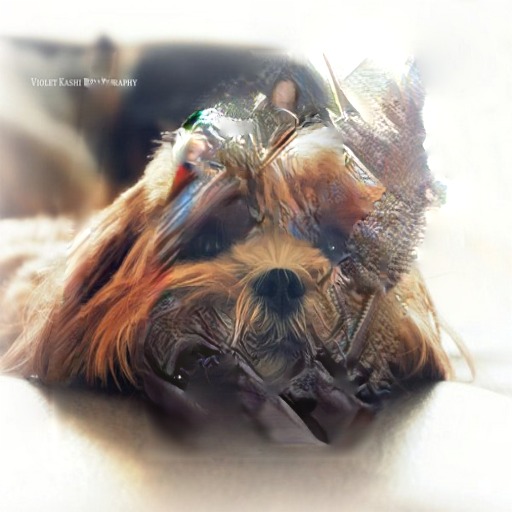} &
    \includegraphics[width=0.153\columnwidth]{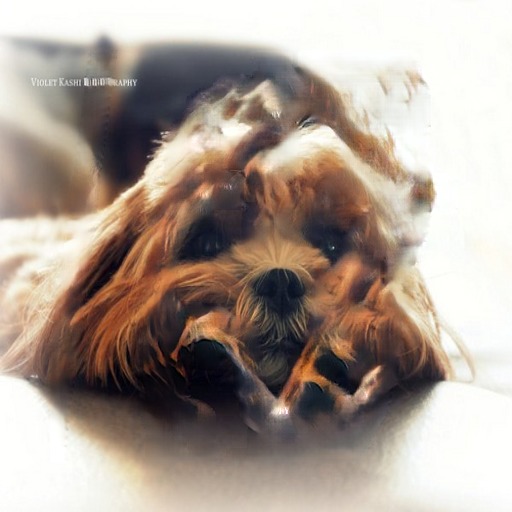} &
    \includegraphics[width=0.153\columnwidth]{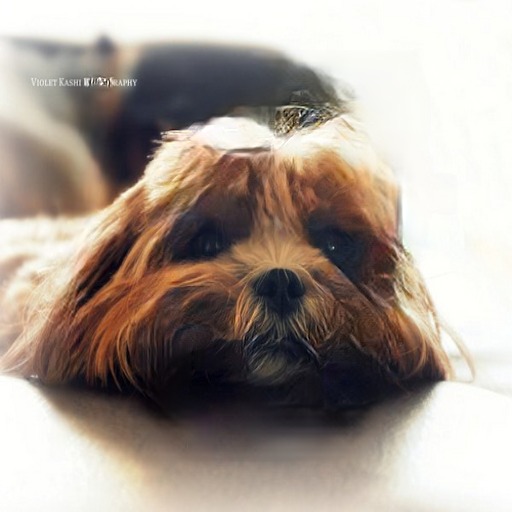} &
    \includegraphics[width=0.153\columnwidth]{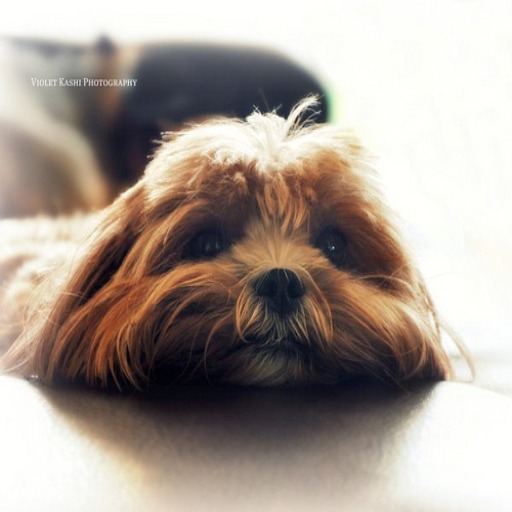} \\      
    \includegraphics[width=0.158\columnwidth]{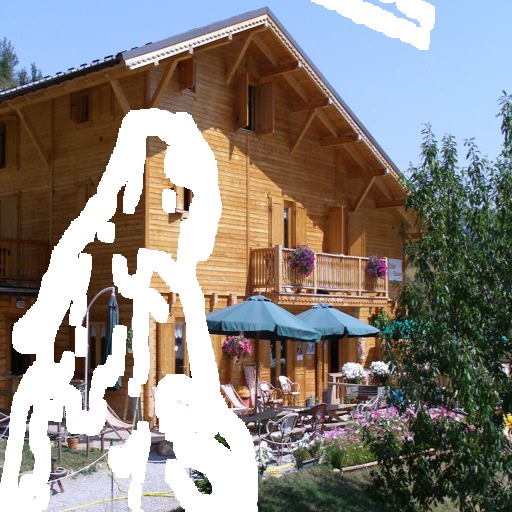} & 
    \includegraphics[width=0.158\columnwidth]{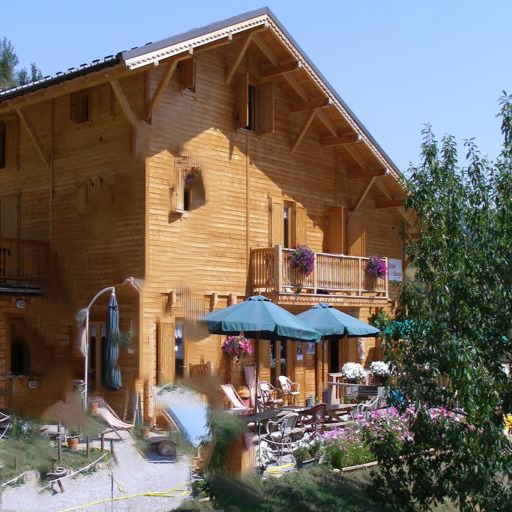} &
    \includegraphics[width=0.158\columnwidth]{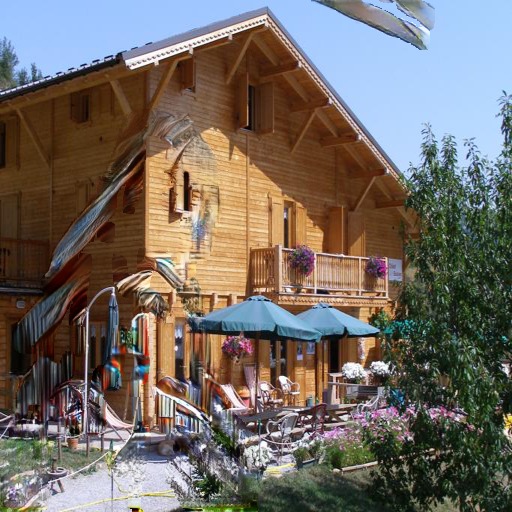} &
    \includegraphics[width=0.158\columnwidth]{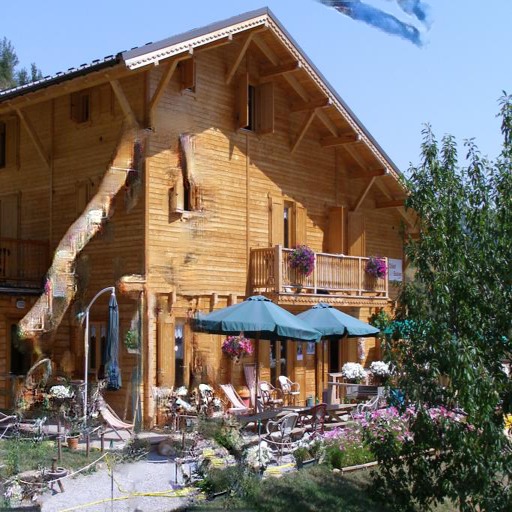} &
    \includegraphics[width=0.158\columnwidth]{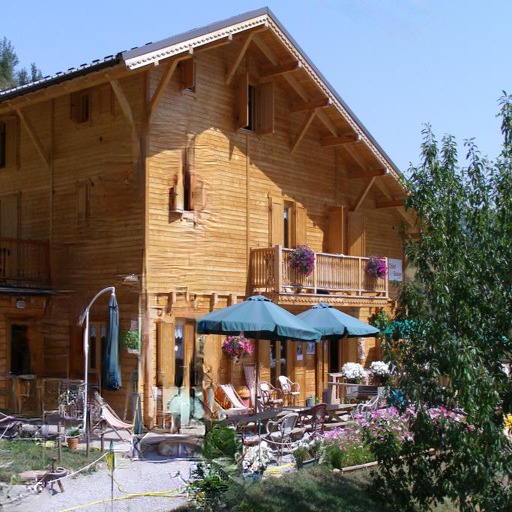} &
    \includegraphics[width=0.158\columnwidth]{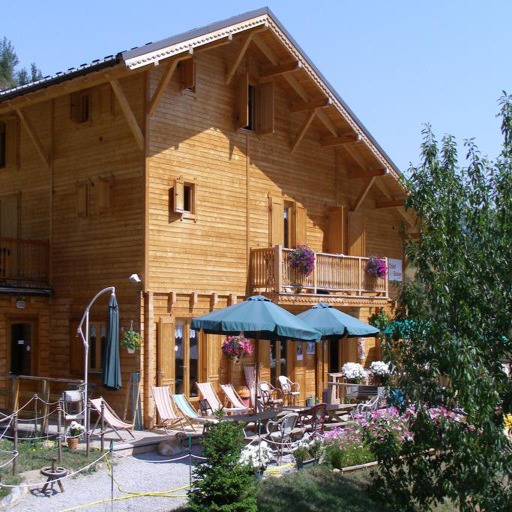} \\    
    \end{tabular}
    }
    \caption{Comparisons on irregular masks. The abbreviations of the notations are the same as Figure~5 and Figure~6 in the paper.}
    \label{fig:compare_irregualr}
\end{figure}

\clearpage

\subsection*{More Comparisons on Regular Masks}

\begin{figure}[h]
    \centering
    \scalebox{0.9}{
    \begin{tabular}{cccccc}
    \multicolumn{6}{c}{}\\
    Input & PM & GL & GntIpt & PConv & GT \\
    \includegraphics[width=0.153\columnwidth]{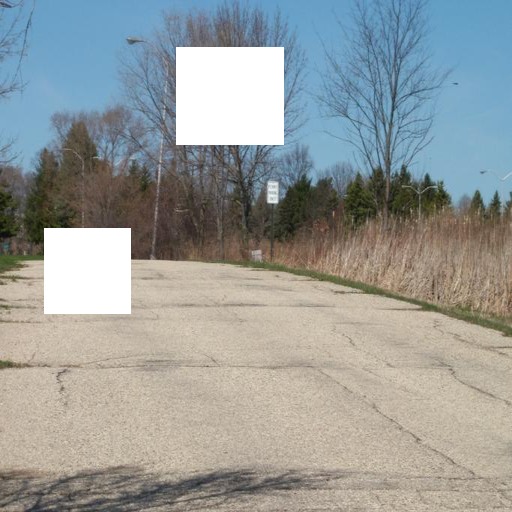} &
    \includegraphics[width=0.153\columnwidth]{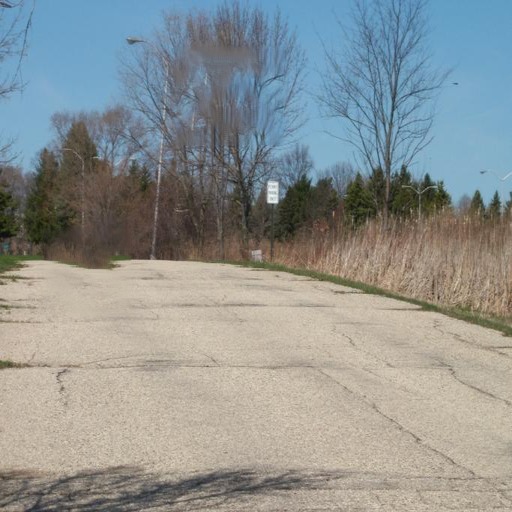} &
    \includegraphics[width=0.153\columnwidth]{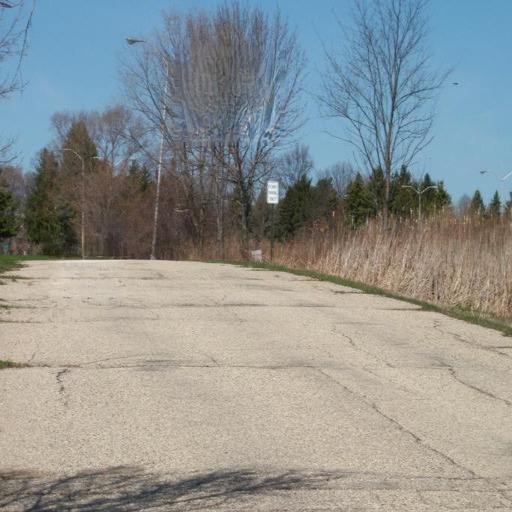} &
    \includegraphics[width=0.153\columnwidth]{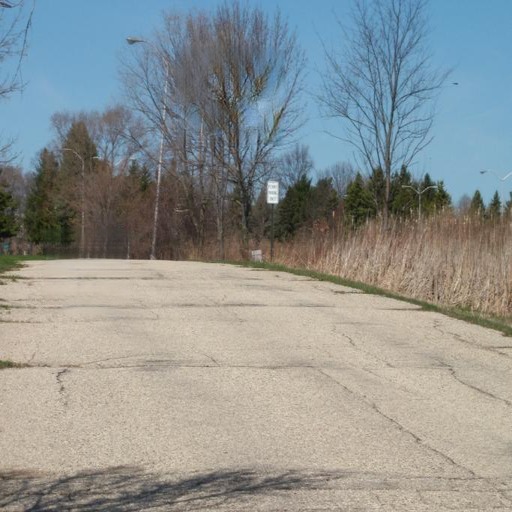} &
    \includegraphics[width=0.153\columnwidth]{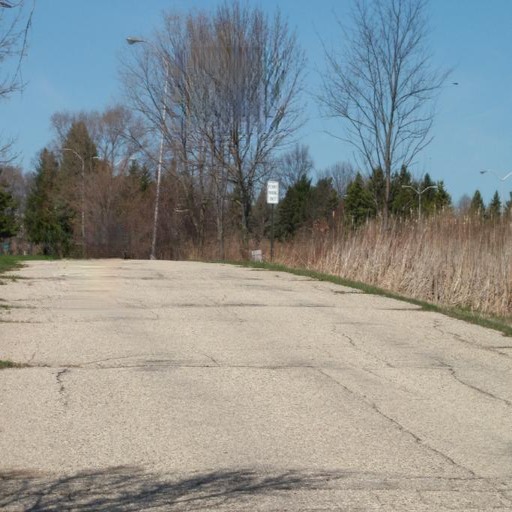} &
    \includegraphics[width=0.153\columnwidth]{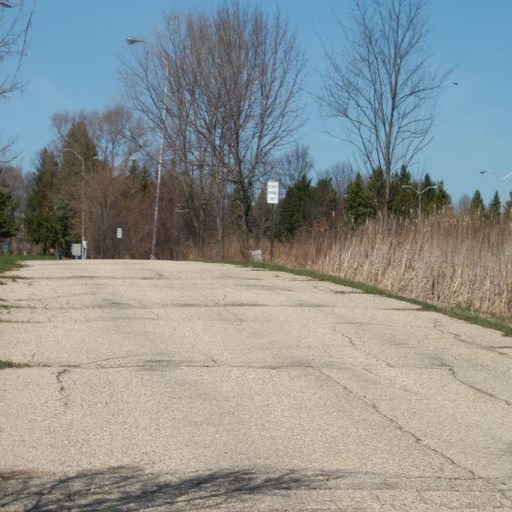} \\    
    \includegraphics[width=0.153\columnwidth]{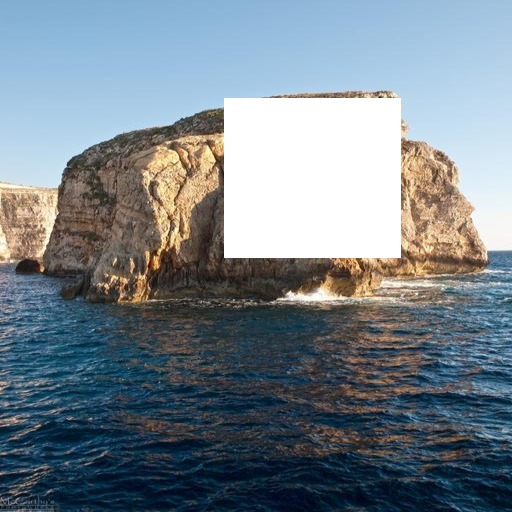} &
    \includegraphics[width=0.153\columnwidth]{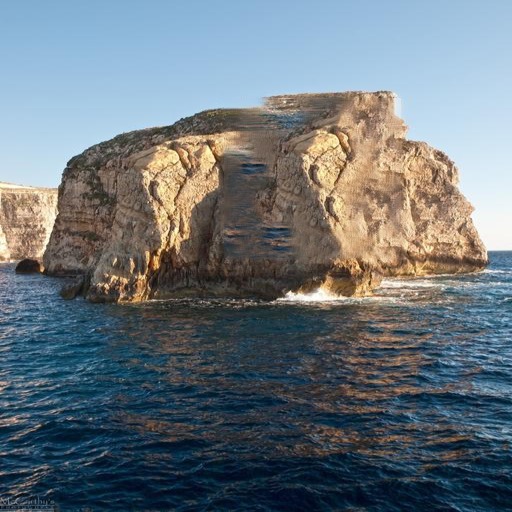} &
    \includegraphics[width=0.153\columnwidth]{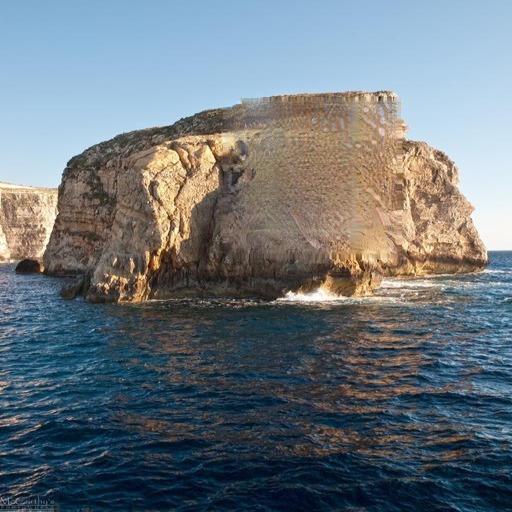} &
    \includegraphics[width=0.153\columnwidth]{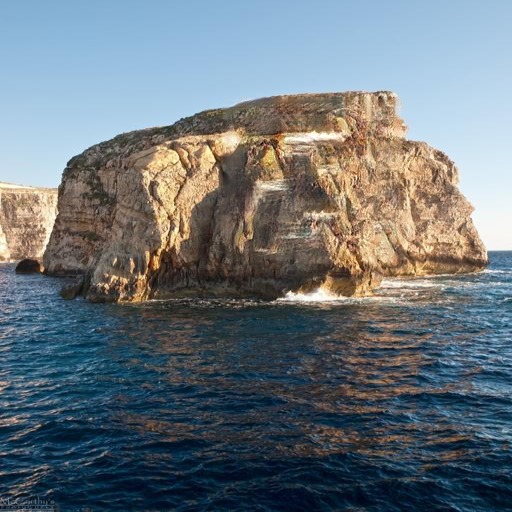} &
    \includegraphics[width=0.153\columnwidth]{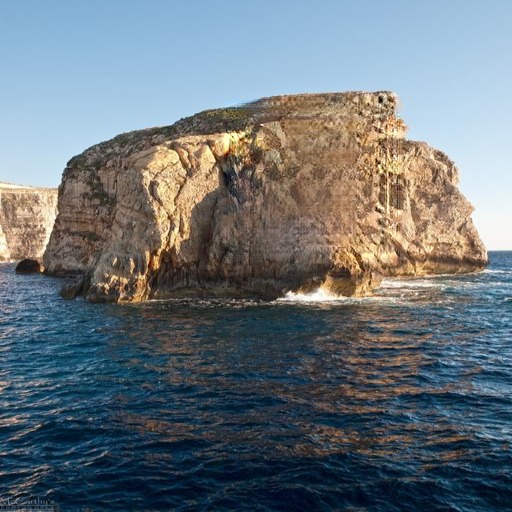} &
    \includegraphics[width=0.153\columnwidth]{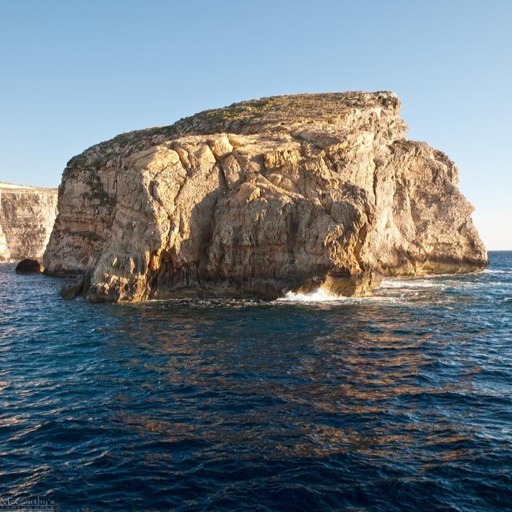} \\      
    \includegraphics[width=0.153\columnwidth]{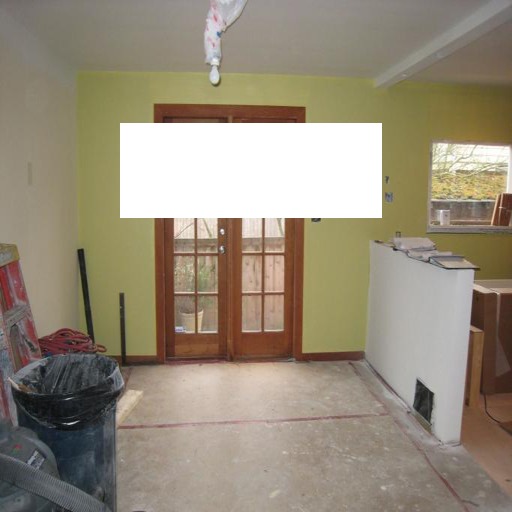} &
    \includegraphics[width=0.153\columnwidth]{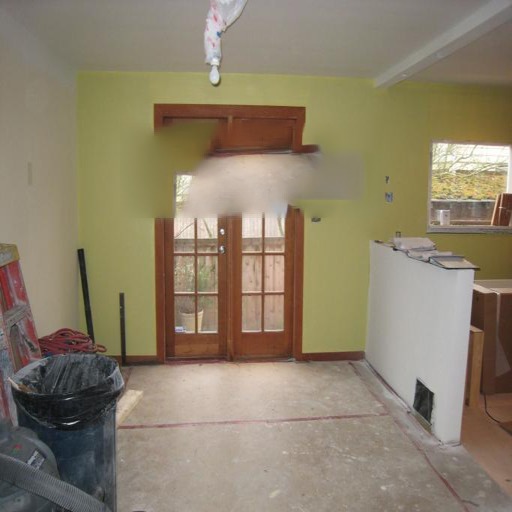} &
    \includegraphics[width=0.153\columnwidth]{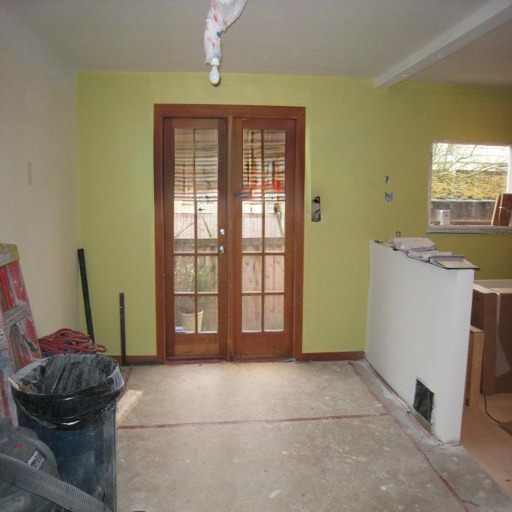} &
    \includegraphics[width=0.153\columnwidth]{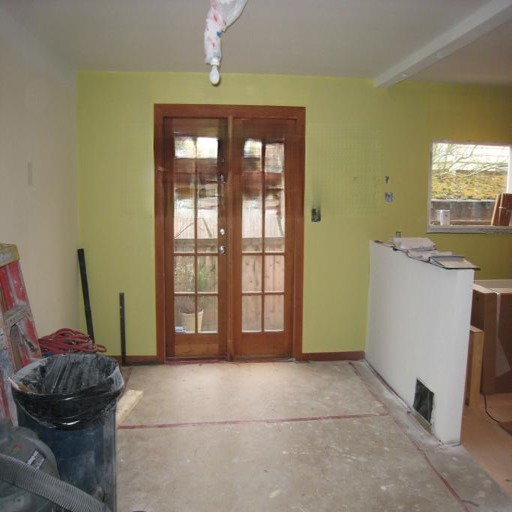} &
    \includegraphics[width=0.153\columnwidth]{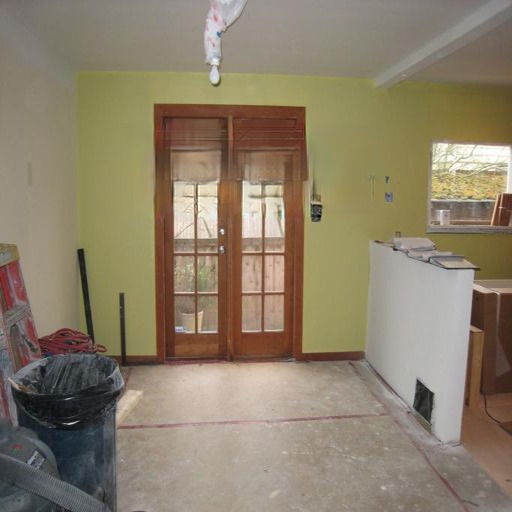} &
    \includegraphics[width=0.153\columnwidth]{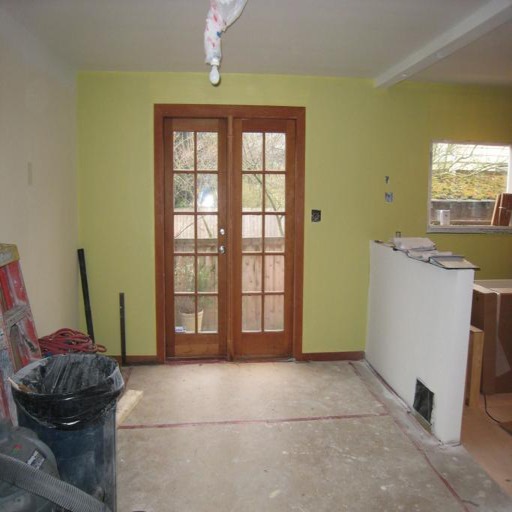} \\
    \includegraphics[width=0.153\columnwidth]{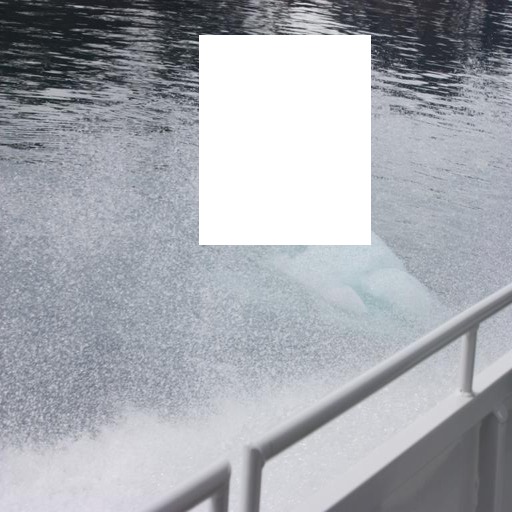} &
    \includegraphics[width=0.153\columnwidth]{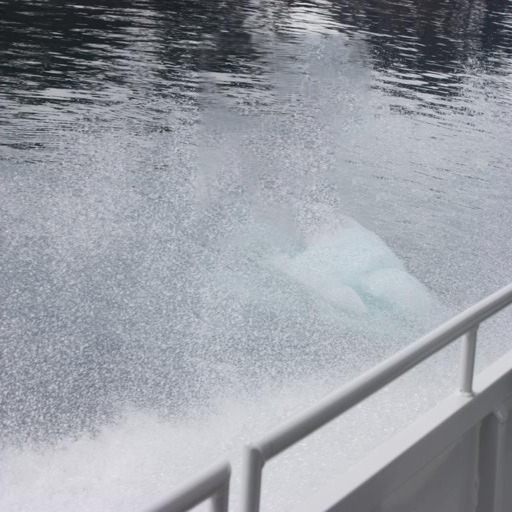} &
    \includegraphics[width=0.153\columnwidth]{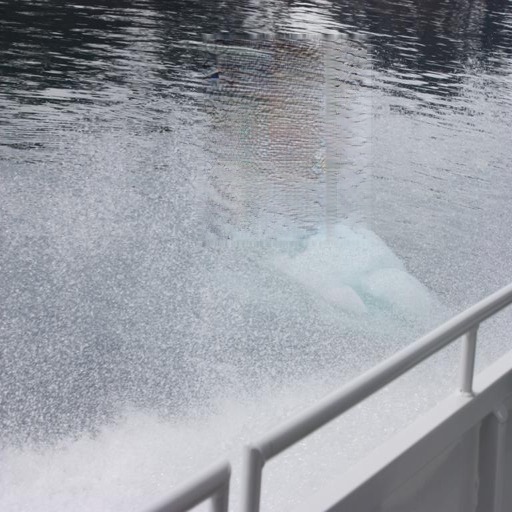} &
    \includegraphics[width=0.153\columnwidth]{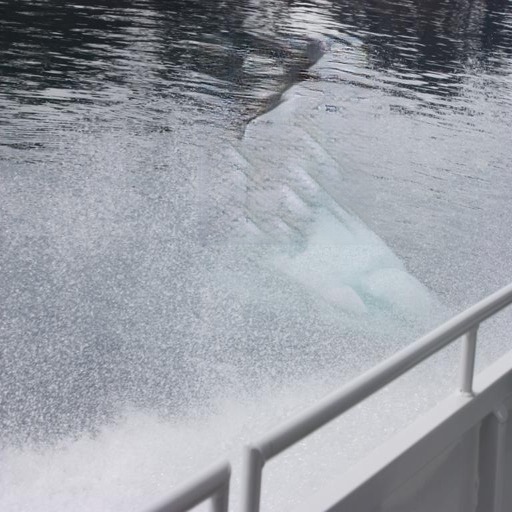} &
    \includegraphics[width=0.153\columnwidth]{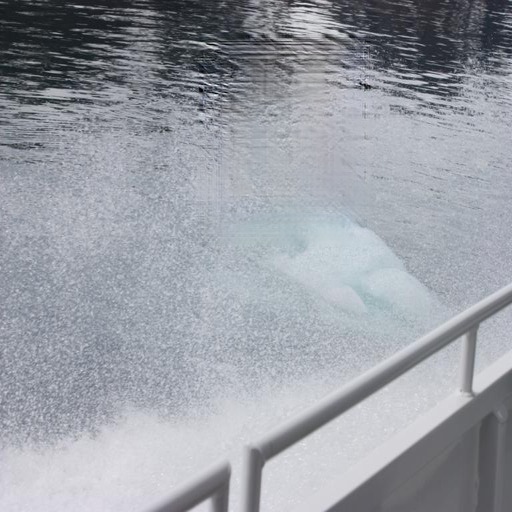} &
    \includegraphics[width=0.153\columnwidth]{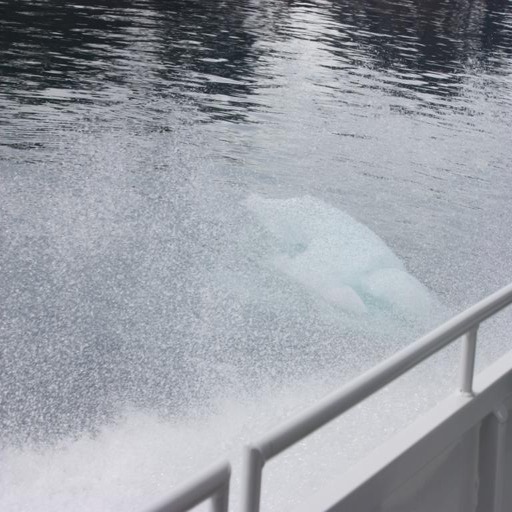} \\         
    \includegraphics[width=0.153\columnwidth]{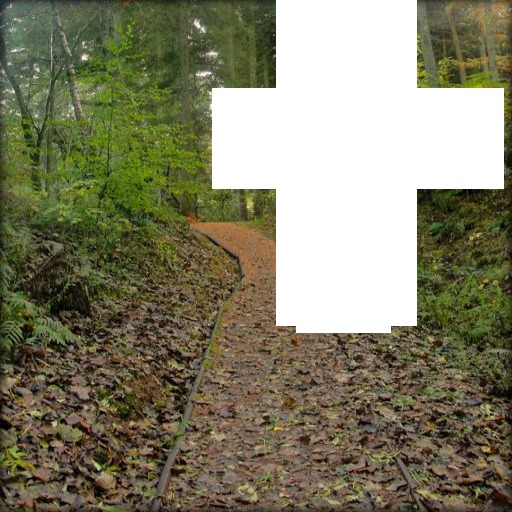} &
    \includegraphics[width=0.153\columnwidth]{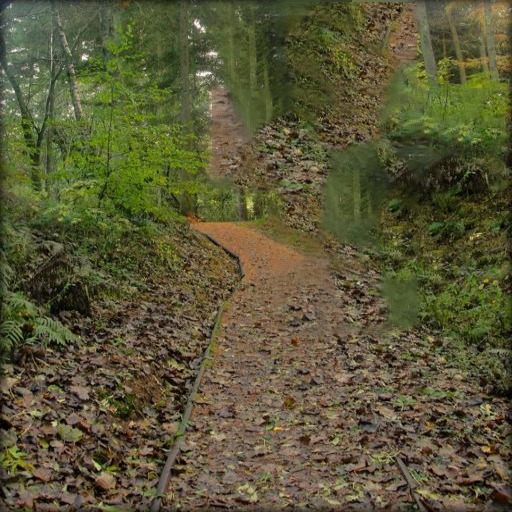} &
    \includegraphics[width=0.153\columnwidth]{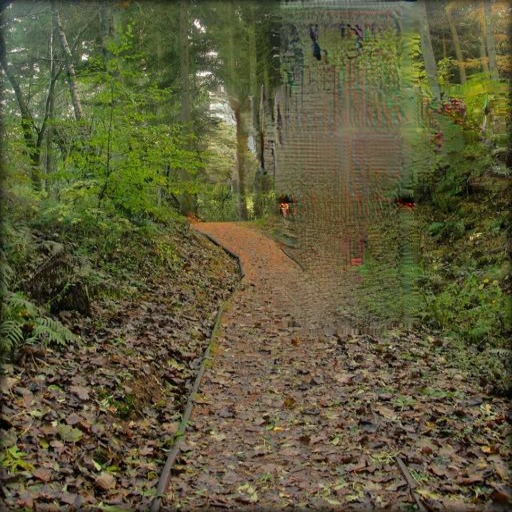} &
    \includegraphics[width=0.153\columnwidth]{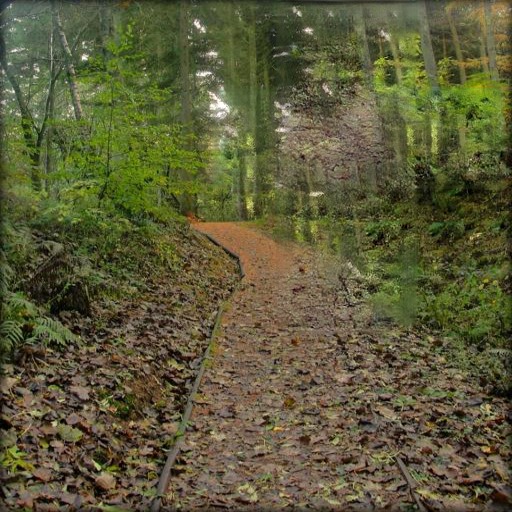} &
    \includegraphics[width=0.153\columnwidth]{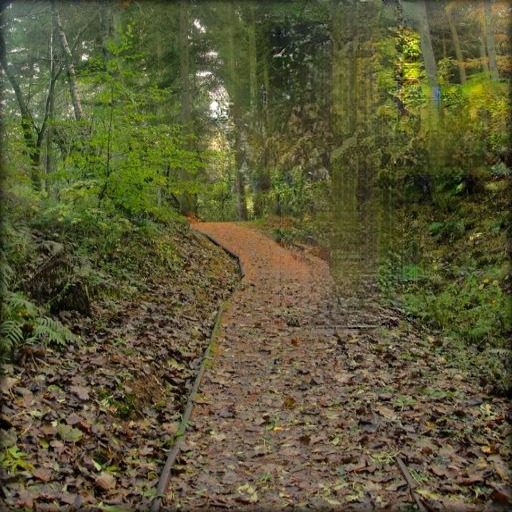} &
    \includegraphics[width=0.153\columnwidth]{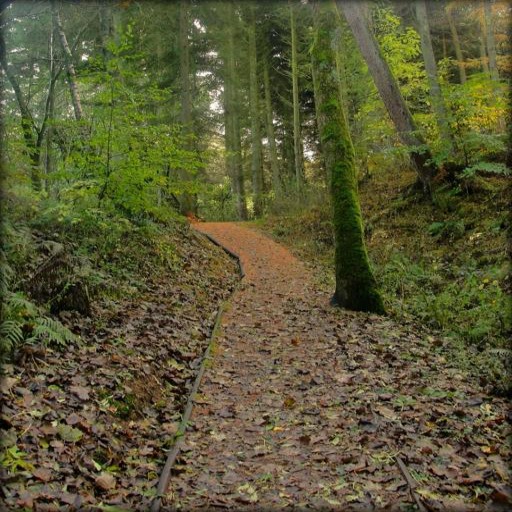} \\     
    \includegraphics[width=0.153\columnwidth]{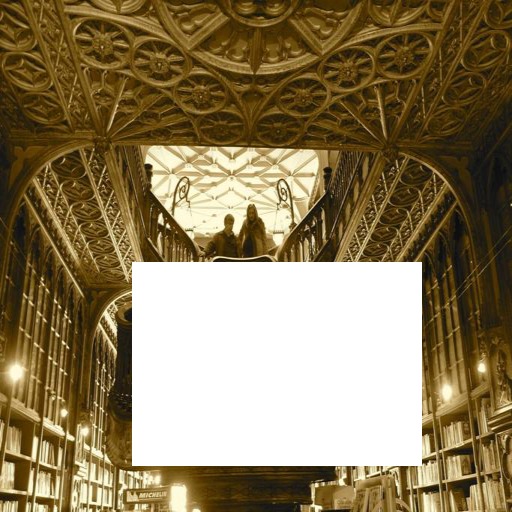} &
    \includegraphics[width=0.153\columnwidth]{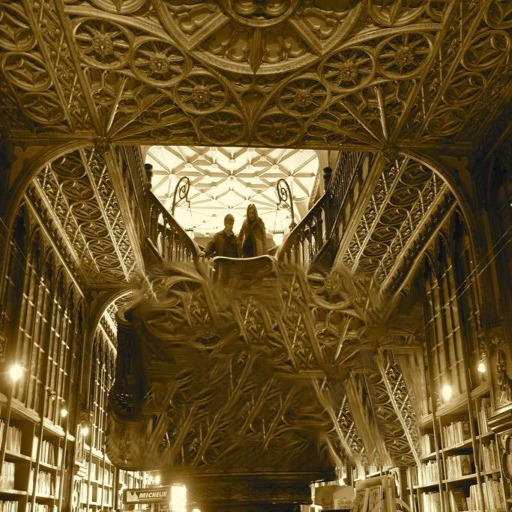} &
    \includegraphics[width=0.153\columnwidth]{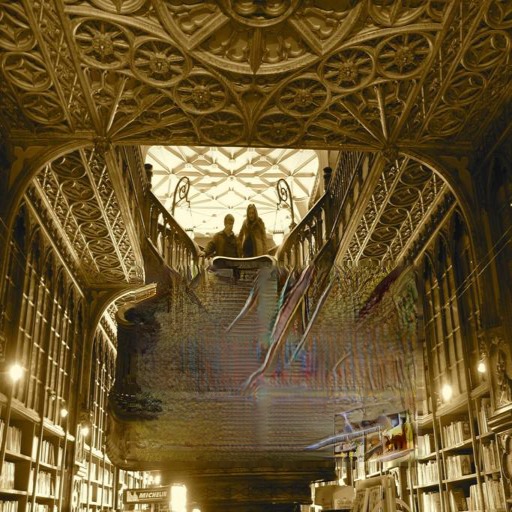} &
    \includegraphics[width=0.153\columnwidth]{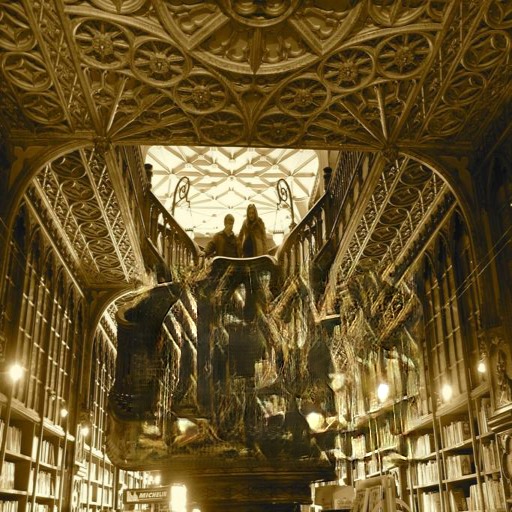} &
    \includegraphics[width=0.153\columnwidth]{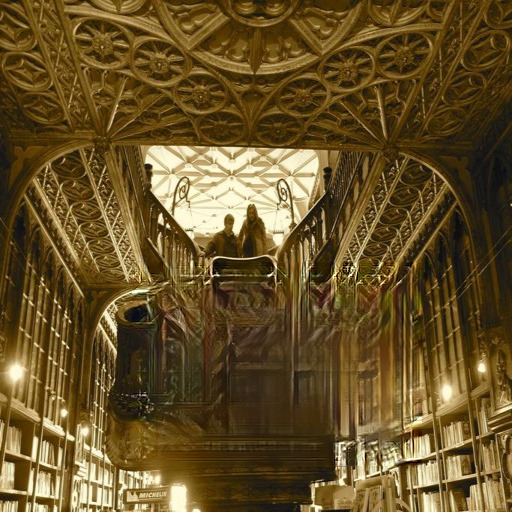} &
    \includegraphics[width=0.153\columnwidth]{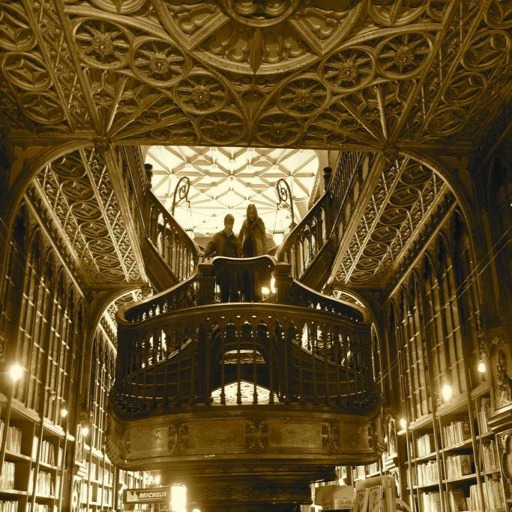} \\     
    \includegraphics[width=0.153\columnwidth]{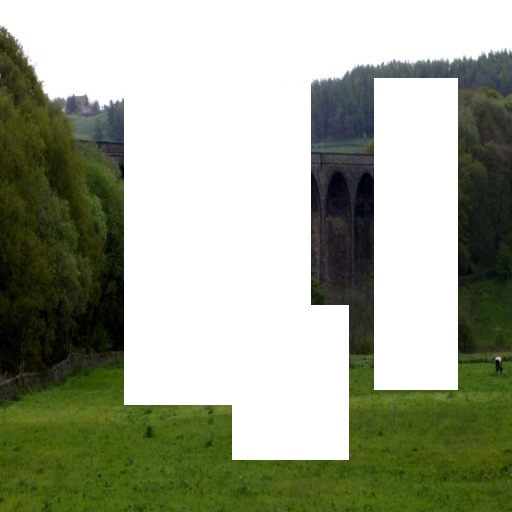} &
    \includegraphics[width=0.153\columnwidth]{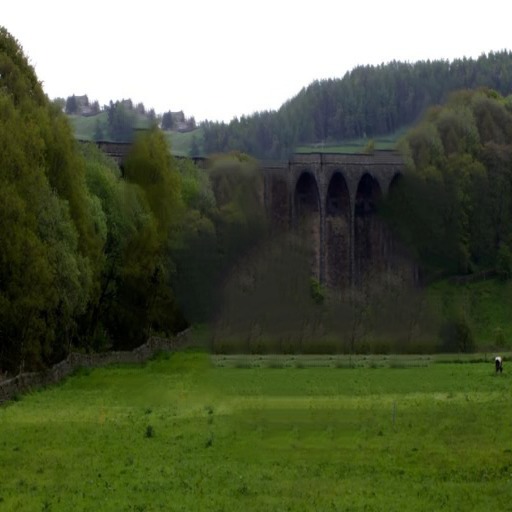} &
    \includegraphics[width=0.153\columnwidth]{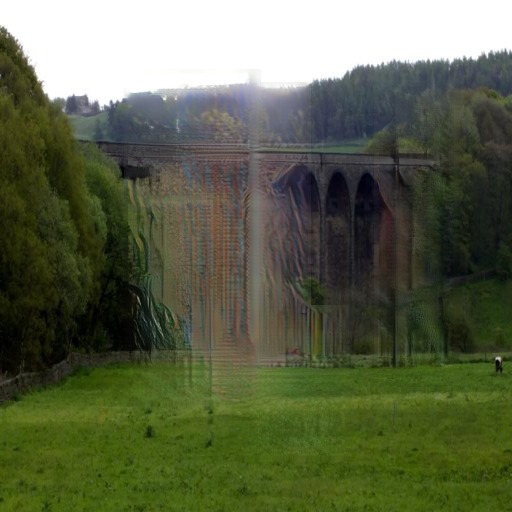} &
    \includegraphics[width=0.153\columnwidth]{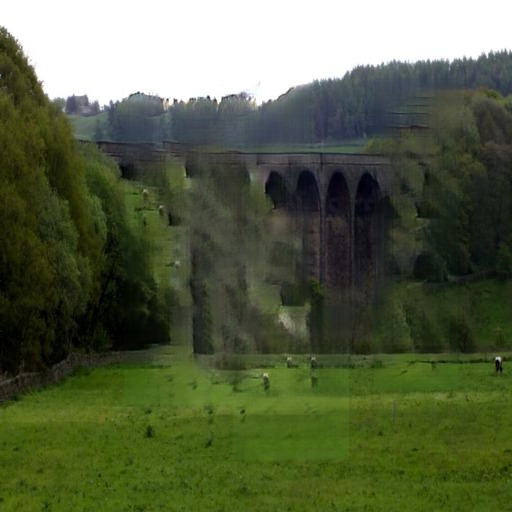} &
    \includegraphics[width=0.153\columnwidth]{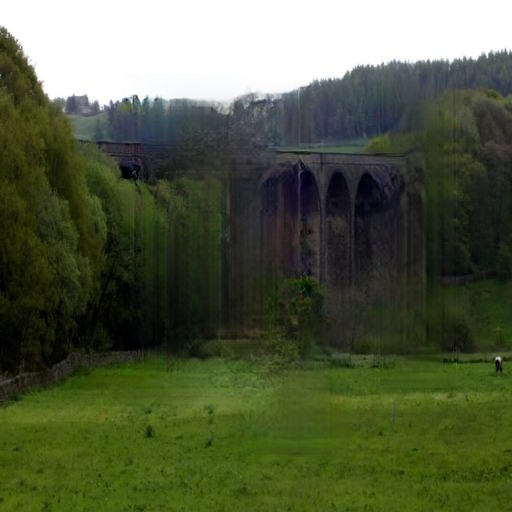} &
    \includegraphics[width=0.153\columnwidth]{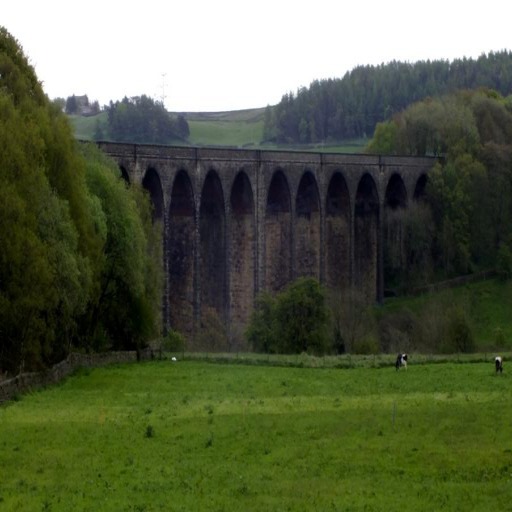} \\      
    \end{tabular}
    }
    \caption{Comparisons on regular masks. The abbreviations of the notations are the same as Figure~5 and Figure~6 in the paper.}
    \label{fig:compare_regualr}
\end{figure}

\clearpage

\subsection*{More Comparisons On Image Super Resolution}

\begin{figure}[h!]
\centering
    \scalebox{0.95}{
    \begin{tabular}{ccccc}
    \multicolumn{5}{c}{}\\
    Bicubic & SRGAN & MDSR+ & PConv & GT \\
    \includegraphics[width=0.19\columnwidth]{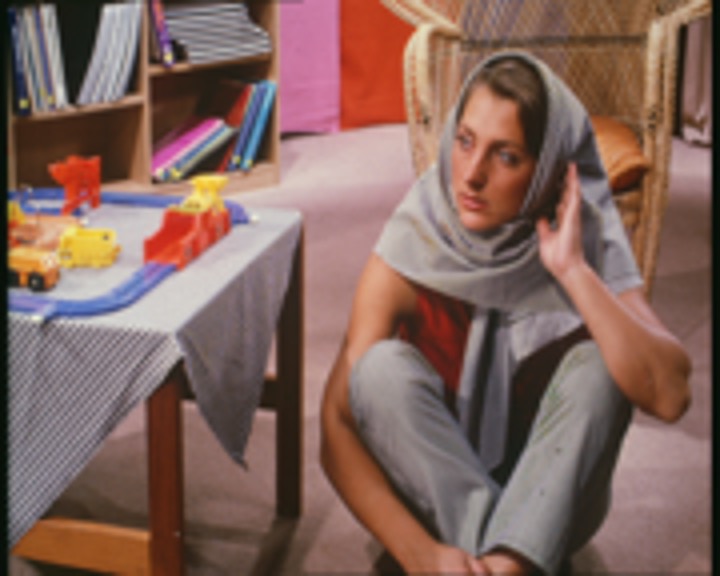} &
    \includegraphics[width=0.19\columnwidth]{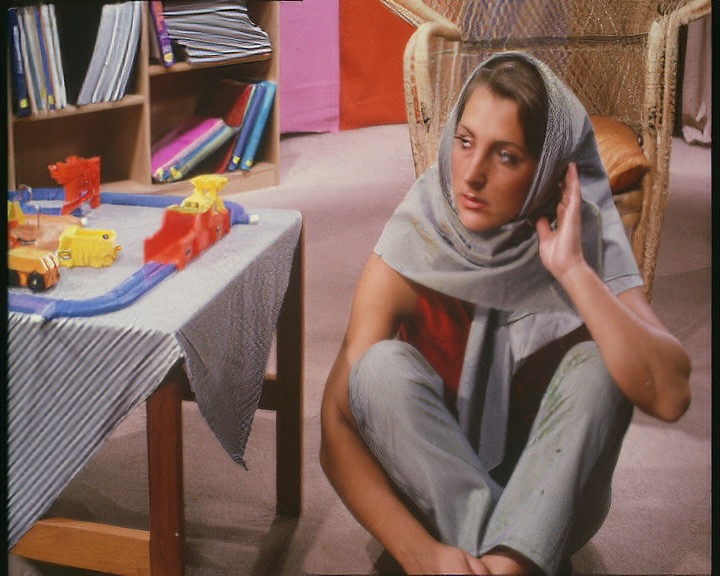} &
    \includegraphics[width=0.19\columnwidth]{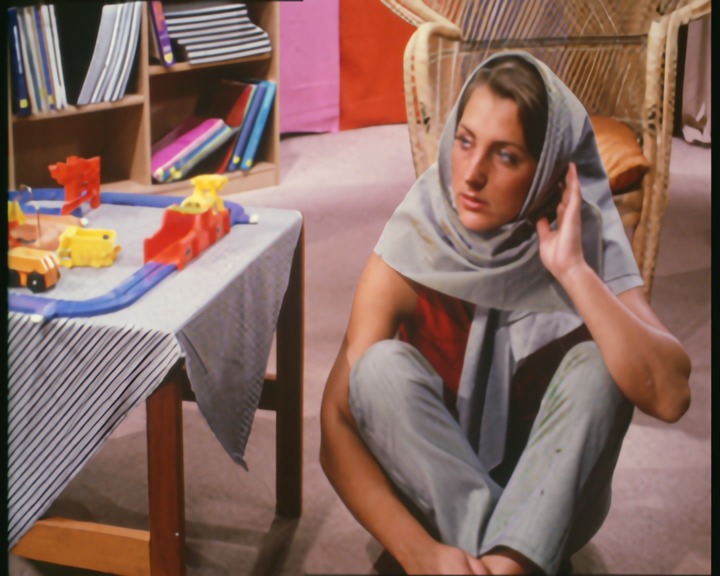} &
    \includegraphics[width=0.19\columnwidth]{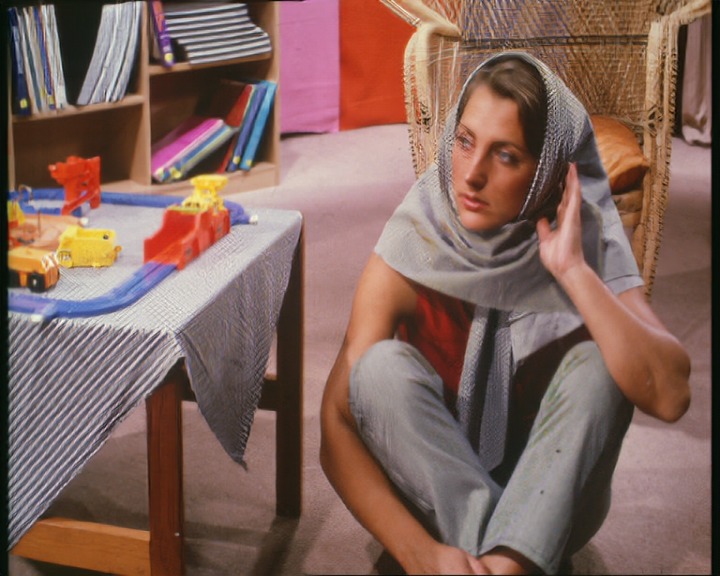} &
    \includegraphics[width=0.19\columnwidth]{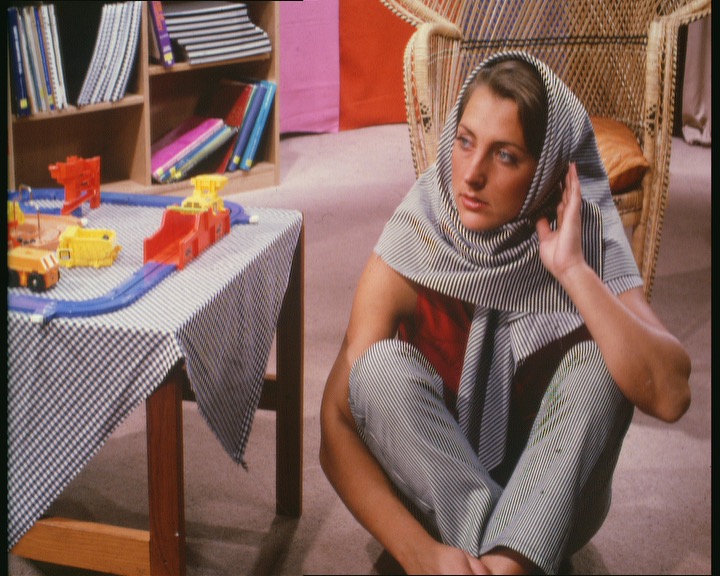} \\    
    \includegraphics[width=0.19\columnwidth]{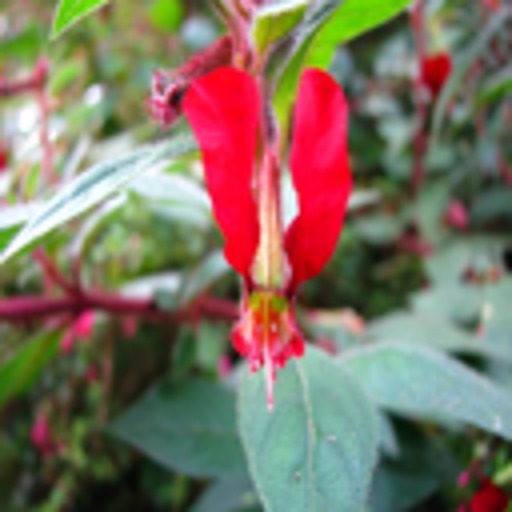} &
    \includegraphics[width=0.19\columnwidth]{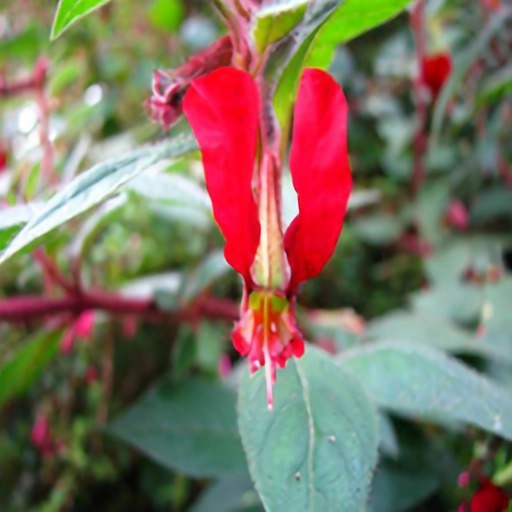} &
    \includegraphics[width=0.19\columnwidth]{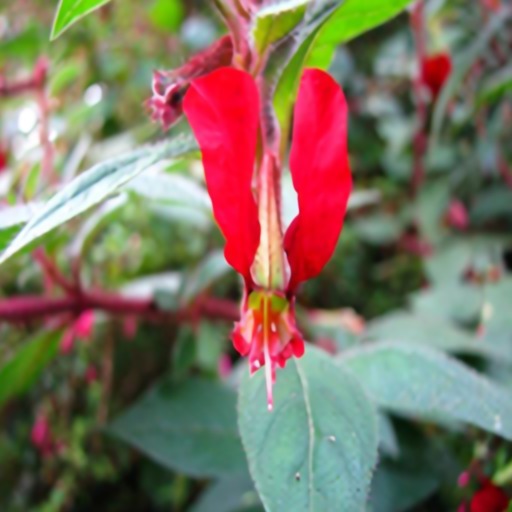} &
    \includegraphics[width=0.19\columnwidth]{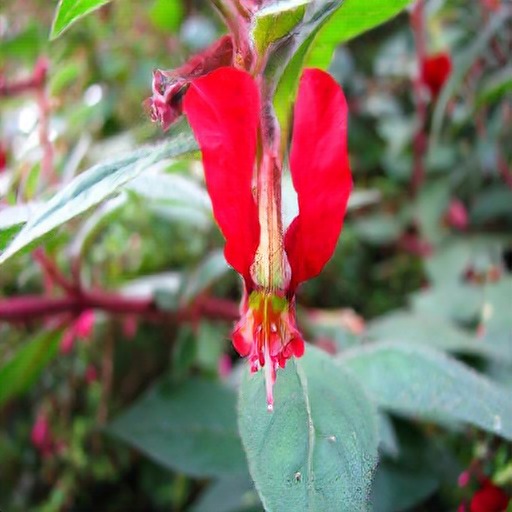} &
    \includegraphics[width=0.19\columnwidth]{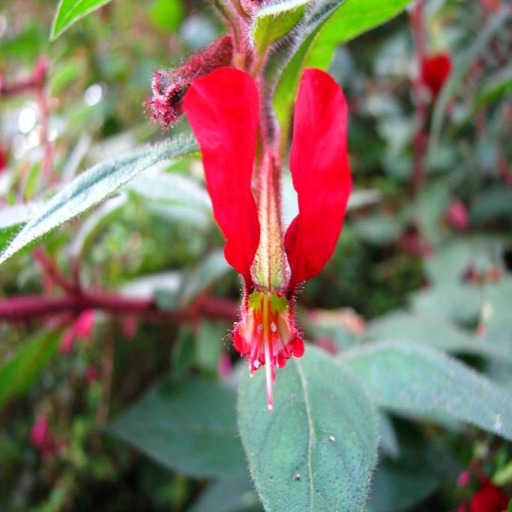} \\    
    \includegraphics[width=0.19\columnwidth]{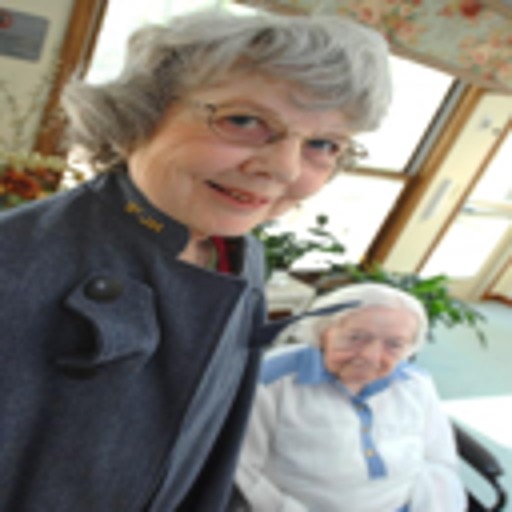} &
    \includegraphics[width=0.19\columnwidth]{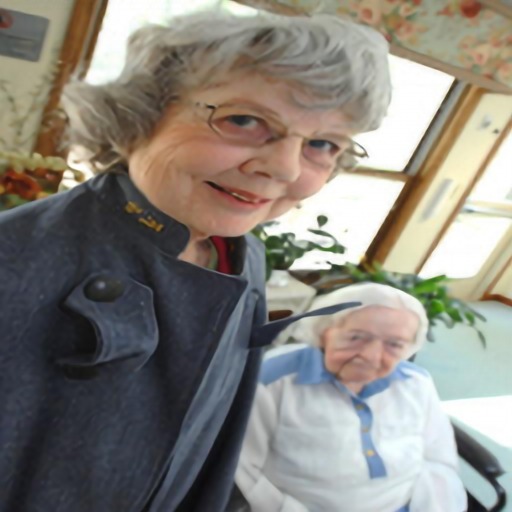} &
    \includegraphics[width=0.19\columnwidth]{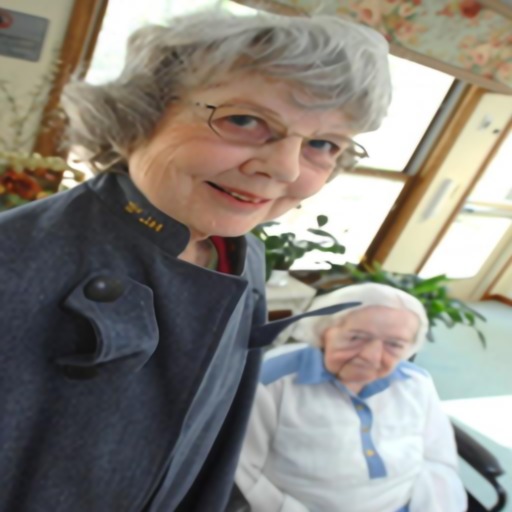} &
    \includegraphics[width=0.19\columnwidth]{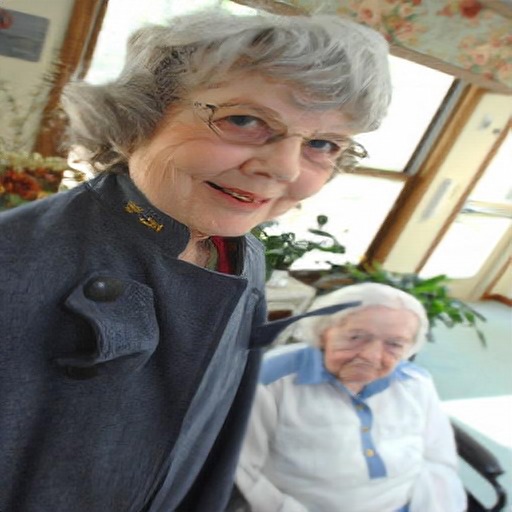} &
    \includegraphics[width=0.19\columnwidth]{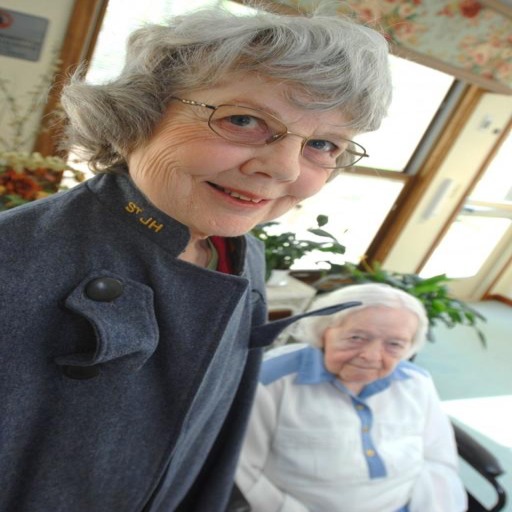} \\      
    \includegraphics[width=0.19\columnwidth]{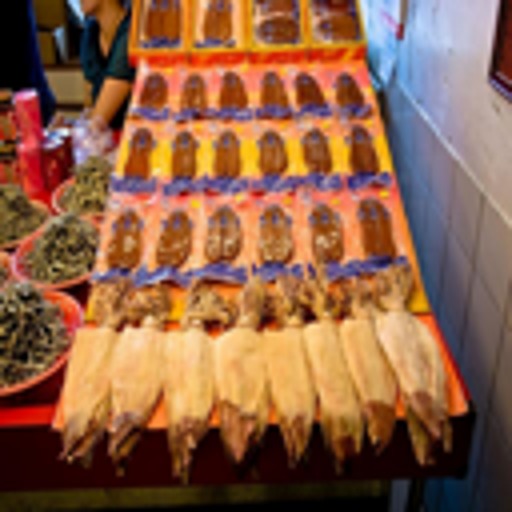} &
    \includegraphics[width=0.19\columnwidth]{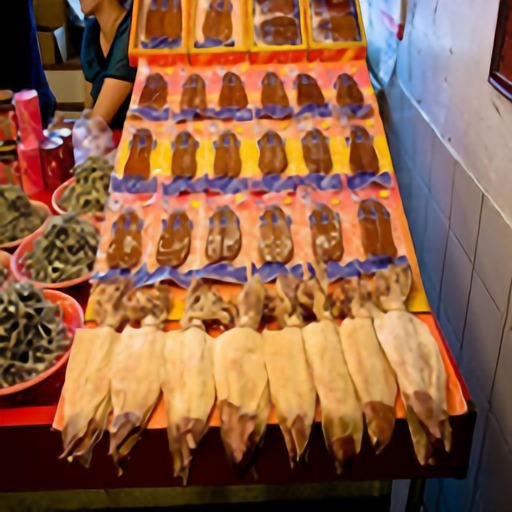} &
    \includegraphics[width=0.19\columnwidth]{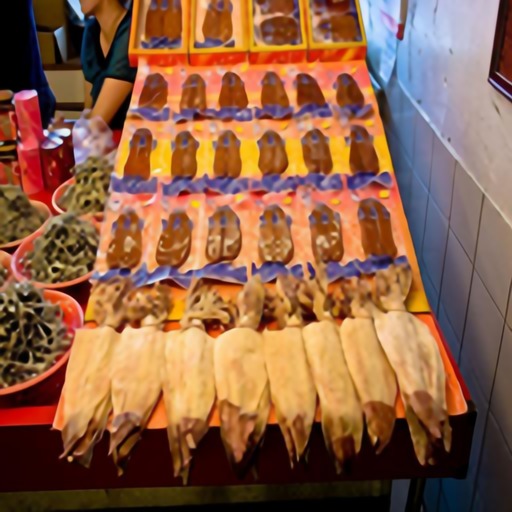} &
    \includegraphics[width=0.19\columnwidth]{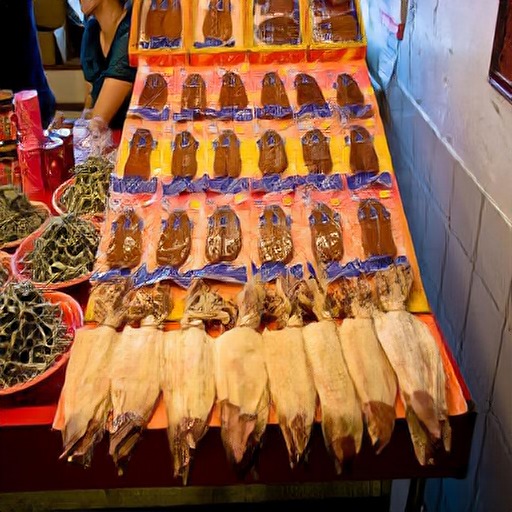} &
    \includegraphics[width=0.19\columnwidth]{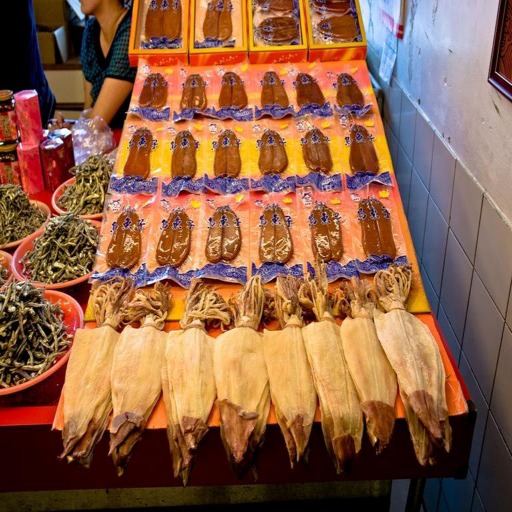} \\        
    \includegraphics[width=0.19\columnwidth]{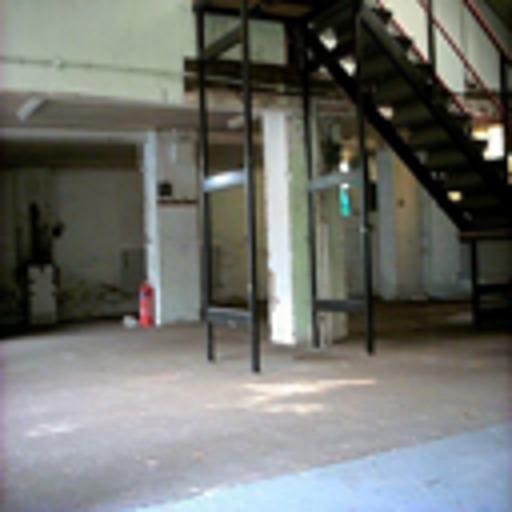} &
    \includegraphics[width=0.19\columnwidth]{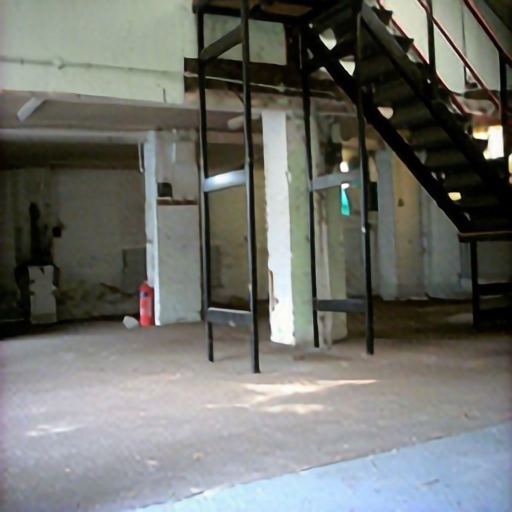} &
    \includegraphics[width=0.19\columnwidth]{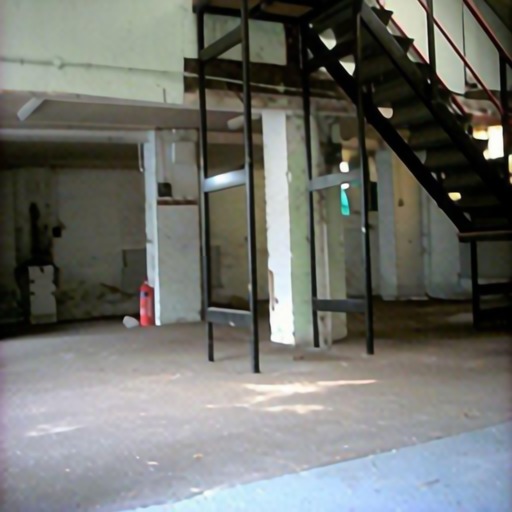} &
    \includegraphics[width=0.19\columnwidth]{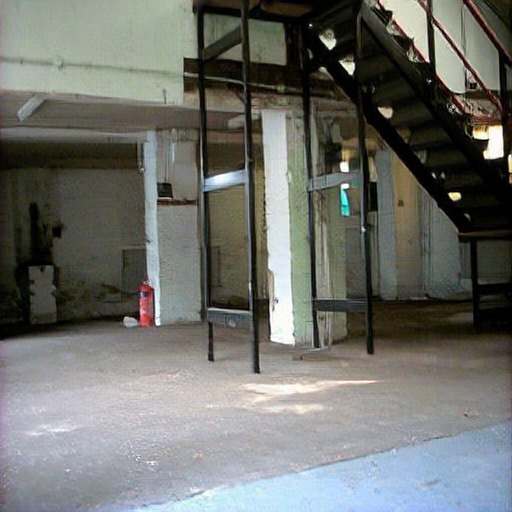} &
    \includegraphics[width=0.19\columnwidth]{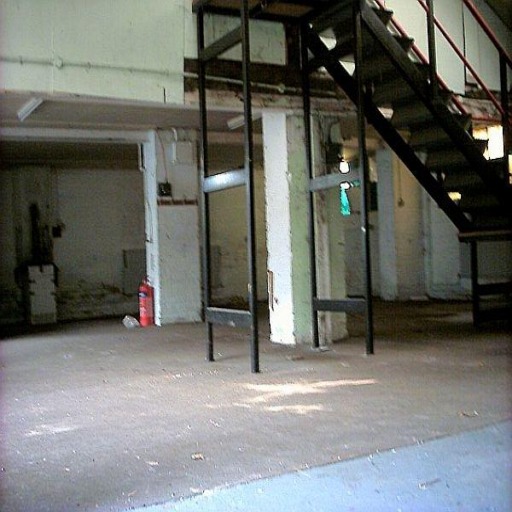} \\        
    \includegraphics[width=0.19\columnwidth]{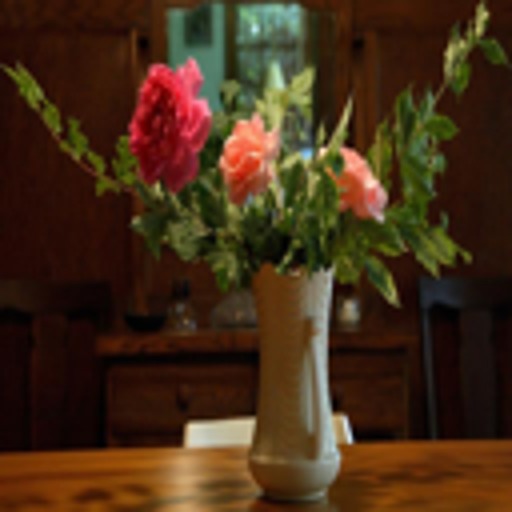} &
    \includegraphics[width=0.19\columnwidth]{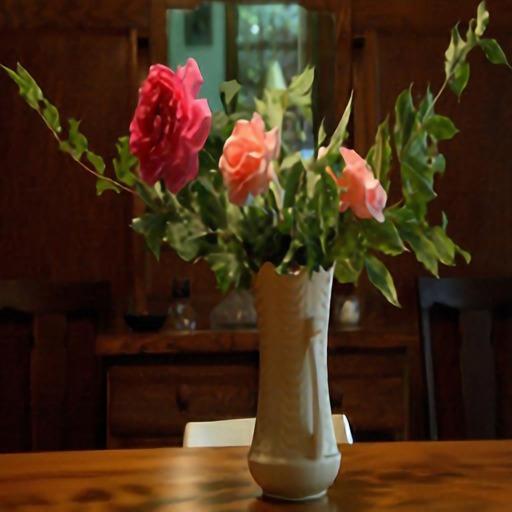} &
    \includegraphics[width=0.19\columnwidth]{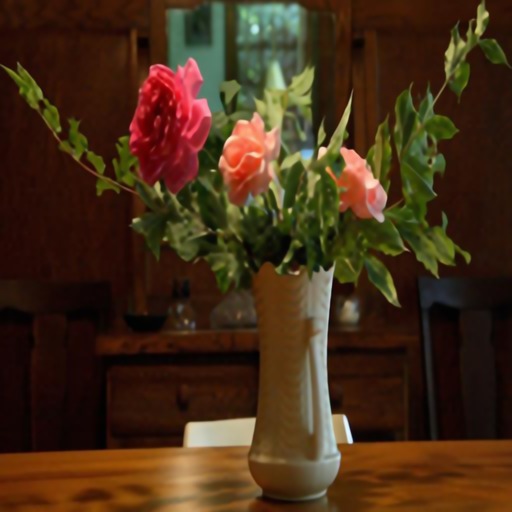} &
    \includegraphics[width=0.19\columnwidth]{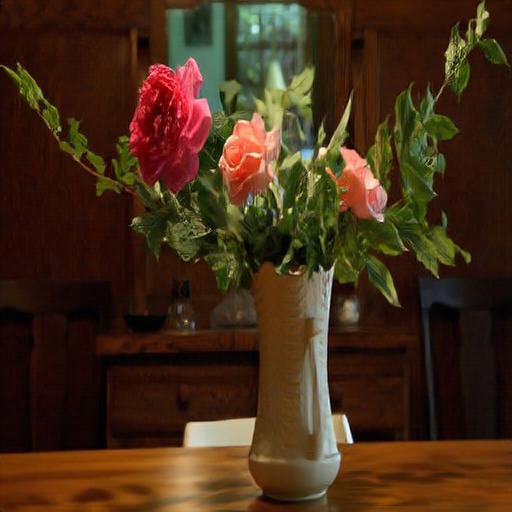} &
    \includegraphics[width=0.19\columnwidth]{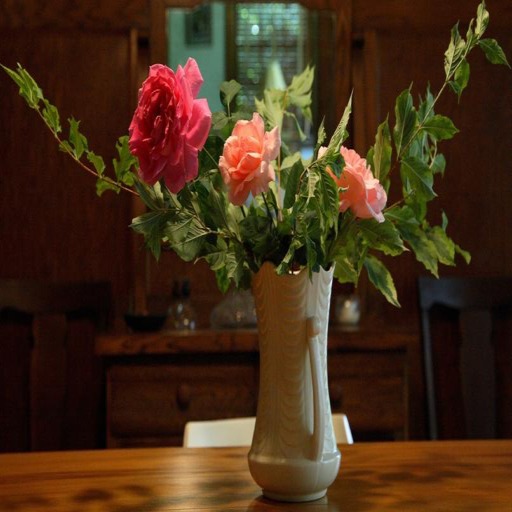} \\      
    \includegraphics[width=0.19\columnwidth]{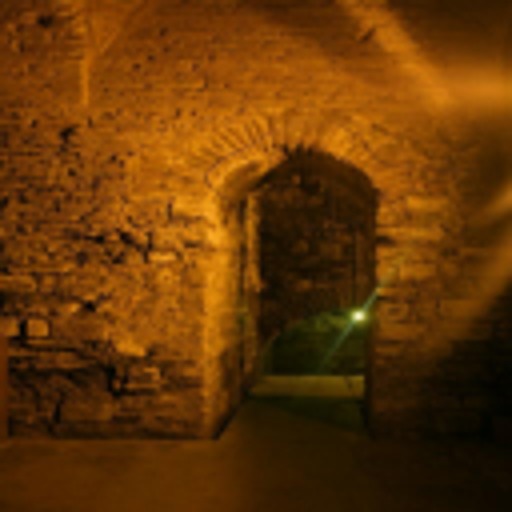} &
    \includegraphics[width=0.19\columnwidth]{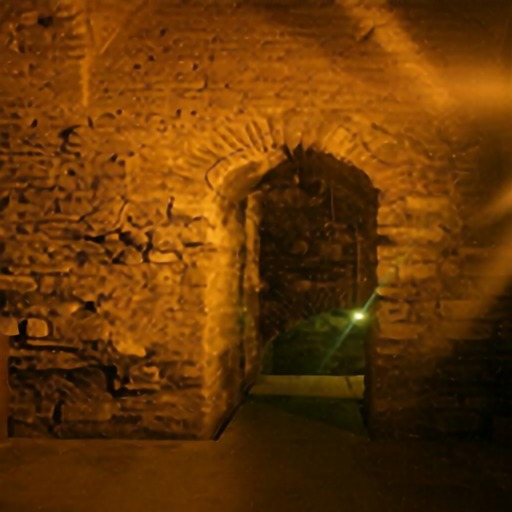} &
    \includegraphics[width=0.19\columnwidth]{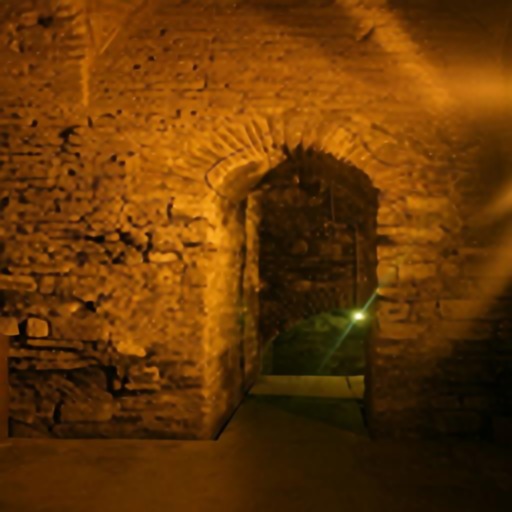} &
    \includegraphics[width=0.19\columnwidth]{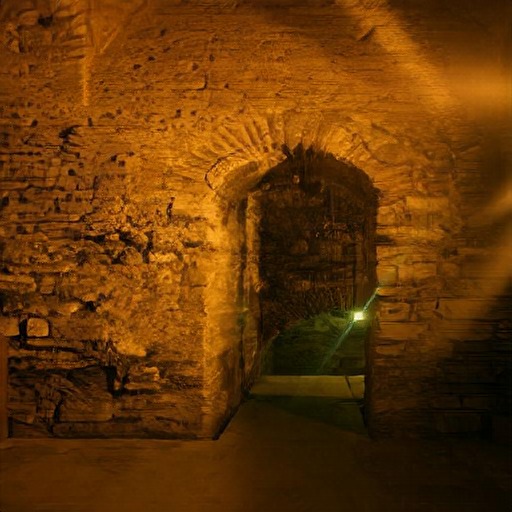} &
    \includegraphics[width=0.19\columnwidth]{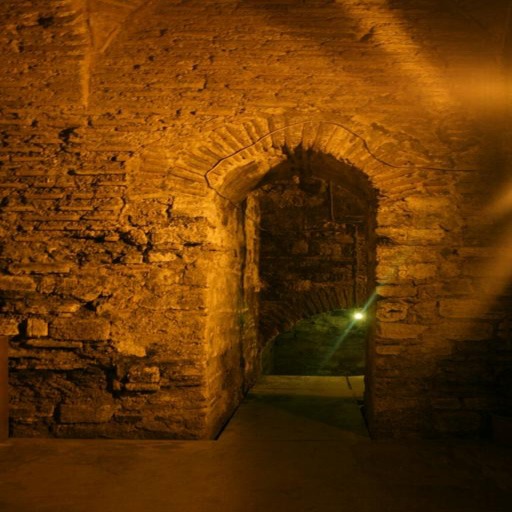} \\             
    \end{tabular}
    }
\label{fig:compare_superres_appendix}
\end{figure}

\clearpage

\subsection*{More Results of Our Approach}
\begin{figure}[h]
    \centering
    \scalebox{0.98}{
    \begin{tabular}{cc|cc}
    \multicolumn{4}{c}{}\\
    Input & PConv & Input & PConv \\
    \includegraphics[width=0.22\columnwidth]{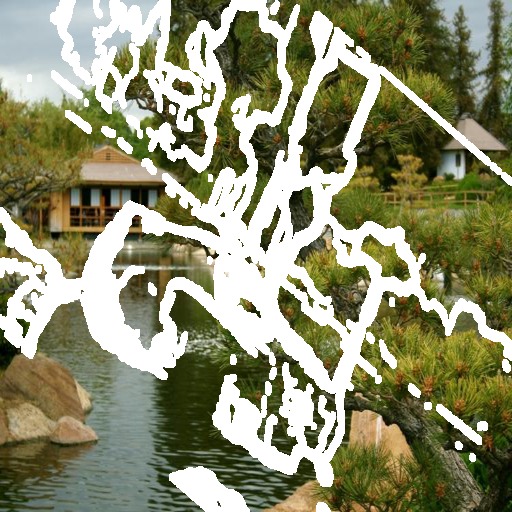} &
    \includegraphics[width=0.22\columnwidth]{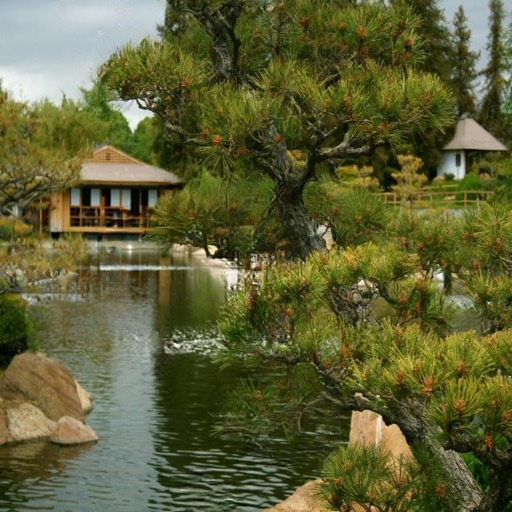} & 
    \includegraphics[width=0.22\columnwidth]{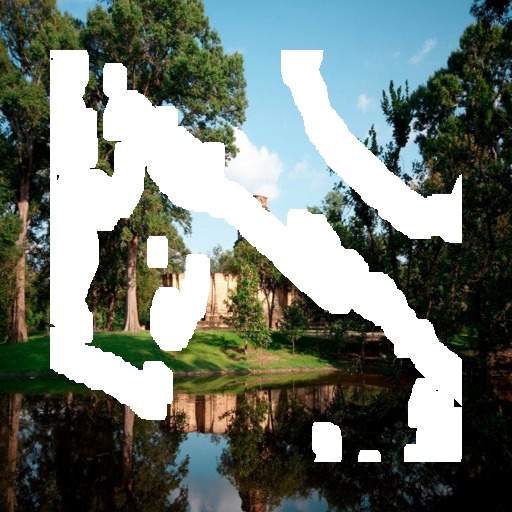} &
    \includegraphics[width=0.22\columnwidth]{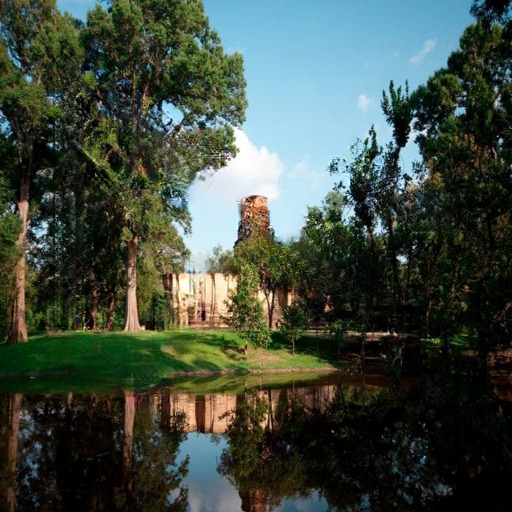} \\
    \includegraphics[width=0.22\columnwidth]{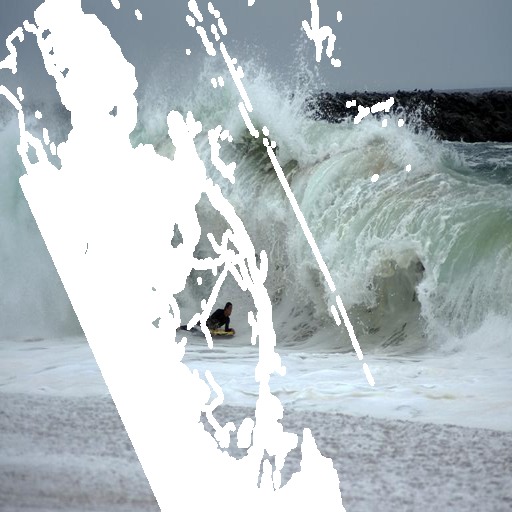} &
    \includegraphics[width=0.22\columnwidth]{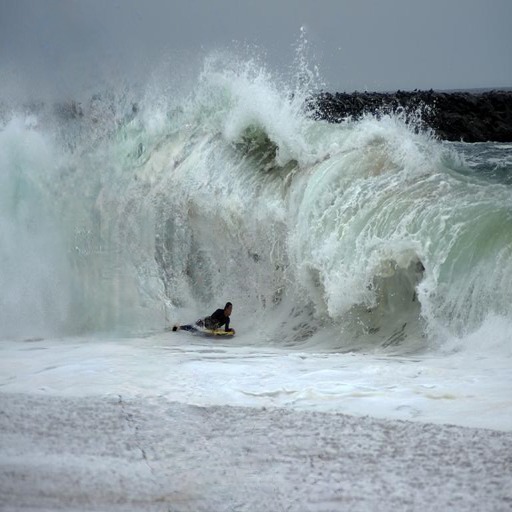} & 
    \includegraphics[width=0.22\columnwidth]{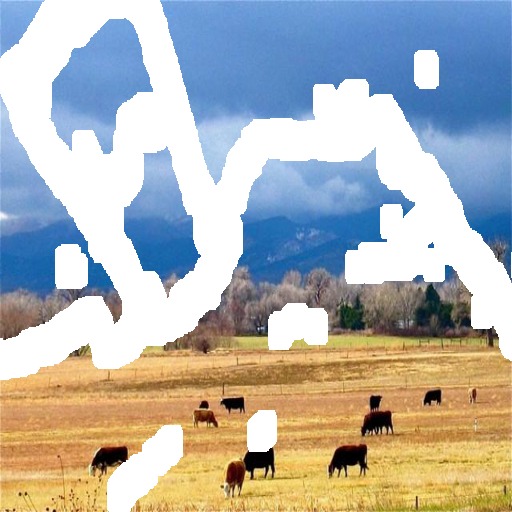} &
    \includegraphics[width=0.22\columnwidth]{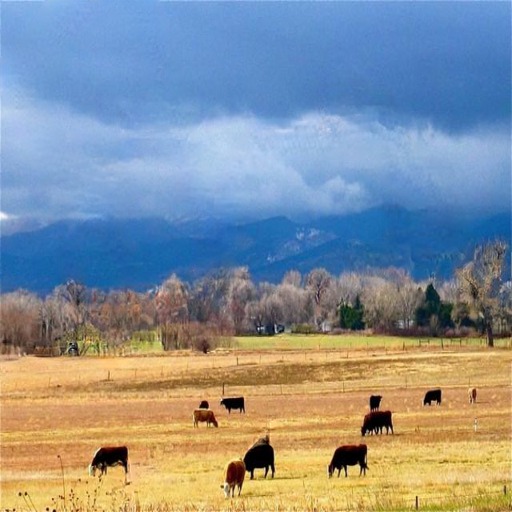} \\    
    \includegraphics[width=0.22\columnwidth]{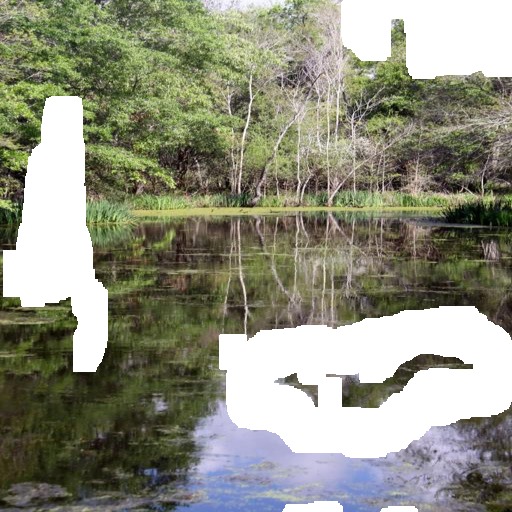} &
    \includegraphics[width=0.22\columnwidth]{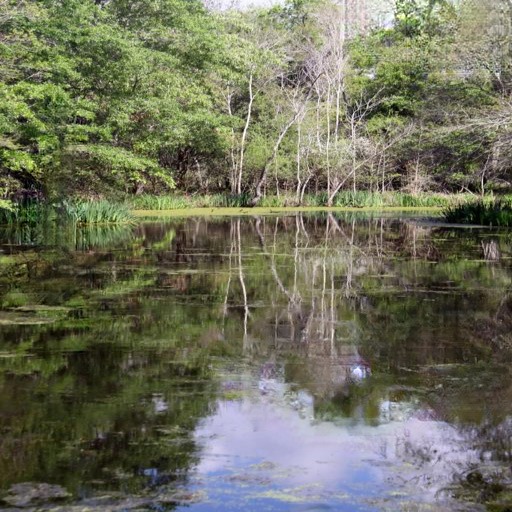} & 
    \includegraphics[width=0.22\columnwidth]{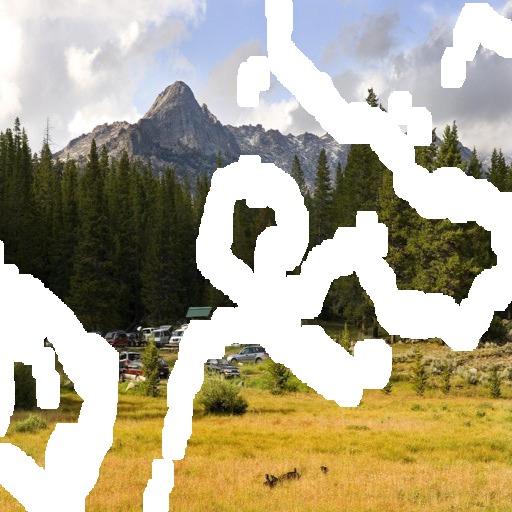} &
    \includegraphics[width=0.22\columnwidth]{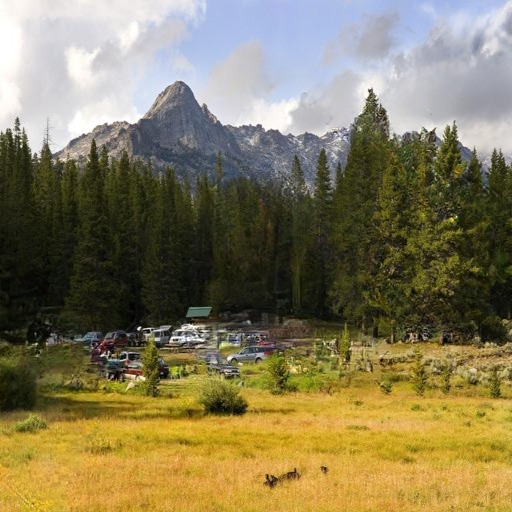} \\     
    \includegraphics[width=0.22\columnwidth]{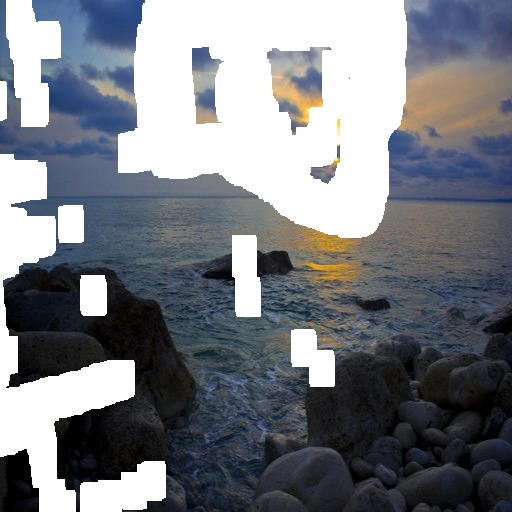} &
    \includegraphics[width=0.22\columnwidth]{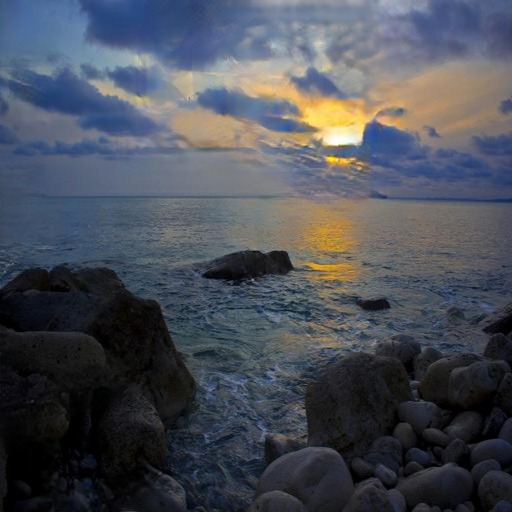} & 
    \includegraphics[width=0.22\columnwidth]{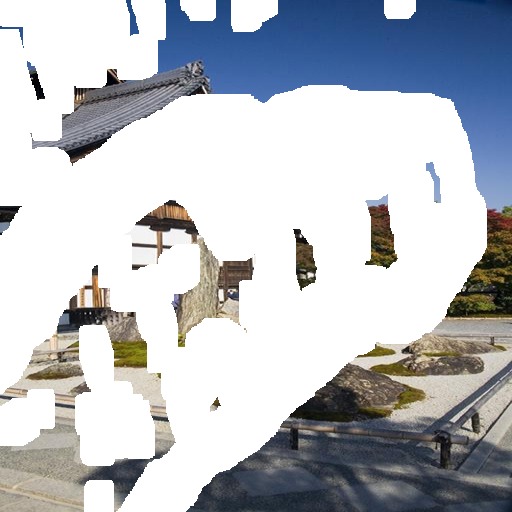} &
    \includegraphics[width=0.22\columnwidth]{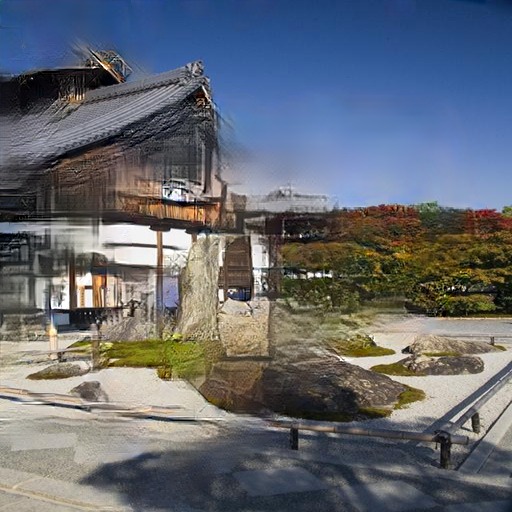} \\     
    \includegraphics[width=0.22\columnwidth]{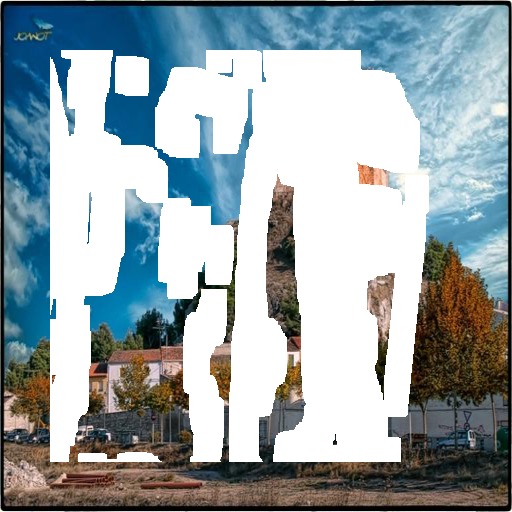} &
    \includegraphics[width=0.22\columnwidth]{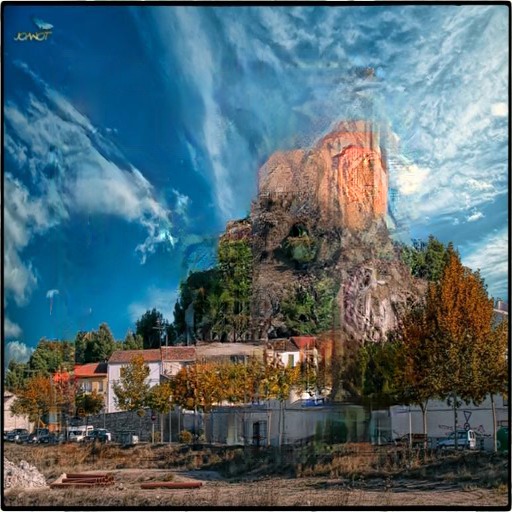} &
    \includegraphics[width=0.22\columnwidth]{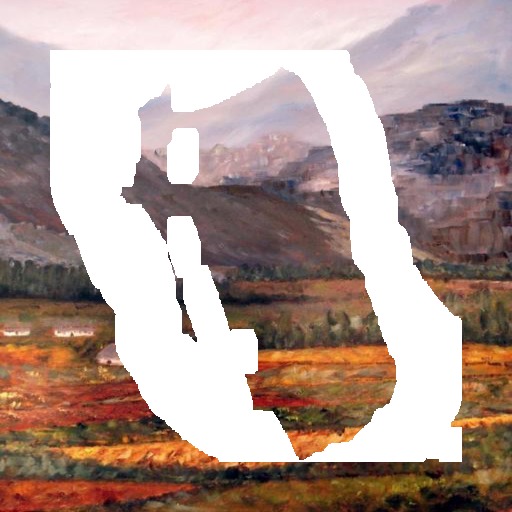} &
    \includegraphics[width=0.22\columnwidth]{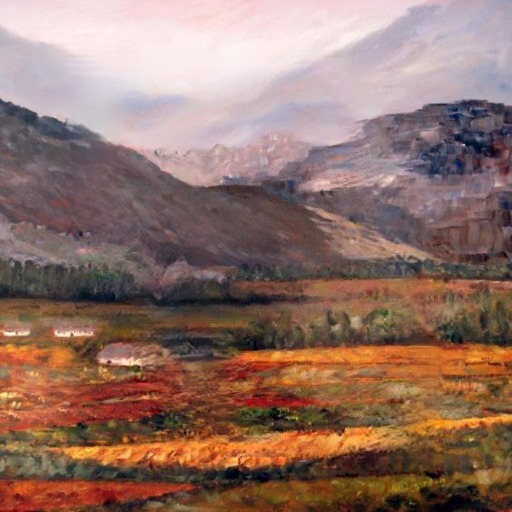} \\   
    \end{tabular}
    }
    % \label{fig:compare_regualr}
\end{figure}

\begin{figure}[h]
    \centering
    \scalebox{0.98}{
    \begin{tabular}{cc|cc}
    \multicolumn{4}{c}{}\\
    Input & PConv & Input & PConv \\
    \includegraphics[width=0.22\columnwidth]{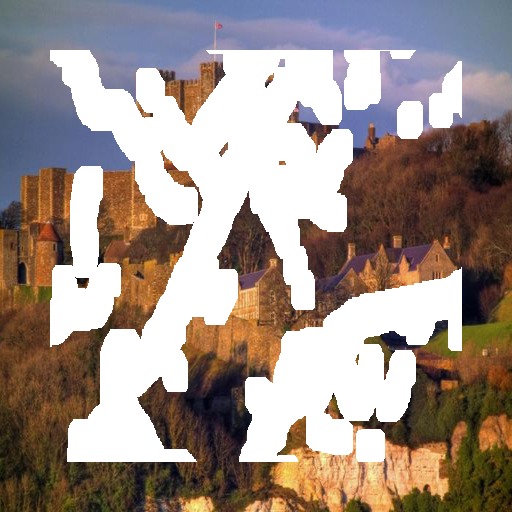} &
    \includegraphics[width=0.22\columnwidth]{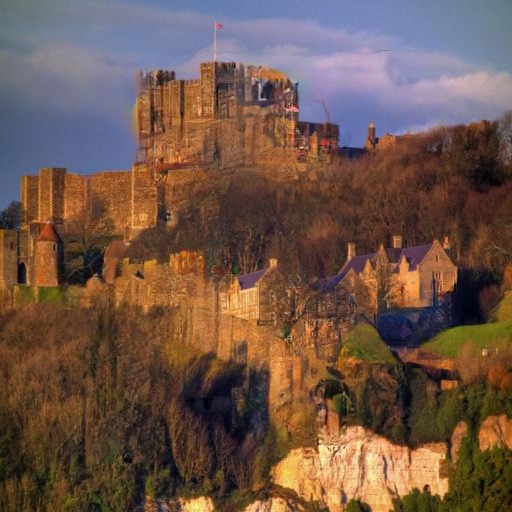} & 
    \includegraphics[width=0.22\columnwidth]{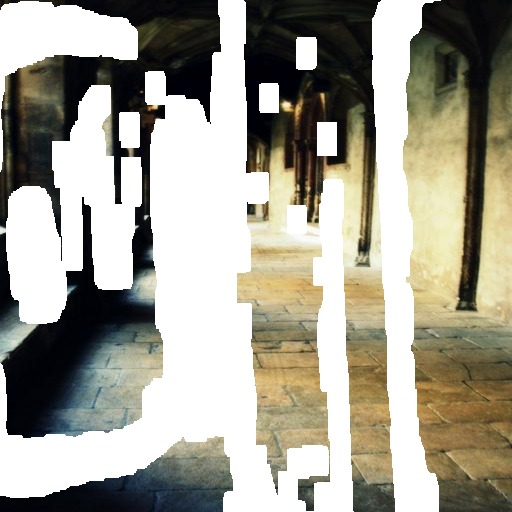} &
    \includegraphics[width=0.22\columnwidth]{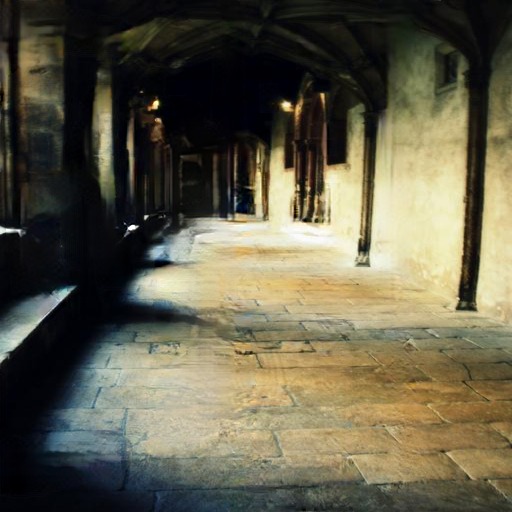} \\
    \includegraphics[width=0.22\columnwidth]{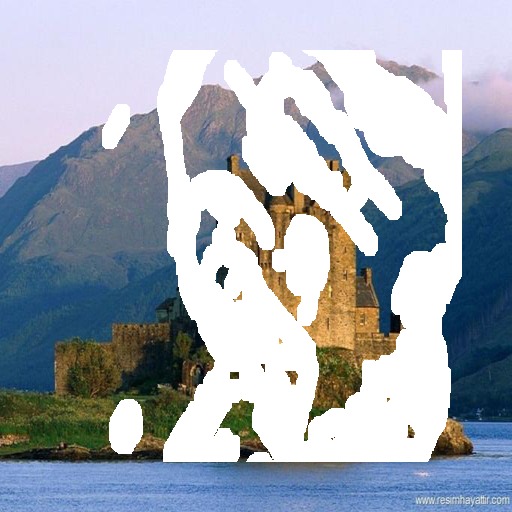} &
    \includegraphics[width=0.22\columnwidth]{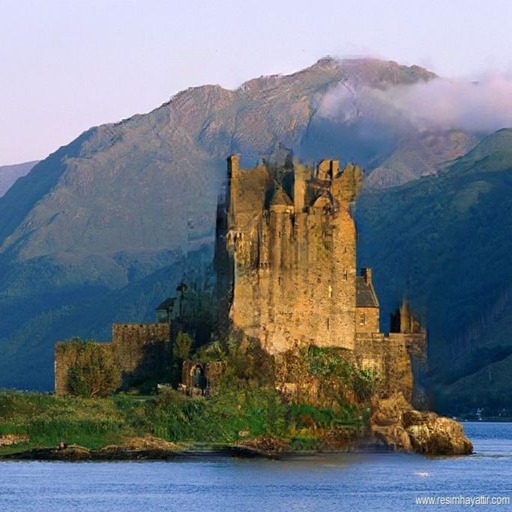} & 
    \includegraphics[width=0.22\columnwidth]{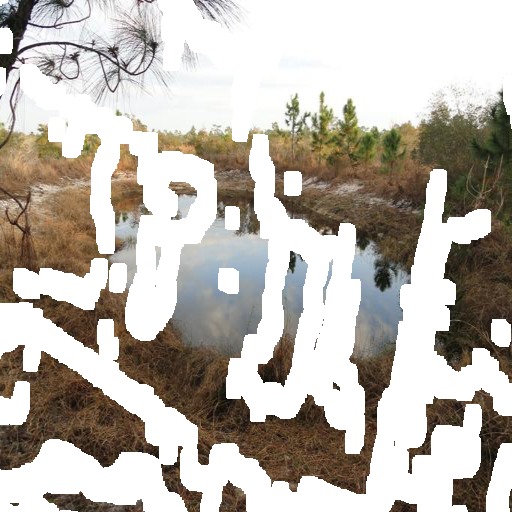} &
    \includegraphics[width=0.22\columnwidth]{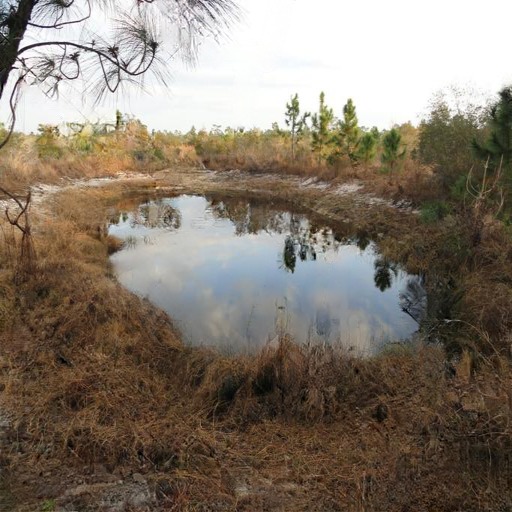} \\    
    \includegraphics[width=0.22\columnwidth]{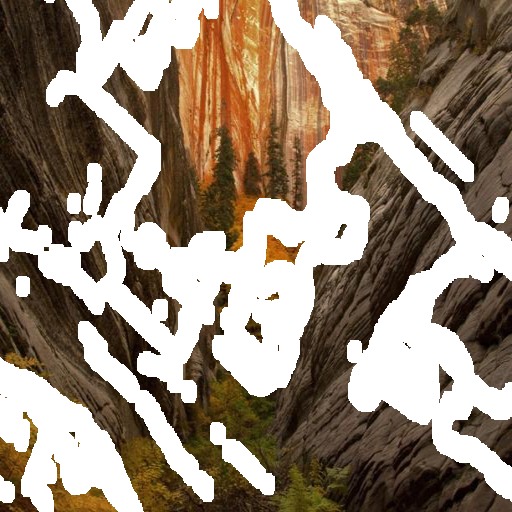} &
    \includegraphics[width=0.22\columnwidth]{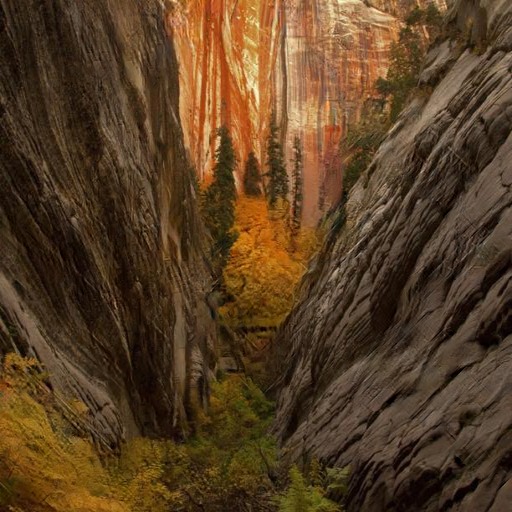} &     
    \includegraphics[width=0.22\columnwidth]{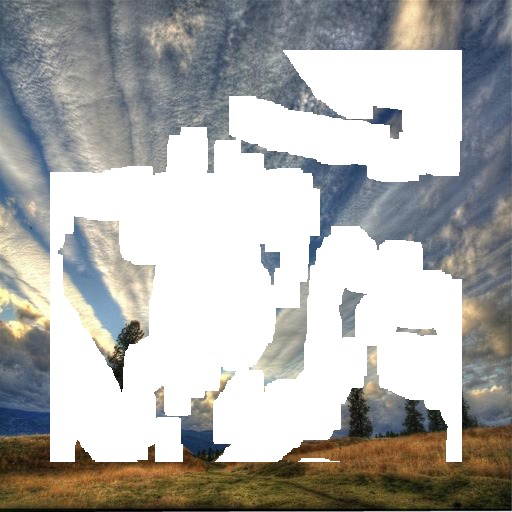} &
    \includegraphics[width=0.22\columnwidth]{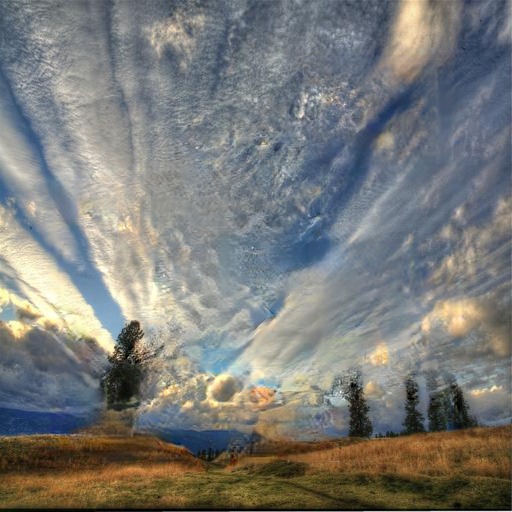} \\ 
    \includegraphics[width=0.22\columnwidth]{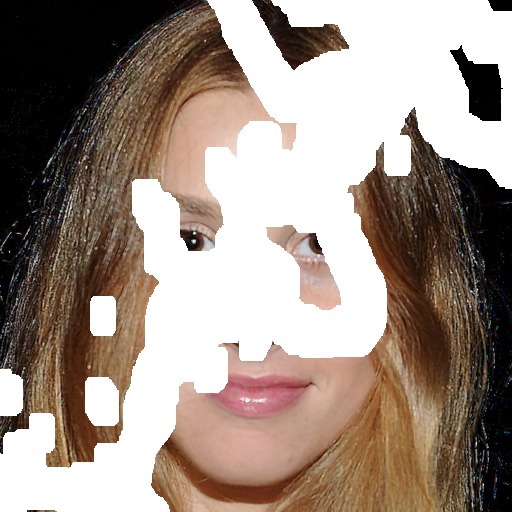} &
    \includegraphics[width=0.22\columnwidth]{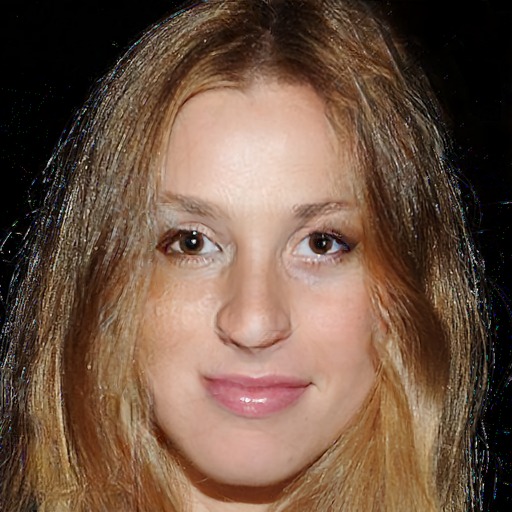} &     
    \includegraphics[width=0.22\columnwidth]{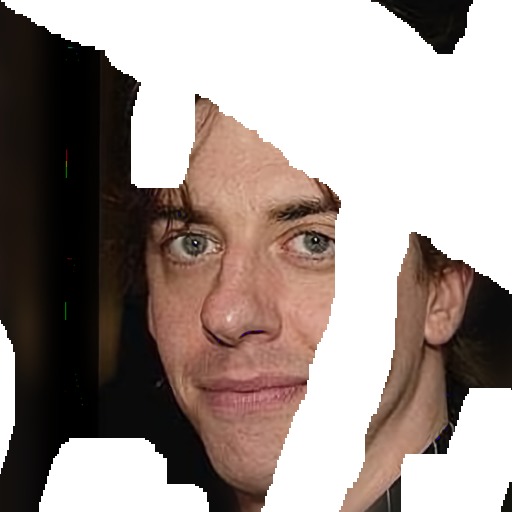} &
    \includegraphics[width=0.22\columnwidth]{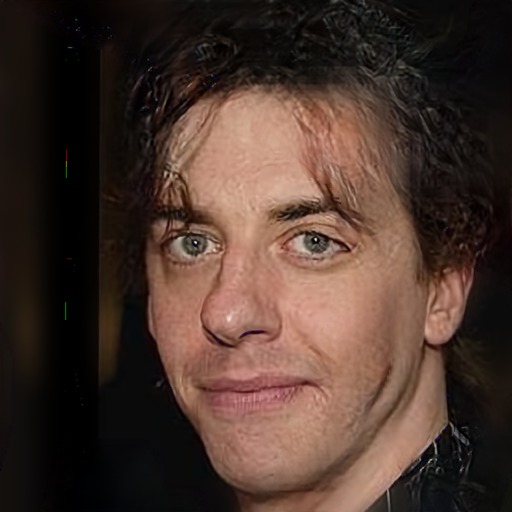} \\ 
    \includegraphics[width=0.22\columnwidth]{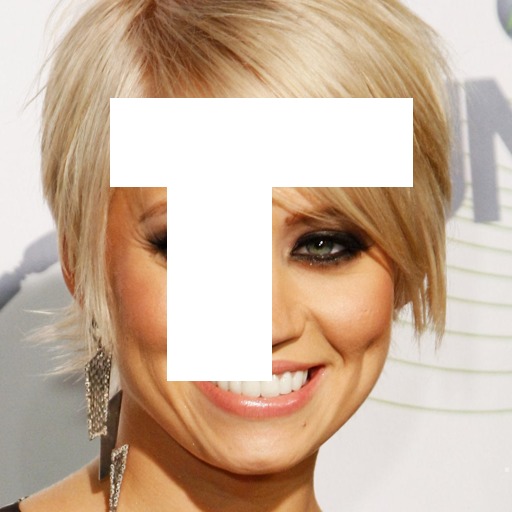} &
    \includegraphics[width=0.22\columnwidth]{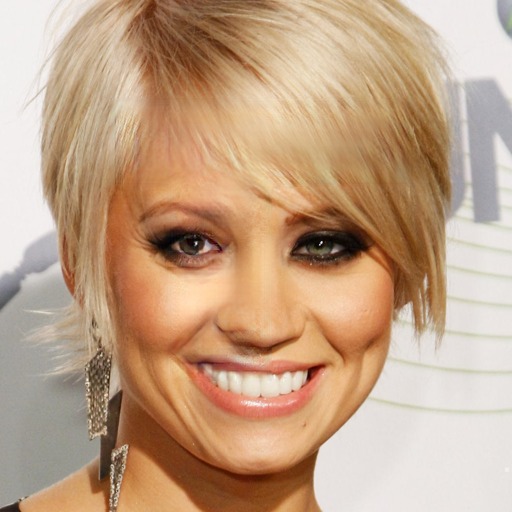} &     
    \includegraphics[width=0.22\columnwidth]{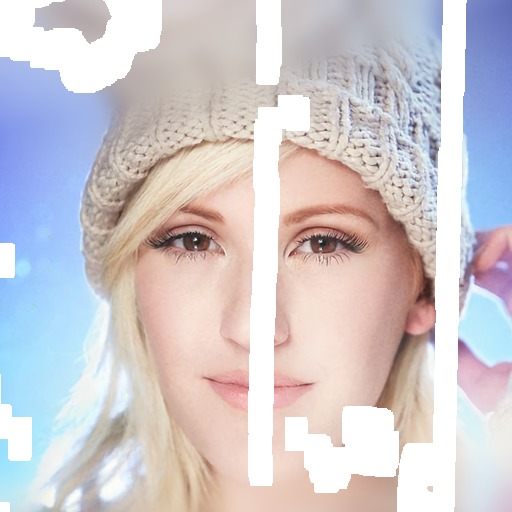} &
    \includegraphics[width=0.22\columnwidth]{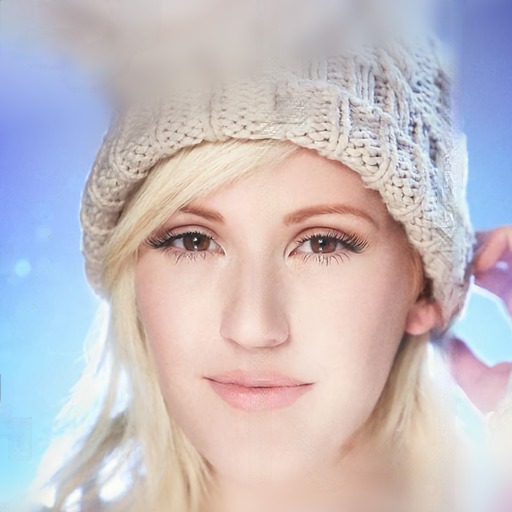} \\     
    \end{tabular}
    }
    % \label{fig:compare_regualr}
\end{figure}

\end{document}